\documentclass[11pt]{article}

\usepackage{amsmath}
\usepackage{graphicx}
\usepackage{enumerate}
\usepackage{natbib}
\usepackage{url} 

\usepackage{amssymb}
\usepackage{amsthm}
\usepackage{color}
\usepackage{mathrsfs}

\usepackage{epsfig}
\usepackage[linesnumbered,ruled]{algorithm2e}

\begin{document}

\def\spacingset#1{\renewcommand{\baselinestretch}%
{#1}\small\normalsize} \spacingset{1}



\newcommand{\be}{\begin{equation}}
\newcommand{\ee}{\end{equation}}
\newcommand{\bes}{\begin{equation*}}
\newcommand{\ees}{\end{equation*}}
\newcommand{\beqn}{\begin{eqnarray}}
\newcommand{\eeqn}{\end{eqnarray}}
\newcommand{\beqns}{\begin{eqnarray*}}
\newcommand{\eeqns}{\end{eqnarray*}}

\newcommand{\lkr}{\left(}
\newcommand{\lkv}{\left[}
\newcommand{\rkv}{\right]}
\newcommand{\rkr}{\right)}
\newcommand{\lfi}{\left\{}
\newcommand{\rfi}{\right\}}
\newcommand{\di}{\displaystyle}

\newcommand{\fr}[1]{(\ref{#1})}

\long\def\ignore#1{}

\newcommand{\bm}[1]{\mathbf{#1}} 

\newcommand{\bs}[1]{\boldsymbol{#1}}

\newcommand{\R}{\mathbb{R}}

\newcommand{\vart}{\vartheta}
\newcommand{\ro}{\varrho}
\newcommand{\ph}{\varphi}

\newcommand{\Te}{\Theta}
\newcommand{\Del}{\Delta}
\newcommand{\del}{\delta}
\newcommand{\dn}{\delta_n}
\newcommand{\af}{\alpha}
\newcommand{\eps}{\epsilon}
\newcommand{\Ga}{\Gamma}
\newcommand{\ga}{\gamma}
\newcommand{\te}{\theta}
\newcommand{\om}{\omega}
\newcommand{\lam}{\lambda}
\newcommand{\Up}{\Upsilon}
\newcommand{\up}{\upsilon}
\newcommand{\dz}{\zeta}
\newcommand{\sig}{\sigma}
\newcommand{\sgmd}{\sigma^2}
\newcommand{\Lam}{\Lambda}
\newcommand{\Om}{\Omega}
\newcommand{\Ups}{\Upsilon}

\newcommand{\EE}{\ensuremath{{\mathbb E}}}
\newcommand{\JJ}{\ensuremath{{\mathbb J}}}
\newcommand{\II}{\ensuremath{{\mathbb I}}}
\newcommand{\ZZ}{\ensuremath{{\mathbb Z}}}
\newcommand{\PP}{\ensuremath{{\mathbb P}}}
\newcommand{\QQ}{\ensuremath{{\mathbb Q}}}
\newcommand{\KK}{\ensuremath{{\mathbb K}}}
\newcommand{\RR}{\ensuremath{{\mathbb R}}}

\newcommand{\MISE}{\mbox{MISE}}
\newcommand{\Span}{\mbox{Span}}
\newcommand{\intg}{\mbox{int}}
\newcommand{\card}{\mbox{card}}
\newcommand{\Range}{\mbox{Range}}
\newcommand{\Var}{\mbox{Var}}
\newcommand{\Cov}{\mbox{Cov}}
\newcommand{\diag}{\mbox{diag}}
\newcommand{\supp}{\mbox{supp}}
\newcommand{\xil}{ {\it et. al }}
\newcommand{\std}{\mbox{std}}
\newcommand{\SNR}{\mbox{SNR}}
\newcommand{\Tr}{\mbox{Tr}}
\newcommand{\proj}{\mbox{proj}}
\newcommand{\Ber}{\mbox{Ber}}
\newcommand{\ERR}{\mbox{ERR}}
\newcommand{\rank}{\mbox{rank}}
\newcommand{\dist}{\mbox{dist}}
\newcommand{\vect}{\mbox{vec}}
\newcommand{\Pen}{\mbox{Pen}}
\newcommand{\Err}{\mbox{Err}}
\newcommand{\su}{\mbox{supp}}
\newcommand{\sgn}{\mbox{sign}}

\newtheorem{theorem}{Theorem}
\newtheorem{lemma}{Lemma}
\newtheorem{corollary}{Corollary}
\newtheorem{proposition}{Proposition}
\newtheorem{remark}{Remark}
\newtheorem{example}{Example}
\newtheorem{definition}{Definition}

\newcommand{\bo}{\mathbf{}}
\newcommand{\ba}{\mathbf{a}}
\newcommand{\bb}{\mathbf{b}}
\newcommand{\bc}{\mathbf{c}}
\newcommand{\bd}{\mathbf{d}}
\newcommand{\boe}{\mathbf{e}}
\newcommand{\bof}{\mathbf{f}}
\newcommand{\bh}{\mathbf{h}}
\newcommand{\bi}{\mathbf{i}}
\newcommand{\bq}{\mathbf{q}}
\newcommand{\bt}{\mathbf{t}}
\newcommand{\bu}{\mathbf{u}}
\newcommand{\bv}{\mathbf{v}}
\newcommand{\bw}{\mathbf{w}}
\newcommand{\bx}{\mathbf{x}}
\newcommand{\by}{\mathbf{y}}
\newcommand{\bz}{\mathbf{z}}

\newcommand{\bA}{\mathbf{A}}
\newcommand{\bB}{\mathbf{B}}
\newcommand{\bD}{\mathbf{D}}
\newcommand{\bI}{\mathbf{I}}
\newcommand{\bM}{\mathbf{M}}
\newcommand{\bP}{\mathbf{P}}
\newcommand{\bQ}{\mathbf{Q}}
\newcommand{\bR}{\mathbf{R}}
\newcommand{\bG}{\mathbf{G}}
\newcommand{\bH}{\mathbf{H}}
\newcommand{\bU}{\mathbf{U}}
\newcommand{\bV}{\mathbf{V}}
\newcommand{\bW}{\mathbf{W}}
\newcommand{\bX}{\mathbf{X}}
\newcommand{\bY}{\mathbf{Y}}
\newcommand{\bZ}{\mathbf{Z}}

\newcommand{\calB}{{\cal B}}

\newcommand{\bzero}{\mathbf{0}}

\newcommand{\bte}{\mbox{\mathversion{bold}$\te$}}
\newcommand{\bupsilon}{\mbox{\mathversion{bold}$\upsilon$}}
 \newcommand{\btheta}{\mbox{\mathversion{bold}$\theta$}}
\newcommand{\bxi}{\mbox{\mathversion{bold}$\xi$}}
\newcommand{\boeta}{\mbox{\mathversion{bold}$\xi$}}
\newcommand{\bbe}{\mbox{\mathversion{bold}$\beta$}}
\newcommand{\bzeta}{\mbox{\mathversion{bold}$\zeta$}}
\newcommand{\bphi}{\mbox{\mathversion{bold}$\ph$}}
\newcommand{\bpsi}{\mbox{\mathversion{bold}$\psi$}}
\newcommand{\beps}{\mbox{\mathversion{bold}$\eps$}}
\newcommand{\bobeta}{\mbox{\mathversion{bold}$\beta$}}
\newcommand{\bgamma}{\mbox{\mathversion{bold}$\gamma$}}
\newcommand{\bmu}{\mbox{\mathversion{bold}$\mu$}}

\newcommand{\hbtheta}{\widehat{\btheta}}

\newcommand{\bbJ}{(\bb_J)}
\newcommand{\tilJ}{\tilde{J}}
\newcommand{\brJ}{\breve{J}}
\newcommand{\tilL}{\tilde{L}}
\newcommand{\tilC}{\tilde{C}}

\newcommand{\bPhi}{\mbox{\mathversion{bold}$\Phi$}}
\newcommand{\bUp}{\mbox{\mathversion{bold}$\Up$}}
\newcommand{\bPsi}{\mbox{\mathversion{bold}$\Psi$}}
\newcommand{\bLam}{\mbox{\mathversion{bold}$\Lambda$}}
\newcommand{\bOm}{\mbox{\mathversion{bold}$\Om$}}

\newcommand{\Jc}{{J_{*}^c}}

\newcommand{\calD}{{\mathcal{D}}}
\newcommand{\calG}{{\mathcal G}}
\newcommand{\calJ}{{\mathcal{J}}}
\newcommand{\calmu}{{\mathcal M}_{\mu}}
\newcommand{\calL}{{\mathcal{L}}}
\newcommand{\calM}{{\mathcal{M}}}
\newcommand{\calO}{{\mathcal{O}}}
\newcommand{\calP}{{\mathcal{P}}}
\newcommand{\calX}{{\cal{X}}}
\newcommand{\calY}{{\cal{Y}}}
\newcommand{\calZ}{{\cal{Z}}}
\newcommand{\calV}{{\cal{V}}}
\newcommand{\calW}{{\cal{W}}}

\newcommand{\calH}{{\cal H}}
\newcommand{\calT}{{\cal T}}
\newcommand{\calS}{{\cal S}}
\newcommand{\calN}{{\cal N}}
\newcommand{\calF}{{\mathcal F}}

\newcommand{\sumjni}{\displaystyle\sum_{j\neq i}}
\newcommand{\sumjn}{\displaystyle\sum_{j=1}^n}
\newcommand{\sumin}{\displaystyle\sum_{i=1}^n}
\newcommand{\sumJ}{\sum_{j \in J_{*}}}
\newcommand{\sumJc}{\sum_{j \in \Jc}}
\newcommand{\sumijn}{\displaystyle\sum_{i,j =1}^n}

\newcommand{\scrP}{\mathscr{P}}
\newcommand{\scrPZ}{\mathscr{P}_{Z}}
\newcommand{\scrPZK}{\mathscr{P}_{Z,K}}
\newcommand{\scrPhZK}{\mathscr{P}_{\hat{Z}_K,K}}

\newcommand{\norm} [1]{\left\|#1 \right\|}
\newcommand{\etal}{{\it et al.\ }}
 
\newcommand{\pijbAZh} {\Pi_{J^{(k,l)}} (A^{(k,l)} (\hat{Z} , \hat{K}))}
\newcommand{\pijbAZ} { \Pi_{J^{(k,l)}} (A^{(k,l)} (Z,K))}
\newcommand{\pijhbAZh} {\Pi_{\hat{J}^{(k,l)}} (A^{(k,l)} (\hat{Z}, \hat{K}))}
\newcommand{\pijhbPsZh} {\Pi_{\hat{J}^{(k,l)}} ({P_{*}}^{(k,l)} (\hat{Z}, \hat{K}))}
\newcommand{\pijbTheta} {\Pi_{J^{(k,l)}} (\Theta^{(k,l)})}
\newcommand{\pijkl}{\Pi_{J^{(k,l)}}}
\newcommand{\pihjkl}{\Pi_{\hat{J}^{(k,l)}}}
\newcommand{\pibrjkl}{\Pi_{\breve{J}^{(k,l)}}}

\newcommand{\pijAZ} {\Pi_{J} (A^{(k,l)}(\hat{Z}, \hat{K})) }
\newcommand{\pijhAZh} {\Pi_{\hat{J} } (A (\hat{Z} , \hat{K})) }
\newcommand{\pijhPsZh} {\Pi_{\hat{J}} ({P_{*}} (\hat{Z}, \hat{K})) }
\newcommand{\pijhEZh} {\Pi_{\hat{J} } (\Xi (\hat{Z}, \hat{K})) }
\newcommand{\pijThetah} {\Pi_{J} (\Theta (\hat{Z}, \hat{K})) }

\newcommand{\tC}{\tilde{C}}
\newcommand{\tibeta}{\tilde{\beta}}

\newcommand{\lowc}{\underline{c}}
\newcommand{\highc}{\bar{c}}

\newcommand{\hbtm}{\hat{\bm t}^{(m)}}

 \newcommand{\oPen}{\overline{\Pen}}

\newcommand{\hn}{\hat{n}}
\newcommand{\rhon}{\rho_n}

\newcommand{\hB}{\widehat{B}}
\newcommand{\hF}{\widehat{F}}
\newcommand{\hC}{\widehat{C}}
\newcommand{\hG}{\widehat{G}}
\newcommand{\hZ}{\widehat{Z}}
\newcommand{\hThe}{\widehat{\Theta}}
\newcommand{\hH}{\widehat{H}}
\newcommand{\hK}{\hat{K}}
\newcommand{\hL}{\widehat{L}}
\newcommand{\hU}{\widehat{U}}
\newcommand{\hW}{\widehat{W}}
\newcommand{\hLam}{\widehat{\Lam}}
\newcommand{\hTe}{\widehat{\Te}}
\newcommand{\hD}{\widehat{D}}
\newcommand{\hbd}{\widehat{\bd}}

\newcommand{\tcalV}{\tilde{\calV}}
\newcommand{\tcalW}{\tilde{\calW}}
\newcommand{\hcalW}{\widehat{\calW}}
\newcommand{\hbG}{\widehat{\bG}}
\newcommand{\hcalH}{\widehat{\calH}}
\newcommand{\hbH}{\widehat{\bH}}
\newcommand{\hUps}{\widehat{\Ups}}
\newcommand{\hPsi}{\widehat{\Psi}}

\newcommand{\tB}{\tilde{B}}
\newcommand{\tH}{\tilde{H}}
\newcommand{\tP}{\tilde{P}}
\newcommand{\tU}{\tilde{U}}
\newcommand{\tLam}{\tilde{\Lam}}
\newcommand{\tilOm}{\tilde{\Omega}}

\newcommand{\wt}{\widetilde}
\newcommand{\wh}{\widehat}
\newcommand{\ovr}{\overline}
\newcommand{\mc}{\mathcal}
\newcommand{\wc}{\check}
 
\newcommand{\scrPZC}{\mathscr{P}_{Z,C}}
\newcommand{\lowC}{\underline C}

\newcommand{\upl}{^{(l)}}
\newcommand{\upm}{^{(m)}}

\newcommand{\tim}{\times_3}
\newcommand{\barG}{\bar{G}}
\newcommand{\hbarG}{\bar{\hG}}

\long\def\ignore#1{}

\newcommand{\bbC}{\mathbb{C}}


\newcommand {\colred}[1] {\textcolor{red}{#1}}
\newcommand {\colblue}[1] {\textcolor{blue}{#1}}
\newcommand {\colyellow}[1] {\textcolor{yellow}{#1}}
\newcommand {\colblack}[1] {\textcolor{black}{#1}}
\newcommand {\colgreen}[1] {\textcolor{green}{#1}}
\newcommand {\colsred}[1] {\textcolor{OrangeRed}{#1}}
 


  \title{\bf Sparse Subspace Clustering in Diverse Multiplex  Network Model}
  \author{Majid Noroozi \\ 
    Department of Mathematical Sciences, University of Memphis\\
    and \\
    Marianna Pensky \thanks{
    The author  gratefully acknowledges partially support  by National Science Foundation
   (NSF)  grant DMS-2014928}\hspace{.2cm}\\
    Department of Mathematics, University of Central Florida }
\date{}
  \maketitle

\bigskip
\begin{abstract}
The paper considers the  DIverse MultiPLEx (DIMPLE) network  model, introduced in Pensky and Wang (2021), 
where all layers of the network have the same collection of nodes and are equipped with the Stochastic Block Models. 
In addition, all layers can be partitioned into groups with the same community structures, although the  layers
in the same group may have different matrices of block connection probabilities. The DIMPLE model generalizes 
a multitude of papers that study multilayer networks with the same community structures in all layers,
as well as the Mixture  Multilayer Stochastic Block Model  (MMLSBM),   where the  layers
in the same group  have identical matrices of block connection probabilities. 
While Pensky and Wang (2021) applied spectral clustering to the proxy of the  adjacency tensor,
the present paper uses Sparse Subspace Clustering (SSC)  for identifying groups of layers with identical community structures.
Under mild conditions, the latter leads to the strongly consistent between-layer clustering. In addition,
SSC  allows to handle much larger networks than methodology of Pensky and Wang (2021), 
and is perfectly suitable for application of parallel computing.
\end{abstract}

\noindent%
\vfill

\newpage
\spacingset{1.0} 




\section{Introduction}
\label{sec:intro}
 
Network models are an important tool  for describing and analyzing complex systems in many areas such as the social, 
biological, physical, and engineering sciences. Originally, almost all studies of networks were focused on a single network, 
 that is completely represented by   a set of  nodes and edges. 
Over the last decade, many models have been introduced to describe more complex networks. Specifically, 
the existence of many real networks with community structure has generated a surge of interest in studying Stochastic 
Block Model(SBM)  and its extensions (see, e.g.,  \cite{JMLR:v18:16-480},  \cite{Karrer2011StochasticBA},
 \cite{doi:10.1080/0022250X.1971.9989788},  \cite{RePEc:bla:jorssb:v:80:y:2018:i:2:p:365-386}).

Recently, the focus has changed to analysis of a   multilayer network \cite{kivela2014multilayer}, a powerful representation of relational data 
in which different individual networks evolve or interact with each other. In addition to a node set and an edge set, a multilayer network 
includes a layer set, whose each layer represents a different type of relation among those nodes. For example, a general 
multilayer network could be used to represent an urban transportation network,
where nodes might be stations in the city and each layer might represent a mode of transportation such as buses, metro, rail, etc.
While the term  ``multilayer network''  is often used in a more general context, we focus on the multilayer networks 
where the same set of nodes appears on every layer, and there are no edges between two different layers. 
Following  \cite{macdonald2021latent}, we call this multilayer network a {\it multiplex network}. 
One such example is a collection of brain connectivity networks of several individuals, where each layer  corresponds 
to a brain connectivity network of an individual.

A   time-varying network representing different states of a single network over time, 
can also be viewed as a particular case of the multiplex network.
The difference between those models and the multilayer network is that, 
in a dynamic network, the layers are ordered according to time instances, while in a multiplex network 
the enumeration of layers is completely arbitrary.

In this paper, we study a multiplex network where each layer is enabled with a community structure.
One of the problems in multilayer and dynamic networks is community detection with many important applications. 
While in such networks different layers have different forms of connections, it is often the case that one underlying 
unobserved community structure is in force. For example, in the multilayer Twitter networks in \cite{greene2013producing}, 
ground truth community memberships can be assigned to the users (nodes) based on some fundamental attributes 
(e.g., political views, country of origin, football clubs) that are independent 
of the observed twitter interactions, whereas the interactions 
provide multiple sources of information about the same latent community structure. Combining information from these multiple 
sources would then lead to enhanced performance in the consensus community detection (\cite{paul2020spectral}).

The assumption of one common community structure may not be true in some applications. It is often the case that 
there are groups of layers that are similar in some sense, and layers within each group share the same community 
structure, but each group has different community structure. One example is the worldwide food trading networks, 
collected by \cite{de2015structural}, which has been widely analyzed in literature 
(see, e.g., \cite{jing2020community}, \cite{macdonald2020latent}, among others). The data present 
an international trading network, in which layers represent different food products, nodes are countries, and 
edges at each layer represent trading relationships of a specific food product among countries. 
Two types of products, e.g. unprocessed and processed foods, can be considered as two groups of layers 
where each group has its own pattern of trading among the countries. While some large countries import/export unprocessed food 
from  and/or to a great number of other countries worldwide, for processed foods, countries 
are mainly clustered by the geographical location, i.e., countries in the same continent have closer trading ties
(\cite{jing2020community}).


In this paper, we consider a multilayer network where each of the  layers is equipped with the Stochastic Block Model (SBM).
Specifically, we are interested in analyzing the DIverse MultiPLEx (DIMPLE) network model introduced in \cite{pensky2021clustering}.
In this model, there are several types of layers, each of them is equipped with a distinct community structure, 
while the matrices of block probabilities can take different values in each of the layers.
 
The DIMPLE model generalizes a multitude of papers where communities persist throughout the network 
(\cite{bhattacharyya2020general}, \cite{lei2021biasadjusted}, \cite{10.1093/biomet/asz068},
 \cite{paul2016}, \cite{paul2020}). 
In particular, it includes the networks where the block probabilities take only finite number of values, as it happens
in ''checker board'' and tensor block models (\cite{JMLR:v21:18-155}, \cite{han2021exact},  \cite{NEURIPS2019_9be40cee}), 
as well as more complex networks, where communities persist through all layers of the network
but the matrices of block probabilities vary from one layer to another (see, e.g., 
\cite{bhattacharyya2020general}, \cite{lei2021biasadjusted}, \cite{10.1093/biomet/asz068},
 \cite{paul2016}, \cite{paul2020}  and references therein).
In fact, the DIMPLE network model can be viewed as a concatenation of the latter type of networks, where the layers
are  scrambled.   In addition, the recently introduced  {\bf M}ixture {\bf M}ulti{\bf L}ayer {\bf S}tochastic
{\bf B}lock {\bf M}odel  (MMLSBM) (see \cite{fan2021alma} and  \cite{TWIST-AOS2079}), where  all layers can be 
partitioned into a few different types, with each   type of   
layers equipped with its own  community structure and a   matrix of connection probabilities, is a particular case 
of the DIMPLE model, where each type of layers has its own specific block probability matrix.

\cite{pensky2021clustering} developed clustering procedures for finding layers with similar community structures, 
and also for finding communities in those layers. The authors showed that the methodologies used in the networks with the 
persistent community structure, as well as the ones designed for the MMLSBM, cannot be applied to the DIMPLE model.
The algorithms in \cite{pensky2021clustering}  are based on the spectral clustering. In particular, community detection is achieved 
by   clustering of the vectorized versions of the spectral projection matrices of the  layer networks.
Consequently, for an $n$-node multilayer network, it requires clustering of vectors in $n(n-1)/2$-dimensional space.
While the methodology works well for smaller $n$, it becomes extremely challenging when $n$ grows. 
For this reason, all simulations in \cite{pensky2021clustering} are carried out for relatively small values of $n$.

In the present paper, we propose to use    Subspace Clustering  for finding groups of layers with similar community structures.
Indeed, in what follows, we shall show  that the vectorized probability matrices of such layers all belong to the same low-dimensional subspace.
The subspace clustering relies on self-representation of the vectors to partition them into clusters. Consequently, one has to solve 
a regression  problem for each vector separately to find the matrix of weights, which is usually of much smaller size. 
Subsequently, some kind of spectral clustering is   applied to the weight matrix.   Subspace Clustering  is a very 
common technique in the  computer vision field. In particular, we apply Sparse Subspace Clustering (SSC) approach to identify those groups. 
We provide a review of the SSC technique in Section~\ref{sec:methods}. Although the SSC approach has been recently used in the some network models 
(see, e.g.,  \cite{noroozi2021hierarchy},  \cite{noroozi2021sparse} and \cite{noroozi2021estimation}), to the best of our knowledge, 
it has not been applied to   multilayer networks. Moreover,   this paper is the first one to offer assessment of 
clustering precision of an SSC-based algorithm   applied to   Bernoulli type data.
This requires a different set of assumptions from a traditional application of SSC to Gaussian data, and a novel 
clustering algorithm.

In this paper, we consider   the problem of clustering of layers into the sets of layers with the identical community structures
(the between-layer clustering)  as well as identification of community structures in the groups of layers.  
We do not study estimation of block probability matrices since those matrices are different in all layers 
and can be viewed as nuisance parameters. 

%

The rest of the paper is organized as follows. Section~\ref{sec:model}  introduces the DIMPLE model considered in this paper, presents notations, and reviews the existing results. Section~\ref{sec:methods} presents the algorithm for the between-layer clustering. Specifically,
it reviews the SSC methodology and explains why it is a good candidate for the job. 
Section~\ref{sec:theory} introduces assumptions and provides theoretical guarantees for the consistency of the between-layer
clustering. Section~\ref{sec:simulations}  contains a limited simulation study. 
Section~\ref{sec:RealData} supplements the paper with a real data example.
Section~\ref{sec:discussion} provides  concluding remarks. All the proofs are given in Appendix~\ref{sec:proofs}.


\section{The DIMPLE model}
\label{sec:model}

\subsection{Review of DIMPLE model}
\label{sec:dimple-review}

This section  reviews   the DIMPLE model introduced in \cite{pensky2021clustering}.
Consider an undirected multilayer network with $L$ layers over a common set of $n$ vertices with no self loops, where 
each of the layers follows  the SBM. Assume that those $L$  layers can be partitioned into
$M \ll L$ groups,  $\calS_1, \ldots, \calS_M$, where each group is equipped with its own community structure.
The latter means that there exists a clustering function $c : [L] \to [M]$ such that $c(l) = m$ if  the $l \in \calS_m$, 
$m=1, ...,M$, where $[N] = \{1,...,N\}$ for any positive integer $N$. 
Nodes in the layer $l \in \calS_m$ follow SBM with the $K_m$   communities $G_{m,1}, \ldots , G_{m,K_m}$, that persist in the layers 
of type $m$. Hence, for every $m \in [M]$, there exists a clustering function $z^{(m)}:[n] \to [K_m]$ with the 
corresponding clustering matrix $Z^{(m)} \in \{0,1\}^{n \times K_m}$, such that  $Z^{(m)}_{i,k}=1$ 
if and only if  $z^{(m)} (i) = k$.
Nonetheless, the block connectivity matrices $B^{(l)} \in [0,1]^{K_m \times K_m}$ can vary from layer to layer.
Therefore, the probability of connection between nodes $i$ and $j$ in layer $l$  is $P^{(l)}_{(i,j)} =  B^{(l)}_{k_1,k_2}$
where $k_1 = z^{(m)}(i)$ and $k_2 = z^{(m)}(j)$. In summary, while the membership function $z^{(m)}: [n] \to [K_m]$
is completely determined by the group $m$ of layers, the block connectivity matrices   $B^{(l)}$ are not, and can be all different   
in the group $m$ of layers. In this case, the matrix of connection probabilities in layer $l$ is of the form 
 \be \label{eq:model}
 P^{(l)} = Z^{(m)} B^{(l)} (Z^{(m)})^T, \quad m=c(l), \quad l = 1, \ldots, L 
\ee

 Furthermore,  we assume that   symmetric adjacency matrices $A^{(l)} \in \{0,1\}^{n\times n}$,
$l=1, ...,L$, are  such that  $A^{(l)}_{i,j}\sim \mbox{Bernoulli}(P^{(l)}_{i,j}),$  
 $1 \leq i < j \leq n,$ where  $A^{(l)}_{i,j}$  are conditionally independent 
given $P^{(l)}_{i,j}$, 
$A^{(l)}_{i,j} = A^{(l)}_{j,i}$  and  $A^{(l)}_{i,i}=0$. 
Denote the three-way tensors with   layers $A^{(l)}$ and $P^{(l)}$,  $l \in [L]$, by 
$\bA, \bP \in \RR^{ n \times n  \times L}$, respectively.

It is easy to see that, for $M=1$, the DIMPLE model reduces to the common multilayer network setting in, e.g.,  
 \cite{bhattacharyya2020general}, \cite{lei2021biasadjusted}, \cite{10.1093/biomet/asz068},
 \cite{paul2016}, \cite{paul2020}, where
the community structures persist throughout the network. On the other hand, it becomes the MMLSBM of 
\cite{fan2021alma}  and  \cite{TWIST-AOS2079} if the block connectivity matrices   $B^{(l)}$ are the same for 
all layers in a group, i.e., $B^{(l)} = B^{(c(l))}$, $l \in [L]$.

While the analysis of a multilayer network above can potentially involve three objectives: finding the partition function 
$c: [L] \to [M]$ for 
the layers of the network (between-layer clustering), finding community structures for each group of layers (within-layer clustering), 
and recovering block probability matrices $B^{(l)}$, $l \in [L]$, in this paper we pursue only the first  two  goals. 
Moreover, while we are using a novel Sparse Subspace Clustering based algorithm for the between layer clustering,
we utilize the within-layer clustering algorithm of \cite{pensky2021clustering}, which is inspired by 
 \cite{lei2021biasadjusted}. However, while the algorithms in the present paper and in \cite{pensky2021clustering}
are the same, the community detection error rates are smaller in the present paper, which is due to a more
accurate between-layer clustering. 
Finally, since block probability matrices  $B^{(l)}$  carry no information about the multilayer structure, they act  like 
a kind of nuisance parameters, and, therefore, are   of no interest. Moreover, if the need to retrieve them occurs, 
one can easily estimate them by averaging the entries of the adjacency matrix $A^{(l)}$ over the estimated community assignment.


\subsection{Notation}
\label{subsec:notation}

For any vector $ \bm v \in \RR^p$, denote  its $\ell_2$, $\ell_1$, $\ell_0$ and $\ell_\infty$ norms 
by $\|  \bm v\|$, $\| \bm v\|_1$,  $\| \bm v\|_0$ and $\| \bm v\|_\infty$, respectively. 
Denote by $\bm 1_m$  the $m$-dimensional column vector with all components equal to one. 
 
For any matrix $A$,  denote its spectral and Frobenius norms by, respectively,  $\|  A \|$ and $\|  A \|_F$. 
The column $j$ and the row $i$ of a matrix $A$ are denoted by $A(:, j)$ and $A(i, :)$, respectively.
Let $\vect(A)$ be the vector obtained from matrix $A$ by sequentially stacking its columns. 
Denote by $A \otimes B$ the Kronecker product of matrices $A$ and $B$. 
Denote the diagonal of a matrix $A$ by $\diag(A)$. Also, denote the $K$-dimensional 
diagonal matrix with $a_1,\ldots,a_K$ on the diagonal by $\diag(a_1,\ldots,a_K)$.

For any matrix $A \in \RR^{n \times m}$, denote its projection on the nearest rank $K$ matrix or its rank $K$ approximation by 
$\Pi_K(A)$, that is, if $\sigma_k$ are the singular values, and $\bm u_k$ and $\bm v_k $ are the left and the right singular 
vectors of $A$, $k = 1, \ldots, r$, then
\bes 
 A=\sum_{k=1}^{r} \sigma_k \bm u_k \bm v_k^T \quad \Rightarrow \quad \Pi_K(A)=\sum_{k=1}^{\min(r,K)} \sigma_k \bm u_k \bm v_k^T.
\ees
Denote
\be
\calO_{n,K} = \left \{A \in \RR^{n \times K} : A^T A = I_K \right \},   \quad \calO_n=\calO_{n,n}.
\ee
A matrix $X \in \{0,1\}^{n_1 \times n_2}$ is a clustering matrix  if it is binary and has exactly one 1 per row.  Also, we denote an absolute constant independent of $n, K, L$ and $M$, which can take different values at different instances, by $\bbC$.


\subsection{Review of the existing results }
\label{subsec:review_exist}
 
To the best of our knowledge, Pensky \& Wang (2021) \cite{pensky2021clustering}
is the only paper that studied the DIMPLE model. In particular, the authors there assumed that 
the number of communities in each group of layers is the same, i.e. $K_1=K_2=\ldots=K_M=K$. 
The motivation for this decision is the fact that the labels of the groups are interchangeable, 
so that, in the case of non-identical numbers of communities, it is hard to choose, 
which of the values correspond to which of the groups.

Pensky and Wang (2021) \cite{pensky2021clustering} used spectral clustering for estimation of the label function $c:[L] \to [M]$ and the corresponding clustering matrix $C$.

%
\begin{algorithm} [t] 
\caption{\ The between-layer clustering}
\label{alg:PW_between}
\begin{flushleft} 
{\bf Input:} Adjacency tensor $\bA \in \{0,1\}^{n \times n \times L}$, number of groups of layers $M$,  number of communities $K$. \\
{\bf Output:} Estimated clustering matrix $\hC \in \calM_{L,M}$. \\
{\bf Steps:}\\
{\bf 1:} For $l=1, ...,L$, find the  SVDs $A\upl  = \hU_{A,l} \hLam_{P,l} (\hU_{A,l})^T$, $\hU_{A,l} \in \calO_{n,K}$. \\
{\bf 2:} Form matrix $\hTe \in \RR^{n^2 \times L}$ with columns $\hTe(:,l) = \vect(\hU_{A,l} (\hU_{A,l})^T)$.\\
{\bf 3:} Construct the SVD   $\hTe = \wt \calV \wt \Lam  \wt \calW$,   $\wt \calV \in \calO_{n^2,L}$,  $\wt \calW \in \calO_L$,
 and obtain matrix $\hcalW = \wt \calW(:, 1:M) \in \calO_{L,M}$.\\
{\bf 4:}  Cluster  $L$ rows of  $\hcalW$ into $M$ clusters using   $(1+\eps)$-approximate $K$-means clustering. Obtain 
estimated clustering matrix $\hC$.   
\end{flushleft} 
\end{algorithm}
%


In order to find  the clustering matrix $C$,  \cite{pensky2021clustering} denoted  
$U_z\upm =  Z\upm (D_z\upm)^{-1/2}$, where  matrices $D_z\upm = (Z\upm)^T  Z\upm$   and 
 $U_z\upm \in \calO_{n,K}$, $m=1,...,M$. They observed that matrices $P\upl$ in \eqref{eq:model} 
can be written as 
\be  \label{eq:expans1}
P\upl =  U_z\upm  (D_z\upm)^{1/2} B\upl (D_z\upm)^{1/2} (U_z\upm)^T, \quad l=1,...,L.
\ee 
In order to extract common information from matrices  $P\upl$,  \cite{pensky2021clustering} considered the singular value decomposition (SVD) of  $P\upl$
\be \label{eq:svd1} 
P\upl = U_{P,l} \Lam_{P,l} (U_{P,l})^T, \quad U_{P,l} \in \calO_{n,K},\ l=1,...,L,
\ee 
and related it to expansion \eqref{eq:expans1}. If  
\be    \label{eq:BDl}
B_D\upl \equiv  (D_z\upm)^{1/2}  B\upl (D_z\upm)^{1/2} = O_z\upl S_z\upl (O_z\upl)^T
\ee  
are the corresponding SVDs of $B_D\upl$, where  $S_z\upl$ are 
$K$-dimensional diagonal matrices, and matrices $B\upl$ are of full rank, then 
$O_z\upl \in \calO_K$, so that $O_z\upl (O_z\upl)^T = (O_z\upl)^T O_z\upl = I_K$.
The latter leads to 
\be \label{eq:main_rel}
U_{P,l} (U_{P,l})^T =  U_z\upm O_z\upl  (O_z\upl)^T (U_z\upm)^T = U_z\upm (U_z\upm)^T, \quad m = c(l),
\ee
so that matrices $U_{P,l} (U_{P,l})^T$ depend  on $l$ only via $m = c(l)$ and are uniquely defined for $l=1,...,L$. 
Since matrices  $U_{P,l} (U_{P,l})^T$ are unavailable, in  \cite{pensky2021clustering}, 
they were replaced by their proxies $\hU_{A,l} (\hU_{A,l})^T$ obtained by the SVD  of layers  $A\upl$ of the tensor $\bA$.
The procedure is summarized in Algorithm~\ref{alg:PW_between}.


After the groups of layers are identified by Algorithm~\ref{alg:PW_between}, 
one can find the  communities by some kind of averaging.  Specifically,
\cite{pensky2021clustering} averaged the 
estimated version of the squares of the probability matrices $P\upl$,
similarly to \cite{lei2021biasadjusted}. 
Pensky and Wang   \cite{pensky2021clustering} introduced matrix $\hPsi$  of the form
\be   \label{eq:hUps}
\hPsi = \hC (\hD_{\hat{c}})^{-1/2} \in \calO_{L,M}, \quad \mbox{with} \quad  \hD_{\hat{c}} = \hC^T \hC.
\ee 
They constructed a tensor $\hbG \in \RR^{n \times n \times L}$ with layers 
$\hG\upl = \hbG(:,:,l)$ of the form
\be \label{eq:calhbG}
\hG\upl =   \lkr A\upl \rkr^2 - \diag(\hbd\upl), \quad l=1,...,L,
\ee
where $\hbd\upl$ is the vector of estimated nodes' degrees.
Subsequently, they averaged layers of the same types, obtaining tensor $\hbH \in \RR^{n \times n \times M}$,   
\be \label{eq:est_tensors_alt}
\hbH = \hbG \times_3 \hPsi^T, 
\ee 
where $\hPsi$ is defined in \eqref{eq:hUps}. They applied spectral clustering
to layers of tensor $\hbH$. 
The procedure follows \cite{lei2021biasadjusted} and is summarized in Algorithm~\ref{alg:PW_within}.

%
\begin{algorithm} [t] 
\caption{\ The within-layer clustering}
\label{alg:PW_within}
\begin{flushleft} 
{\bf Input:} Adjacency tensor $\bA \in \{0,1\}^{n \times n \times L}$, number of groups of layers $M$,  number of communities $K$,
estimated layer clustering matrix $\hC \in \calM_{L,M}$. \\
{\bf Output:} Estimated community assignments $\hZ\upm \in \calM_{n,K}$, $m=1,...,M$. \\
{\bf Steps:}\\
{\bf 1:} Construct tensor $\hbG$ with layers $\hG\upl =   \lkr  A\upl \rkr^2 - \diag(A\upl\, 1_n)$,  $l=1,...,L$.\\
{\bf 2:} Construct tensor $\hbH$ using formula $\hbH = \hbG \times_3 \hPsi^T$.  \\
{\bf 3:} Construct the SVDs of layers $\hH\upm =  \wt U_{\hH}\upm \hLam_{\hH}\upm (\wt U_{\hH}\upm)^T$,  
 $m=1, ...,M$.\\
{\bf 4:} Find $\hU_{\hH}\upm = \wt U_{\hH}\upm (:,1:K) = \Pi_K(\wt U_{\hH}\upm)$, 
 $m=1, ...,M$.\\
{\bf 5:}  Cluster    rows of  $\hU_{\hH}\upm$ into $K$ clusters using  $(1+\eps)$-approximate $K$-means clustering. Obtain 
clustering matrices $\hZ\upm$, $m=1,...,M$.   \\
\end{flushleft} 
\end{algorithm}
%


\section{Between-Layer  Clustering Procedure}
\label{sec:methods}


\subsection{Finding the matrix of weights}
\label{subsec:alg1}

In this paper, similarly to \cite{pensky2021clustering}, we assume that the number of communities in each group of layers is the same, 
i.e. $K_1=K_2=\ldots=K_M=K$.

 If one is unsure that each of the layers of the network has the same number of communities,
one can use a different number of communities $K\upl$ in each layer. After groups of layers are identified, the number of layers 
in each group should be re-adjusted, so that  $K^{(l)}= K_m$  if $m = c(l)$.
 One can, of course, assume that the values of $K_m$,
$m=1,...,M$, are known. However, since group labels are interchangeable, in the case of non-identical subspace dimensions 
(numbers of communities), it is hard to choose, which of the values correspond to which of the  groups.  
This is actually the reason why \cite{TWIST-AOS2079} and \cite{fan2021alma}, who imposed this assumption,
used it only in theory while their simulations and real data examples are all restricted to the case 
of equal $K_m$, $m=1,...,M$.  On the contrary, knowledge of $K\upl$ allows one to deal 
with different ambient dimensions (number of communities) in the groups of layers in simulations and real data examples.

In addition, for the purpose of methodological developments, we assume that the 
number of communities  $K$ in each layer of the network is known.  
Identifying the number of clusters is a common  issue in data clustering, and it is 
a separate problem from the process of actually solving the clustering problem
with a known number of clusters.  
A common method for finding the number of clusters is the so called ``elbow'' method 
that looks at the fraction of the variance 
explained as a function of the number of clusters. The method is based on the
idea that one should choose the smallest number of clusters, such that adding
another cluster does not significantly improve fitting of the data by a model.
There are many ways to determine the ``elbow''. For example, one can base its detection on 
evaluation of the clustering error in terms of an objective function, as in, e.g., \cite{Zhang2012}.
Another possibility is to  monitor  the eigenvalues of the non-backtracking matrix or the 
Bethe Hessian matrix, as it is done in~\cite{Le2015EstimatingTN}. 
One can also  employ a simple technique of checking the eigen-gaps, 
as it has been discussed in \cite{vonLuxburg2007}, 
or use a scree plot as it is done in \cite{ZHU2006918}. 
\\

In order to partition the layers of the network into groups with the distinct community structures,
note that 
\be  \label{eq:span1}
 \vect(P^{(l)}) = (Z^{(m)} \otimes  Z^{(m)}) \bm b^{(l)}, \quad \bm b^{(l)} = \vect(B^{(l)}), \quad m=c(l), \quad  l \in [L]
\ee 
Hence, for  $m=c(l)$, vectors $\vect(P^{(l)})$ belong to distinct subspaces  $\ovr \calS_m = \Span(Z^{(m)} \otimes  Z^{(m)})$.
Denote 
\begin{align*}
    D^{(m)} & = (Z^{(m)})^T (Z^{(m)}) = \diag(n_{1}^{(m)}, ..., n_{K}^{(m)}), \\  
    U^{(m)} & = Z^{(m)} (D^{(m)})^{-1/2},  
      \quad m \in [M], 
\end{align*}
and observe that $U^{(m)} \in \calO_{n,K}$. Therefore, \eqref{eq:span1} can be rewritten as 
\be \label{eq:span2}
\vect(P^{(l)}) = (U^{(m)} \otimes  U^{(m)}) \left(\sqrt{D^{(m)}} \otimes   \sqrt{D^{(m)}}\right) \, \bm b^{(l)}, 
\quad m=c(l), \  l \in [L],
\ee 
so that $\ovr \calS_m = \Span(U^{(m)} \otimes  U^{(m)})$. Equations \eqref{eq:span1}  and \eqref{eq:span2} confirm 
that vectors $\vect(P^{(l)})$ lie in distinct subspaces $\ovr \calS_m$ with $m = c(l)$ and, hence, possibly can be partitioned 
into groups using subspace clustering.

Yet, there is one potential complication in applying   subspace clustering to the problem above. 
Indeed, the subspace clustering works well when the subspaces do not intersect or have insignificant intersection. 
However, each of the subspaces $\ovr \calS_m$ includes $n^{-1}\, \bm 1_{n^2}$ as its main basis vector. 
The latter is likely to compromise the precision of   subspace clustering techniques. 
However, luckily, it  is relatively easy to remove this vector from all subspaces. 
Consider a projection matrix
\be \label{eq:scrP}
\mathscr{P}  =  n^{-1}\, \bm 1_n \bm 1_n^T, \quad \mathscr{P}^2 =\mathscr{P}   
\ee 
Then, for 
\begin{align}  \label{eq:tilPl}
\wt P^{(l)} & = (I - \mathscr{P})  P^{(l)} (I - \mathscr{P})  
= (I - \mathscr{P}) Z^{(m)} B^{(l)} (Z^{(m)})^T (I - \mathscr{P}), \\
%
\label{eq:tilUm}
\wt U^{(m)} & = (I - \mathscr{P}) U^{(m)} = (I - \mathscr{P}) Z^{(m)} (D^{(m)})^{-1/2}, \quad m = 1, \ldots, M,
\end{align} 
and $\bm b^{(l)}$ defined in \eqref{eq:span1},   one has, for $m = c(l)$
\be \label{eq:q_l}
\bm q^{(l)}= \vect(\wt P^{(l)})  = (\wt U^{(m)} \otimes \wt U^{(m)}) \tilde {\bm{b}}^{(l)}, \, \,  
\tilde {\bm{b}}^{(l)} = \left( \sqrt{D^{(m)}} \otimes \sqrt{D^{(m)}} \right) \bm b^{(l)}
\ee 
%
Consider subspaces $\calS_m = \Span(\wt U^{(m)} \otimes \wt U^{(m)})$ with dimension $(K-1)^2 = \rank (\wt U^{(m)} \otimes \wt U^{(m)})$.
In many scenarios, the new subspaces $\calS_m$
have very little or no intersection and, hence, can be well separated using the subspace clustering technique.

Subspace clustering has been widely used in computer vision and, for this reason, it is a very
well studied and developed methodology.
Subspace clustering is designed for separation of points that lie in the union of subspaces. 
Let $\bm x^{(j)} \in \R^{D}$, $j=1, \ldots,L$ be a given set of points drawn from an unknown union of 
$M \geq   1$ linear or affine subspaces $\calS_{i}$, $i=1,\ldots,M$, of unknown dimensions 
$d_{i}= \text{dim}(\calS_{i})$, $0<d_{i} <D$, $i=1,...,M$. In the case of linear subspaces, 
the subspaces can be described as 
$$
\calS_{i}=\lfi \bm{x} \in \R^{D} : \bm{x}=  \mathcal{U}^{(i)}\bm{f} \rfi, \quad i=1,...,M 
$$
where $\mathcal{U}^{(i)} \in \R^{D \times d_{i}}$ is a basis for subspace $\calS_{i}$ 
and $\bm{f} \in \R^{d_{i}}$ is a low-dimensional representation for point $\bm{x}$. 
The goal of subspace clustering is to find the number of subspaces $M$, their dimensions 
$d_{i}$,  $i=1,\ldots,M$, the subspace bases $\mathcal{U}^{(i)}$,  $i=1,\ldots,M$,  and the segmentation 
of the points according to the subspaces.

Several methods have been developed to implement subspace clustering such as  algebraic methods 
( \cite{vidal2005generalized}), 
iterative methods   (\cite{tseng2000nearest})
and spectral clustering based methods (\cite{Elhamifar:2013:SSC:2554063.2554078}, 
 \cite{soltanolkotabi2014},   \cite{vidal2011subspace}). 
In this paper, we shall use the latter group of techniques. 
Spectral clustering algorithms rely on  construction of  an affinity matrix 
whose entries are based on some distance measures between the points. 
For example, in the case of the SBM, adjacency matrix itself serves as the affinity matrix, 
while for the Degree Corrected Block Model (DCBM) (\cite{Karrer2011StochasticBA}), 
the affinity matrix is obtained by normalizing rows/columns of the adjacency matrix. 
In the case of the  subspace clustering problem, one cannot use the typical distance-based affinity measures
because two points could be very close to each other, but lie in different subspaces, while they 
could be far from each other, but lie in the same subspace. One of the solutions is to construct 
the affinity matrix using self-representation of the points, with the expectation that a point is more likely to 
be presented as a linear combination of points in its own subspace rather than from a different one.  
A number of approaches such as Low Rank Representation (\cite{Liu2010RobustSS}) and Sparse Subspace Clustering (SSC) 
(\cite{Vidal:2009aa} and \cite{Elhamifar:2013:SSC:2554063.2554078})
have been proposed   for the solution of this problem.

In this paper we  use  the self-representation version of  the SSC
developed in  \cite{Elhamifar:2013:SSC:2554063.2554078}. The technique
is based on   representation of each of the vectors as a sparse linear combination of
all other vectors. The weights obtained by this procedure are used to form the affinity matrix which, 
in turn, is partitioned using the spectral clustering methods.
If vectors $\bm{q}^{(l)},$ $l=1,\ldots,L$, in \eqref{eq:q_l} were known, the weight matrix $W$ would be based 
on writing every vector as a sparse linear combination of all other vectors by minimizing the number of nonzero coefficients 
\begin{equation}  \label{mn:opt_prob1}
 \min_{\bw^{(l)}} \left\|\bw^{(l)} \right\|_{0}    \quad   \mbox{s.t.}    \quad   \bm{q}^{(l)}=\sum_{k \ne l} W_{k,l} \bm{q}^{(k)}, 
\quad  \bw^{(l)}  = W(:,l)
\end{equation}
The affinity matrix of the SSC is the symmetrized version of the weight matrix $W$. 
%
Since the problem \eqref{mn:opt_prob1}  is NP-hard, one usually solves its convex relaxation,
with $\|\bw^{(l)}\|_{0}$ in \eqref{mn:opt_prob1} replaced by $\|\bw^{(l)}\|_{1}$.

In the case of the DIMPLE model, vectors $\bm{q}^{(l)}$, $l=1, ..., L$, are unavailable. 
Instead, we use their proxies based on the adjacency matrices. Specifically, we consider matrices 
\be \label{eq:AY}
\wh {\wt P}^{(l)} = \Pi_{K-1} (\wt A^{(l)}), \quad \wt A^{(l)} = (I - \mathscr{P})  A^{(l)} (I - \mathscr{P}),
\quad  \mathscr{P}  =  n^{-1}\, \bm 1_n \bm 1_n^T.
\ee
Here $\wh {\wt P}^{(l)}$ is the rank $(K-1)$ approximation of $\wt A^{(l)}$.
Construct matrices  $Y, \wh Q \in \RR^{n^2 \times L}$ with columns  $\by^{(l)}$ and 
$ \hat{\bq}^{(l)}$, respectively, given by
\be  \label{eq:matrY}
\by^{(l)} = Y(:,l) =   \hat{\bq}^{(l)}/\|\ \hat{\bq}^{(l)}\|, \quad 
 \hat{\bq}^{(l)} = \vect \lkr \widehat{\widetilde P}^{(l)}\rkr, \ \ l=1, \ldots, L
\ee
In the case of data contaminated by noise, the SSC algorithm does not attempt to write  each
${\bm{y}}^{(l)}$  as an exact linear combination of other points. Instead, the SSC is built 
upon   solutions of the LASSO problems  
\begin{equation} \label{mn:lasso}
 \widehat{\bw}^{(l)}  \in  \underset{\bm w \in \RR^{L}, \bm w_l=0}{\text{argmin}}   
\lfi  \left\| {\bm{y}}^{(l)} - Y \bm w \right\|^{2} + 2 \lam \left\|\bm w \right\|_{1}  \rfi, \ l=1,\ldots,L, 
\end{equation}
where $\lam > 0$ is the tuning parameter. 
We solve  \eqref{mn:lasso} using a fast version of the LARS algorithm 
implemented  in  SPAMS Matlab toolbox  \cite{mairal2014spams}.


{\spacingset{1.4}

\begin{algorithm}[t] 
\caption{\ Finding the matrix of weights}
%
\label{mn:SSC:alg}
\begin{flushleft} 
{\bf Input:} Tensor $\bA$; 
the number of communities $K$ in each layer; parameter $\lambda$.\\
{\bf Output:} 
matrix  $\wh{\wt W}$ of weights. \\
{\bf Steps:}\\
{\bf 1:} For $l=1, ...,L$, find pre-conditioned rank $(K-1)$ approximations $\wh {\wt P}^{(l)}$ of $A^{(l)} = \bA(:,:,l)$,
using formula \eqref{eq:AY}.\\
{\bf 2:}  Construct matrix $Y \in \RR^{n^2 \times L}$ with columns ${\bm{y}}^{(l)}$, $l=1,\ldots,L$, defined in \eqref{eq:matrY}.  \\
{\bf 3:} Find a matrix of weights, $\widehat W \in \RR^{L \times L}$ with columns 
$\widehat{\bw}^{(l)}  = \widehat{W} (:,l)$
and $\diag(\widehat{W})=0$, by solving the LASSO problem \eqref{mn:lasso} for  $l=1,\ldots,L$.\\
{\bf 4:} Construct matrix $\wh{\wt W} = |\wh W| + |\wh W^T| $ of weights.
%
\end{flushleft} 
\end{algorithm}

}


Given $\widehat{W}$, the clustering function  $\hat{c}: [L] \to [M]$ is  obtained by applying spectral clustering 
to the affinity matrix   $|\widehat{W}| + |\widehat{W}^{T}|$, 
where, for any matrix $B$, matrix $|B|$ has absolute values of elements of $B$ as its entries.
Algorithm~\ref{mn:SSC:alg} summarizes the methodology described above.


{\spacingset{1.4}

\begin{algorithm}[t]   
\caption{\ The between-layer clustering}
\label{alg:SSC:between-clust}
\begin{flushleft} 
{\bf Input:}  Matrix  $\wh{\wt W} \in \RR^{L \times L}$ of weights; the number of communities $K$ in each layer; the number of groups of layers $M$,
threshold $T$.\\
{\bf Output:} The clustering function $\hat c: [L] \rightarrow [M]$ and the corresponding clustering matrix $\wh C$.\\
%
{\bf Steps:}\\
{\bf 1:} Find  $\hat {\tilde{c}}: [L] \to [M]$ by applying spectral clustering to $\wh{\wt W}$.
Find the corresponding clustering matrix $\wh {\wt C}$.\\
{\bf 2:} Find  $\wh \calD = \diag(\wh{\wt W} \bm 1)$ and the Laplacian $\calL = \wh \calD - \wh{\wt W}$. 
Find $\wt M$, the number of disconnected components of $\calL$ and the clustering function  $\phi : [L] \to [\wt M]$.\\
{\bf 3:} If $\wt M \leq M$, then $\hat c = \hat {\tilde{c}}$ and $\wh C = \wh {\wt C}$.\\
{\bf 4:} If  $\wt M > M$, then construct matrix $\wh \Up \in \RR^{L \times L}$ with elements 
$\wh \Up_{l_1, l_2}= | (\bm y^{(l_1)})^T \bm y^{(l_2)}|$, where $l_1, l_2 = 1, \ldots, L$, 
and ${\bm{y}}^{(l)} = Y(:, l)$ are defined in \eqref{eq:matrY}. \\
{\bf 5:} Let $\Phi  \in \{ 0, 1 \}^{L \times \wt M}$ be the clustering matrix corresponding to the clustering function $\phi$.
Let $D_{\Phi} =  \Phi^T \Phi$.   Construct matrix 
$\wh{\wt \Up} = (D_{\Phi})^{-1/2}\, \Phi^T \, \wh \Up \, \Phi \, (D_{\Phi})^{-1/2} \in \RR^{\wt M \times \wt M}$
and its thresholded version  $\wh G \in \{ 0, 1 \}^{\wt M \times \wt M}$ with elements 
$\wh G_{\wt m_1, \wt m_2} = I(\wh {\wt \Up}_{\wt m_1, \wt m_2} > T)$, $\wt m_1, \wt m_2= 1, \ldots, \wt M$.\\
{\bf 6:} Find the SVD $\wh G= U_{\wh G} \, \Lambda_{\wh G} \, (U_{\wh G})^T$ of $\wh G$, and cluster rows of $U_{\wh G}(:, 1: M)$ into $M$ clusters. 
Obtain  clustering function $\theta : [\wt M] \rightarrow [M]$ and the corresponding clustering matrix $\Theta$.\\
{\bf 7:}  Set $\wh C = \Phi \Theta$ and $\hat c (l) = \theta (\phi(l)), \, l= 1, \ldots, L$, superposition of $\theta$ and $\phi$.
\end{flushleft} 
\end{algorithm}
}


\subsection{Between-layer clustering}
\label{sec:alg2}

As a result of Algorithms~\ref{mn:SSC:alg}, one obtains a matrix $\wh{\wt W} = |\wh W| + |\wh W^T| $ of weights.
Then, one can apply spectral clustering to $\wh{\wt W}$, partitioning $L$ layers into $M$ clusters. 

The success of clustering relies on the fact that the weight matrix $\widehat{W}$ is such that 
$\widehat{W}_{k,l}\neq 0$ only if points $k$ and $l$ lie in the same subspace, which guarantees that 
vectors $\bm y^{(l)}$ are represented by vectors in their own cluster only. This notion is formalized as 
the Self-Expressiveness Property.
Specifically, we say that the weight matrix 
$W \in \RR^{L \times L}$ satisfies the {\it Self-Expressiveness Property} (SEP)  if $|W (i,j)| > 0$ implies $c(i)=c(j)$,
where $c: [L] \to [M]$ is the true clustering function. Hence, for the success of clustering, 
we would like to ensure that matrix $\wh W$ with columns 
$ \widehat{\bw}^{(l)}$, $l=1, \ldots, L$, defined in \eqref{mn:lasso}, satisfies the SEP with high probability.
Indeed, if  SEP  holds, then no two layer networks  from different groups of layers can have a nonzero weight in the matrix  
$\wh{\wt W}$. 

However, it is known that SEP alone does not guarantee perfect clustering since 
the similarity graph obtained on the basis of $\wh{\wt W}$ can be poorly connected (see, e.g.,  \cite{5995679}). 
Indeed, if the similarity graph has $\wt M  > M$  disconnected components, then one would obtain
spurious clustering errors due to the incorrect grouping of those components. 
It is possible to have $\wt M > M$ since, within one subspace, one can have a group of vectors 
that can be expressed as weighted sums of each other.  The connectivity 
issue has been addressed in, e.g., \cite{pmlr-v51-wang16b}, where the authors proved that the SSC
achieves correct  clustering with high probability under the restricted eigenvalue assumption.
They propose an innovative algorithm for merging subspaces by using single linkage clustering of the 
 disconnected components.
Since we cannot guarantee that the restricted eigenvalue assumption holds in our case, we suggest    
a different novel  methodology for clustering the disconnected components  into $M$ clusters. 
The method is summarized in Algorithm~\ref{alg:SSC:between-clust}. Algorithm~\ref{alg:SSC:between-clust} requires milder conditions and is easier to implement than the 
respective technique in \cite{5995679}.


\section{Theoretical guarantees}
\label{sec:theory}

\subsection{Assumptions }
\label{sec:assump}

In this paper, we assume  that a DIMPLE network is generated by randomly sampling the nodes
similarly to how this is done in SBM models \cite{doi:10.1073/pnas.0907096106}, \cite{10.1214/13-AOS1124}.  Consider vectors 
$\ovr {{\bs \varpi}} = (\varpi_1, ..., \varpi_M) \in [0,1]^M$ and 
$\ovr {\bs{\pi}}^{(m)}  = (\pi_1^{(m)}, ..., \pi_K^{(m)})  \in [0,1]^K$, $m \in [M]$,
such that 
\bes
\sum_{m=1}^{M} \varpi_m = 1, \quad \sum_{k=1}^{K} \pi_k^{(m)}  = 1, \quad m \in [M].
\ees
For each layer $l \in [L]$, we generate its group membership $c(l) \sim {\rm Multinomial}(\ovr {\bs{\varpi}})$.
For each node $j  \in [n]$ in a layer of type $m \in [M]$,
the  membership function $z^{(m)}$  is generated as $z^{(m)}(j)  \sim {\rm Multinomial}(\ovr {\bs{\pi}}^{(m)})$.
Hence, $\varpi_m$ is the probability of a layer of type $m$, and 
 $\pi_k^{(m)}$, $k= 1,...,K$, is the   probability of the $k$-th community in a layer of type $m$.

While, in general, the values of $\pi_k^{(m)}$ can be different for different   $m$, 
in this paper, we assume that $\pi_k^{(m)} = \pi_k$, $m \in [M]$, $k \in [K]$. The latter means that 
for a node $j$ in a group of layers $m$, its community membership can be generated as 
\begin{align} \label{eq:rand_gen1}
& \xi_j^{(m)}  \sim \text{Multinomial}(\bar {\bs{\pi}} , K)  \quad \mbox{with} \quad  
\bar {\bs{\pi}} = (\pi_1, \ldots, \pi_K), \quad j  \in [n] \\
\label{eq:rand_gen2}
& Z_{j,k}^{(m)} = I (\xi_j^{(m)} = k),\quad \PP(Z_{j,k}^{(m)} =1) = \pi_k, 
 \quad k \in [K], \ j  \in [n], \ m \in [M]
\end{align}
After layers' memberships and nodes' memberships in groups of layers are generated, 
the set of matrices $B^{(l)}$ is chosen independently from the groups of layers and community assignments.

In order to derive theoretical guarantees for the SEP, one needs to impose conditions that 
ensure  that the layer networks maintain some regularity and are not too sparse. We   also need to ensure that 
the subspaces, that represent the layer networks, are sufficiently separated, and   are also well represented by the 
sets of vectors $\bm q^{(l)}$ with $c(l)=m$, where $\bm q^{(l)}$ are defined in \eqref{eq:q_l}.  
For this purpose, we introduce matrices $Q, X \in   \RR^{n^2 \times L}$ with columns $\bm q^{(l)}$ and $\bm x^{(l)}$, respectively, where
\be  \label{eq:matrX}
\bm x^{(l)} = X(:,l) = \bm q^{(l)} / \|\bm q^{(l)}\|, \quad \bm q^{(l)} = \vect ( \widetilde P^{(l)}), \ \ 
l=1,...,L, 
\ee
Matrix $X$ can be viewed as the ``true'' version of matrix $Y$ in \eqref{eq:matrY}. 
We impose the following assumptions:

\noindent{\bf A1.} For some positive constants $\underline C$ and $\bar C$, $0 < \underline C \le \bar C < \infty$ , one has
\be \label{eq:Bol}
B^{(l)} = \rho_n B^{(l)}_0 \quad \mbox{with} \quad 
\underline C \le \| B^{(l)}_0 \| \le \bar C 
\ee

\noindent{\bf A2.} For some positive constant  $C_{\sig,0}$, one has
\be \label{eq:lammin_Bol}
\min_{l=1,...,L} \    \sig_{\min}(B^{(l)}_0)/\sig_{\max}(B^{(l)}_0)  \ge C_{\sig,0}  
\ee

\noindent{\bf A3.} For some positive constant  $C_{\rho}$, one has
\bes
\rho_n \ge C_\rho n^{-1} \log n
\ees

\noindent{\bf A4.} For some positive constants $\lowc_{\varpi}$, $\highc_{\varpi}$, $\lowc_{\pi}$ and $\highc_{\pi}$, one has
\be \label{eq:bounds} 
\lowc_{\varpi}/M \leq \varpi_m \leq \highc_{\varpi}/M; \quad   
\lowc_{\pi}/K \leq   \pi_k \leq \highc_{\pi}/K; \quad   k \in [K], \ m \in [M]
\ee

\noindent{\bf A5.} Matrices $B^{(l)}$ are such that, for   any $l \in [L]$ with $c(l)=m$, 
there exists representation $\bm x = \wt X_* \bm w_*$ of $\bm x \equiv \bx^{(l)}$ 
via other columns of $X$ in $\calS_m$, such that $\| \bw_* \|_1 \leq \aleph_{w,K}$ where $\aleph_{w,K}$ can only depend on $K$.  
\\

Assumptions {\bf A1}-{\bf A4} are common regularity assumptions for network papers. 
Since majority of networks are sparse, Assumption {\bf A1} introduces a sparsity factor $\rho_n$ and confirms that all matrices 
$B^{(l)}$ maintain approximately the same level of sparsity.  Assumption {\bf A2} requires that all matrices $B^{(l)}_0$, 
$l=1, ...,L$, are well conditioned. Assumption  {\bf A3}   guarantees that the eigenvectors  of the subspaces constructed on the basis 
of the adjacency matrices are close to those that are defined by  the matrices of probabilities of connections. 
Assumption {\bf A4} ensures that groups of layers in the network, as well as communities in each
of the groups, are balanced, i.e., the number of members have the same order of magnitude  when $n$ and $L$ grow. 
Denote 
\be \label{eq:hnk_hLm}
\widehat{L}_m =  \sum_{l=1}^M I(c(l)=m), \quad \hat n_k^{(m)} = \sum_{j=1}^{n} I(\xi_j^{(m)} = k), \quad k \in [K], \ m \in [M] 
\ee 
Then,   it turns out that, under  Assumption {\bf A4}, there is a set $\Om_t$ such that, for $\om \in \Om_t$ 
\be \label{eq:bal_cond}
\min_m  \hL_m  \ge  C_0\, L/M,  \quad  
  \wt C_0\, n/K \le \hn_{k}^{(m)}  \le \wt{ \wt C}_0\,  n/K, \quad m \in [M], \, \, k \in [K] 
\ee
It follows from Lemma~\ref{lem:balanced_groups} in Section~\ref{sec:th1_proof} that, if $L$ and $n$ are sufficiently large, 
\eqref{eq:bal_cond} holds    with 
\be \label{eq:balance_const}
C_0 = \lowc_{\varpi}/2, \quad \wt C_0 = \lowc_{\pi}/2,\quad  \wt{ \wt C}_0 = 3 \highc_{\pi}/2
\ee
on a set $\Omega_t$ with $\PP(\Om_t) \geq 1- 2 L^{-t} - 2 K\, n^{-t}$.
It turns out that Assumption {\bf A4}  also ensures that groups of layers of the  network are well separated.

Assumption~{\bf A5} replaces much more stringent conditions, which are present in majority of papers that provide theoretical guarantees 
for the sparse subspace clustering, specifically, the assumption of sufficient sampling density and spherical symmetry of the residuals.
While neither of these above conditions holds in our setting, Assumption~{\bf A5} is much easier to satisfy.
It actually requires that the {\it low-dimensional} vectors $\bm b^{(l_0)}$ are easily represented by other vectors 
$\bm b^{(l)}$, where $c(l) = c(l_0)$ and $l \neq l_0$.  Assumption~{\bf A5}  is valid under  
a variety of sufficient conditions. Some examples of those conditions are presented in the following lemma.

\begin{lemma} \label{lem:condA5_a}
(a)\ Consider vectors $\bb_0^{(l)} = \vect(B_0^{(l)})$  where matrices $B_0^{(l)}$ are defined in \eqref{eq:Bol}.
Let, for  any $m \in [M]$ and any $l_0$ with $c(l_0)=m$, there exist a set of indices ${\cal L}_0$ such that
$l_0 \not\in {\cal L}_0$ and $c(l) =m$ for $l \in {\cal L}_0$, and matrix  $\calB_0$ with columns $\bb_0^{(l)}$,  
$l \in {\cal L}_0$, is a full-rank matrix with the lowest singular value $\sigma_{min} (\calB_0) \ge \sig_{0,K}$, 
where $\sig_{0,K}$ can only depend on $K$. Then,  Assumption {\bf A5}  holds with 
\bes
 \aleph_{w,K} = \frac{(\bar C)^2\, \wt{ \wt C}_0}{\underline C\, C_{\sig,0}\, \wt C_0}\, \frac{K \sqrt{K}}{\sig_{0,K}}
\ees
(b)\ If, for $m \in [M]$,  matrices $B^{(l)}$ with $c(l)=m$ take  only  $M_m$ distinct values, with at least two matrices  $B^{(l)}$ 
taking identical values, then  Assumption~{\bf A5}  holds with $\aleph_{w,K} = 1$.
\end{lemma}

Note that part (a) of Lemma~\ref{lem:condA5_a} just prevents the situation where all but one of the vectors $\bb^{(l)}$
are positioned in close proximity of one another. 
Part (b) of Lemma~\ref{lem:condA5_a} includes  the   MMLSBM as its particular case, 
which means that our theoretical results also hold for the MMLSBM.


\subsection{Between-layer clustering precision guarantees }
\label{sec:clust_precision}

The success of clustering relies on the fact that the weight matrix $\widehat{W}$ 
with columns $ \widehat{\bw}^{(l)}$, $l=1, \ldots, L$, defined in \eqref{mn:lasso}, satisfies the SEP with high probability.
It turns out that Assumption {\bf A3}   ensures that subspaces  $\calS_m$, $m \in [M]$, 
corresponding to different types of layers, do not have large intersections  
and allow sparse representation of vectors within each subspace.
The following statement guarantees that this is true for  the weight matrix $\wh W$ in Algorithm~\ref{mn:SSC:alg}.


\begin{theorem} \label{th:self_express}
Let Assumptions  {\bf A1}-{\bf A5} hold and $t > 0$. Define
\be \label{eq:delnk}
\delta_{n,K,t}  = C_{t, \del}\,  K\, (n \rho_n)^{-1/2},
\ee 
where  $C_{t, \del}$ is a constant that depends only on $t$ and constants in Assumptions {\bf A1}-{\bf A4}. 
Let $\wh W$ be a solution of problem \eqref{mn:lasso} with $\lambda = \lam_{n,K}$ such that 
\be \label{eq:param_cond_th1}
\lam_{n,K} \leq (4\, \aleph_{w,K})^{-1}, \quad 
\lim_{n \to \infty} \frac{ \delta_{n,K,t} \  \aleph_{w,K} }{\lam_{n,K}} = 0 ,
\ee
where $\aleph_{w,K}$ is defined in Assumption~{\bf A5}.
If $n$ is large enough and $t>0$ satisfies
\be \label{eq:tcond}
t  < \min \lkr \lowc_{\varpi}^2\,  L\, (2 M^2 \log L)^{-1},\quad  \lowc_{\pi}^2\,  n\, (2 K^2 \log n)^{-1} \rkr,
\ee 
then  matrix $\wh W$ (and, consequently, $\wh{\wt W}$)  satisfies the SEP
on a set $\Omega_t$ with  
\be \label{eq:Omt}
\PP(\Om_t) \ge 1 - 2\, L^{-t}- L \, n^{-t} - 2\, K M (M+2) n^{-t}.
\ee 
%
\end{theorem}

\medskip

\ignore{
\begin{remark}  \label{rem1}
{\rm 
{\bf CHECK THIS!!}\\
Theorem~\ref{th:self_express} requires that $n$ is large enough. 
Specifically, one needs 
\bes
\lambda_{n,K} - \delta_{n,K,t}   -  2\,(1 + \aleph_{w,K}) 
\left ( 1 +    \delta_{n,K,t}^2 (1 + \aleph_{w,K})/(2\, \lambda_{n,K})  \right) (2 \delta_{n,K,t} + 2 \tau  + \delta_{n,K,t}^2) >0,
\ees
which is guaranteed by conditions \eqref{eq:param_cond}.
} 
\end{remark}
} 
 

We would like to point out the fact that although the statement in Theorem~\ref{th:self_express} is
relatively standard, its proof follows completely different path than  proofs of SEP known to us.
Indeed, those proofs   (see, e.g.,  \cite{soltanolkotabi2012},  \cite{soltanolkotabi2014}, \cite{10.5555/2946645.2946657})  
are tailored to the case of Gaussian errors and are based on the idea that the errors are rotationally invariant.
In addition, those proofs require  that the sampled vectors uniformly cover each of the subspaces. 
It is easy to observe that rotational invariance fails in the case of the Bernoulli random vectors,
so our proof is totally original. Moreover, we do not require the sampling condition as in, e.g.,  \cite{soltanolkotabi2014}
and \cite{10.5555/2946645.2946657}. Observe that condition {\bf A5} does  not require uniform sampling or sufficient sampling density.
Instead, condition~{\bf  A5} guarantees that each vector has a sparse representation via the vectors in the same subspace.

The following theorem states that, if the threshold $T = T_{n,K}$ in  Algorithm~\ref{alg:SSC:between-clust} satisfies certain conditions,
$n$ is large enough and the SEP holds, then Algorithm~\ref{alg:SSC:between-clust} leads  to perfect recovery of clusters with high probability.
The latter  implies that our clustering procedure is strongly consistent.


\begin{theorem} \label{th:SSC_ClustErrors}
Let Assumptions {\bf A1} - {\bf A5} hold and the clustering function $\hat c: [L] \rightarrow [M]$ be obtained by 
Algorithm~\ref{alg:SSC:between-clust}. Let 
$T \equiv T_{n,K}$ be such that 
\be  \label{eq:thresh_cond_new}
\lim_{n \to \infty} T_{n,K} =0; \quad  \lim_{n \to \infty}   \left ( \frac{K^2\, \log n}{T_{n,K}\, n}  
+ \frac{K }{T_{n,K}\,  \sqrt{n\, \rho_n}} \right )=0.
\ee
If $n$ is large enough and $t>0$ satisfies \eqref{eq:tcond}, then, 
up to permutation of $M$ cluster labels, for $\om \in \Om_t$ defined in \eqref{eq:Omt}, one has $\PP  (\hat c = c)$,
i.e., the clustering procedure is strongly consistent. 
\end{theorem}


Note that Algorithm~\ref{alg:SSC:between-clust} is very different from Algorithm~2 of \cite{pmlr-v51-wang16b} 
which relies on subspaces recovery and merging. Also, Theorem~\ref{th:SSC_ClustErrors} above holds under milder 
and more intuitive assumptions than Theorem~3.2 of  \cite{pmlr-v51-wang16b}.  
In conclusion, Theorem~\ref{th:SSC_ClustErrors} establishes strong consistency of SSC for data that is not rotationally invariant.


\subsection{Within-layer clustering precision   guarantees }
\label{sec:within_precision}

After the between-layer clustering has been accomplished, the within layer clustering can be carried out by 
Algorithm~\ref{alg:PW_within} of \cite{pensky2021clustering}. 

 Since the clustering is unique only up to a permutation of clusters, 
denote the set of $K$-dimensional permutation functions of $[K]$ by $\aleph(K)$ and the set of $K \times K$ permutation matrices by 
$\mathfrak{F} (K)$. 
The   local community detection error  in the layer of type $m$ is then given by 
\be \label{eq:err_wihin_def}
R_{WL} (m) = (2n)^{-1} \ \min_{\scrP_m \in \mathfrak{F}  (K)}\  \|\wh Z\upm  - Z\upm \, \scrP_m \|^2_F,\ \quad m=1, ...,M,
\ee
where $Z\upm$ is defined in \eqref{eq:model}.
Note that, since the numbering of layers is defined also up to a permutation,
the errors $R_{WL} (1)$, ..., $R_{WL} (M)$ should be minimized over the set of permutations $\aleph(M)$.
The average  error rate of the within-layer clustering is then given by
\begin{align} \label{eq:err_within_ave}
R_{WL}  & =   \frac{1}{M} \ \min_{\aleph(M)}\ \sum_{m=1}^M R_{WL} (m)\\
        & = \frac{1}{2\, M\,n} \ \min_{\aleph(M)}\ 
\sum_{m=1}^M  \lkv \min_{\scrP_m \in \mathfrak{F}  (K)}\  \|\wh Z\upm  - Z\upm \, \scrP_m \|^2_F \rkv \nonumber
\end{align}
With these definitions, one obtains the following statement. 
\\

\begin{theorem} \label{th:SSC_WithinErrors} 
Let Assumptions {\bf A1} - {\bf A5} hold and the between-layer clustering function 
$\hat c: [L] \rightarrow [M]$ be obtained by using 
Algorithm~\ref{alg:SSC:between-clust}. Let 
$T \equiv T_{n,K}$ satisfy condition \eqref{eq:thresh_cond_new} and $t>0$ obeys \eqref{eq:tcond}.
Then, for $n$ large enough, there exists a set $\wt \Omega_t$ and an absolute positive constant $C_t$ such that 
\be \label{eq:tilOm}
\PP(\wt \Omega_t) \geq  1 - 2L^{-t} - C_t (K M^2 + L + n^2)\,  n^{1-t} 
\ee 
and, for any $\om \in \wt \Omega_t$,  the 
average within-layer clustering error  $R_{WL}$   satisfies
\be   \label{eq:within_er}
R_{WL} \leq C_t  \, \lkv \frac{M  K^4 \log(L+n)}{L n \rho_n}\   + \frac{K^4}{n^2}  \rkv.
\ee 
 \end{theorem}



\section{Simulations}
\label{sec:simulations} 
 
{\spacingset{1.4}
 
\begin{figure}[t]
\[ \includegraphics[width=2.47in] {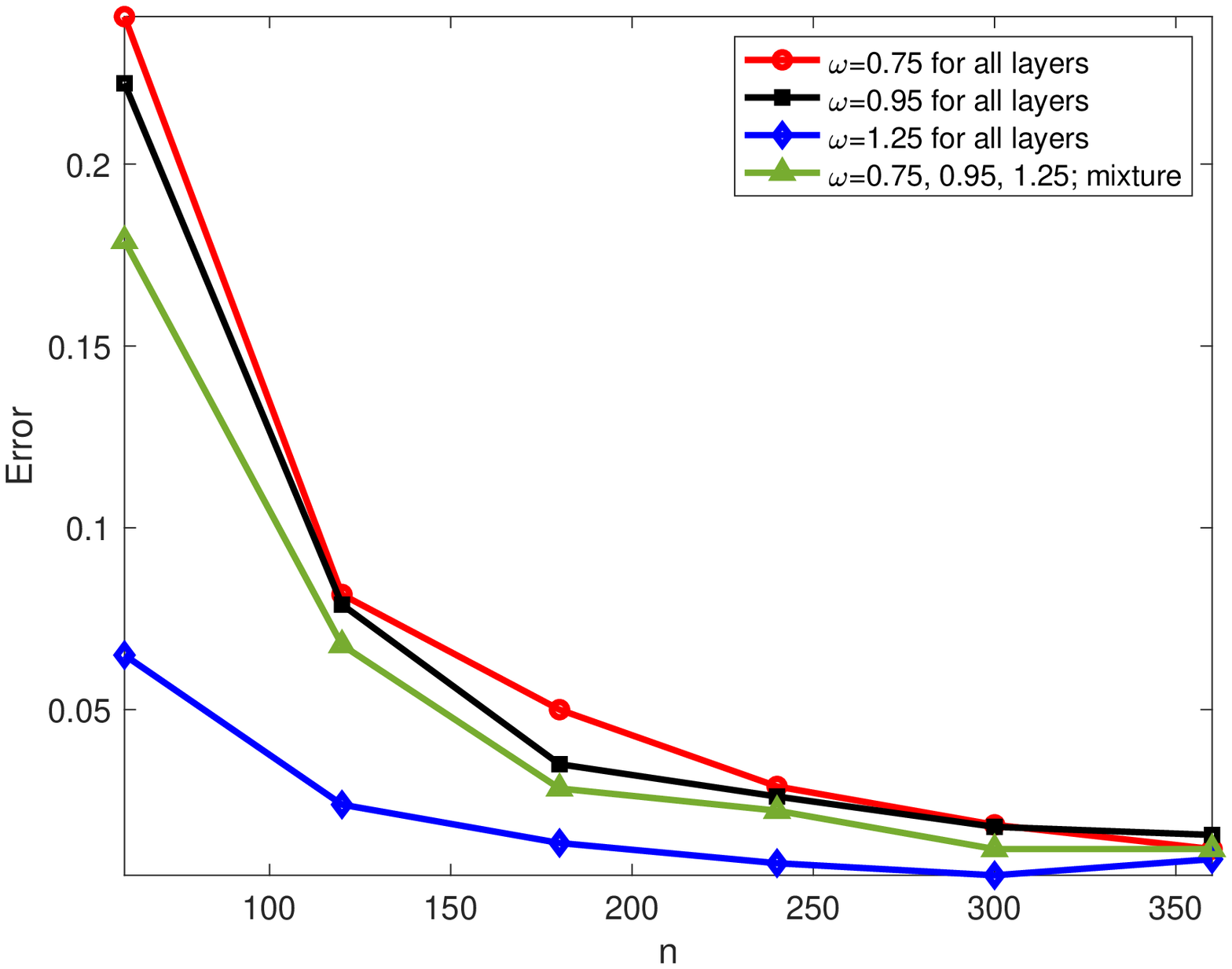}  
\includegraphics[width=2.47in]{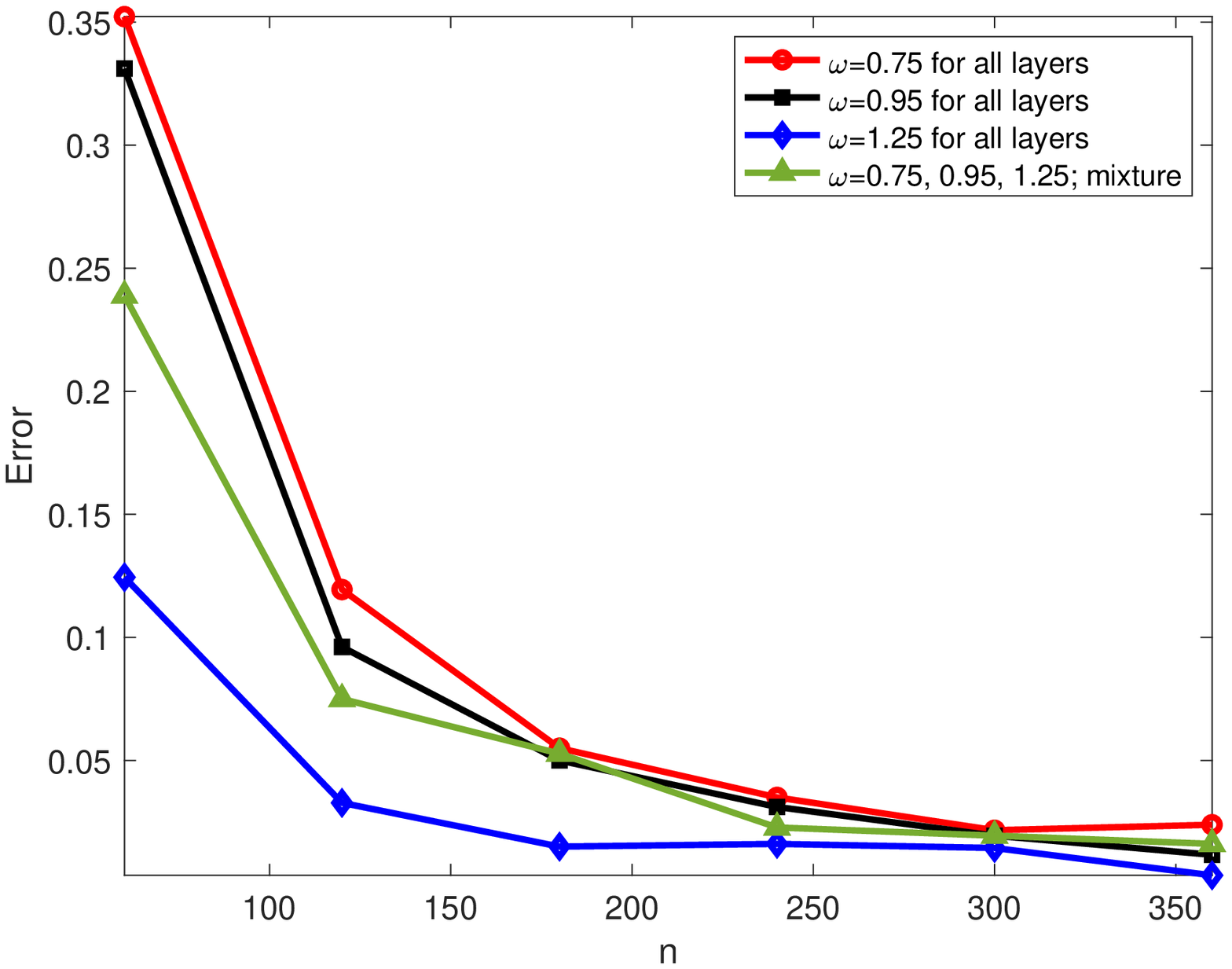}
\] 
\[ \includegraphics[width=2.47in] {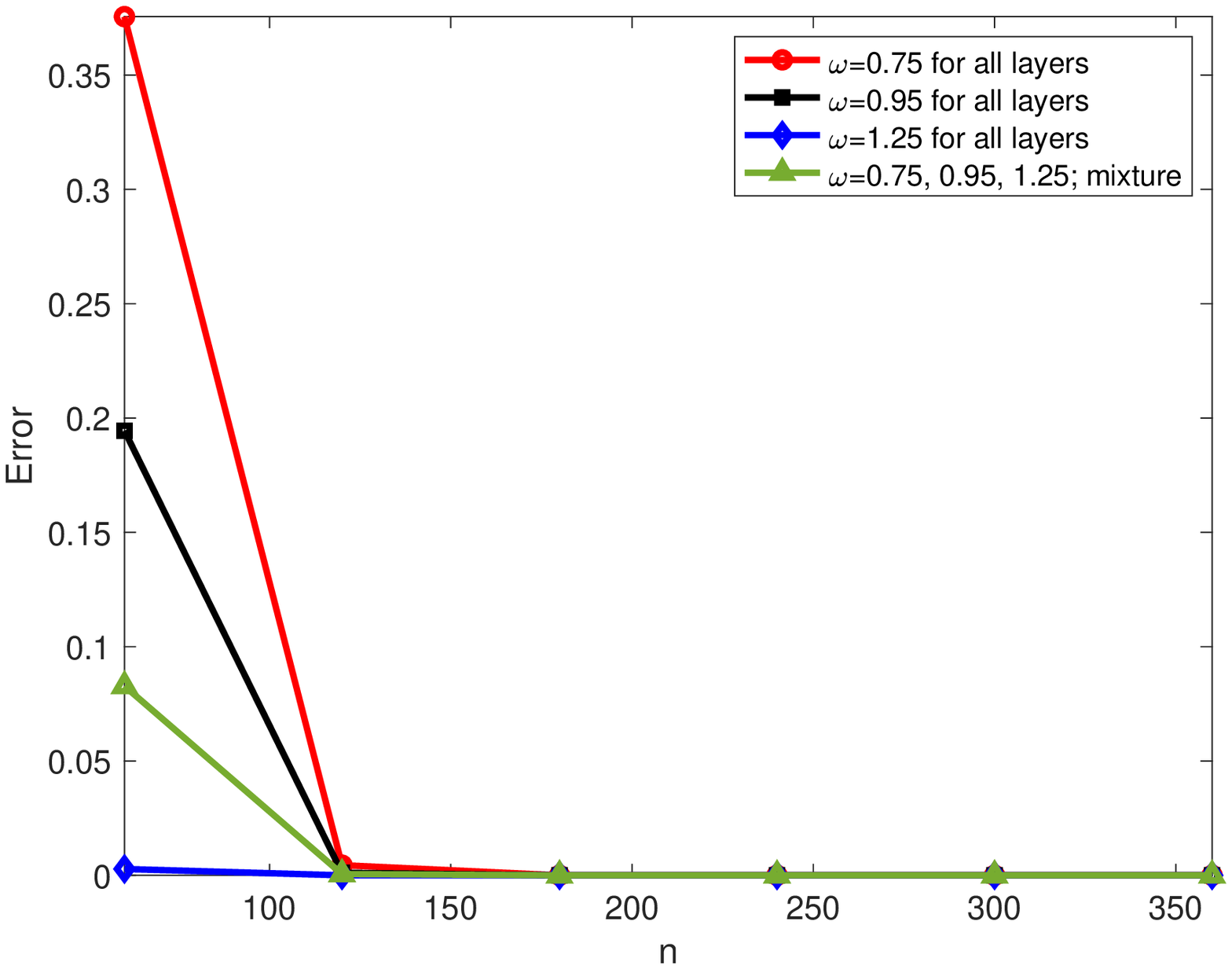} 
\includegraphics[width=2.47in]{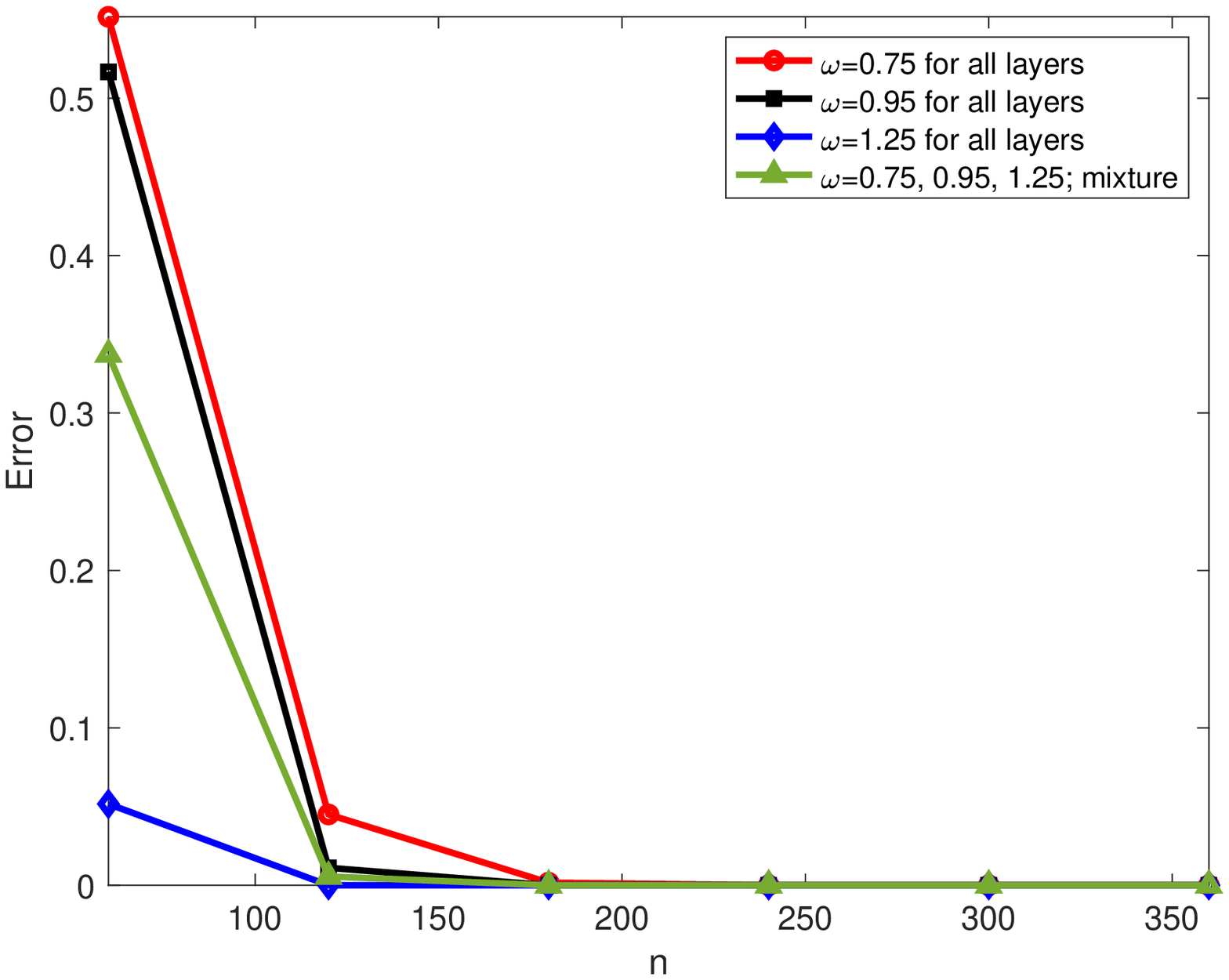}
\] 
\caption{ The clustering errors for $K=3$, $M=3$ (top left), $K=3$, $M=6$ (top right), $K=6$, $M=3$ (bottom left), 
and $K=6$, $M=6$ (bottom right). The number of nodes ranges from $n=60$ to $n=360$ with the increments of 60. 
The number of layers, $L=60$, is fixed for all panels. The lines with different colors represent the results for different values of $\om$: 
$\om=0.75$  (red); $\om=0.95$  (black); $\om=1.25$  (blue); $\om=0.75, 0.95, 1.25$  (green). 
The errors are evaluated over 100 simulation runs.  
}
\label{mn:fig1}
\end{figure}

}

In this section, we carry out a limited  simulation study to illustrate the performance  of our clustering method 
for finite values of $n$ and $L$.
To this end, we investigate the effect of various combinations of model parameters on the clustering errors obtained by our algorithms. 
The proportion of misclassified layers (the between-layer clustering error) is  evaluated as 
 \be \label{eq:misclustered}
\Err(C, \hC) =  (2L)^{-1}\, \underset{\mathscr{P}_M \in \mathfrak{F} (M)} {\min}  \| \hC - C \mathscr{P}_M   \| _F^2  
 \ee 
where $C, \hC  \in \{0,1\}^{L \times M}$ are, respectively,  the true and the estimated clustering matrices. 

{\spacingset{1.4}

\begin{figure}[t]
\[ \includegraphics[width=2.47in] {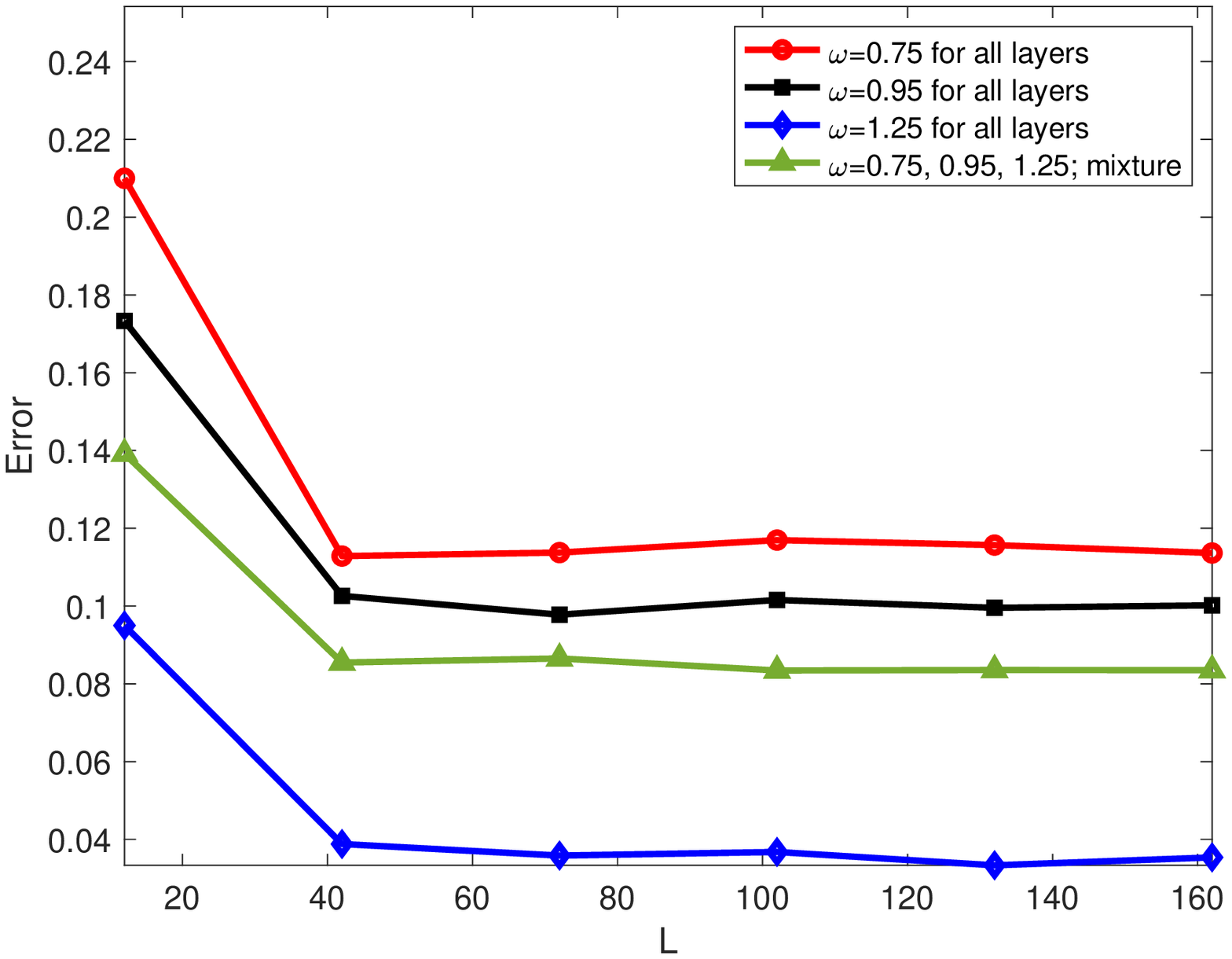}  
\includegraphics[width=2.47in]{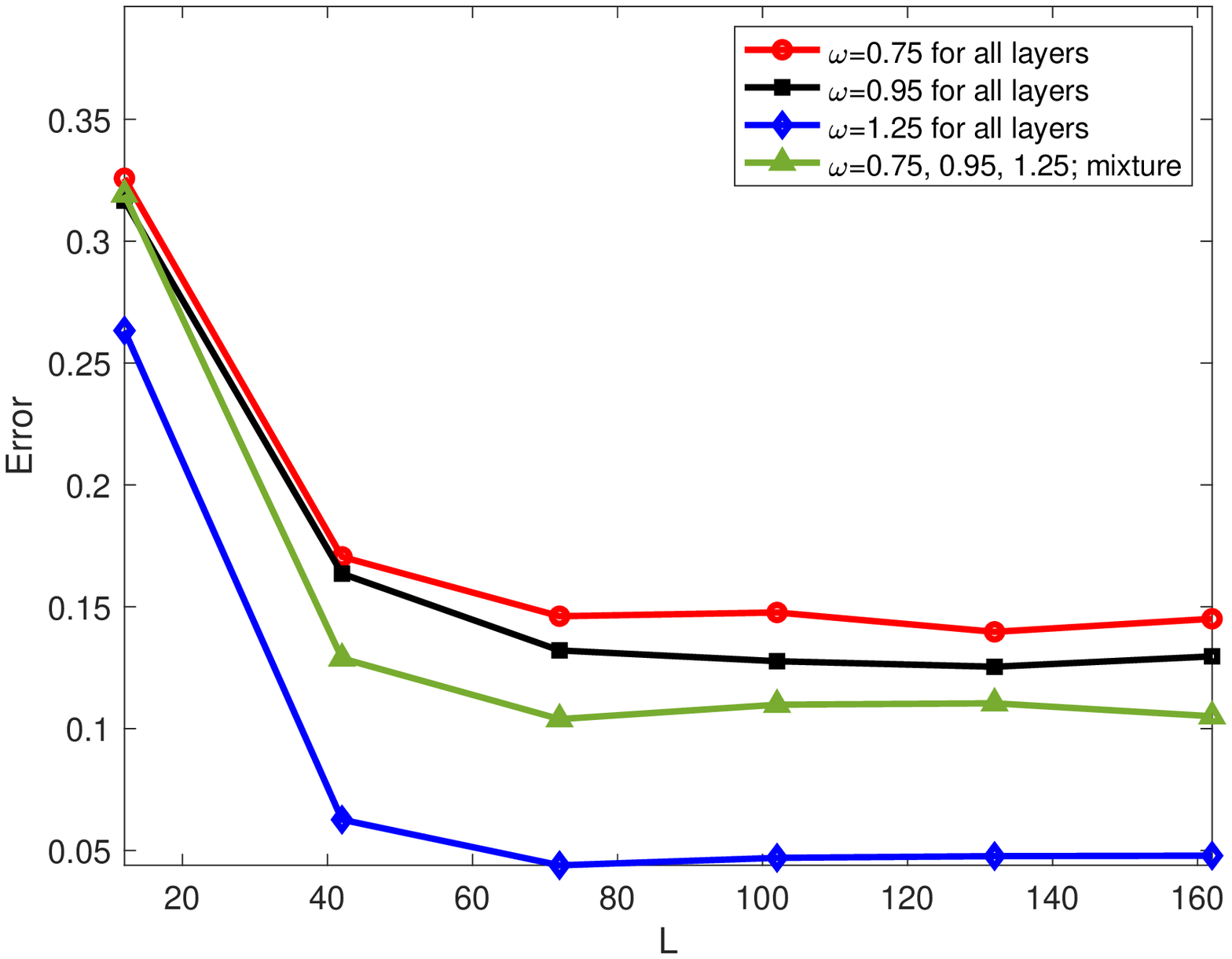}
\] 
\[ \includegraphics[width=2.47in] {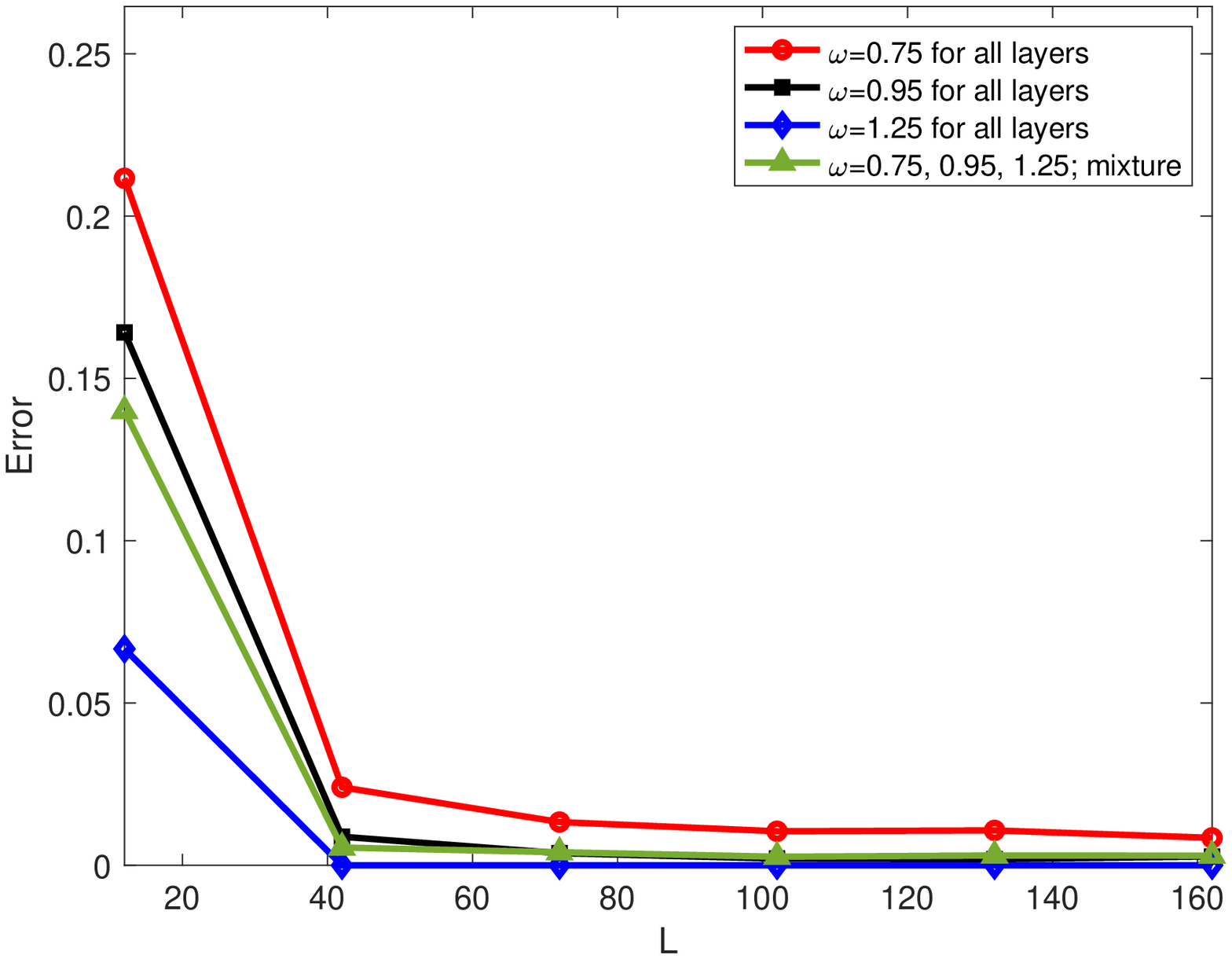} 
\includegraphics[width=2.47in]{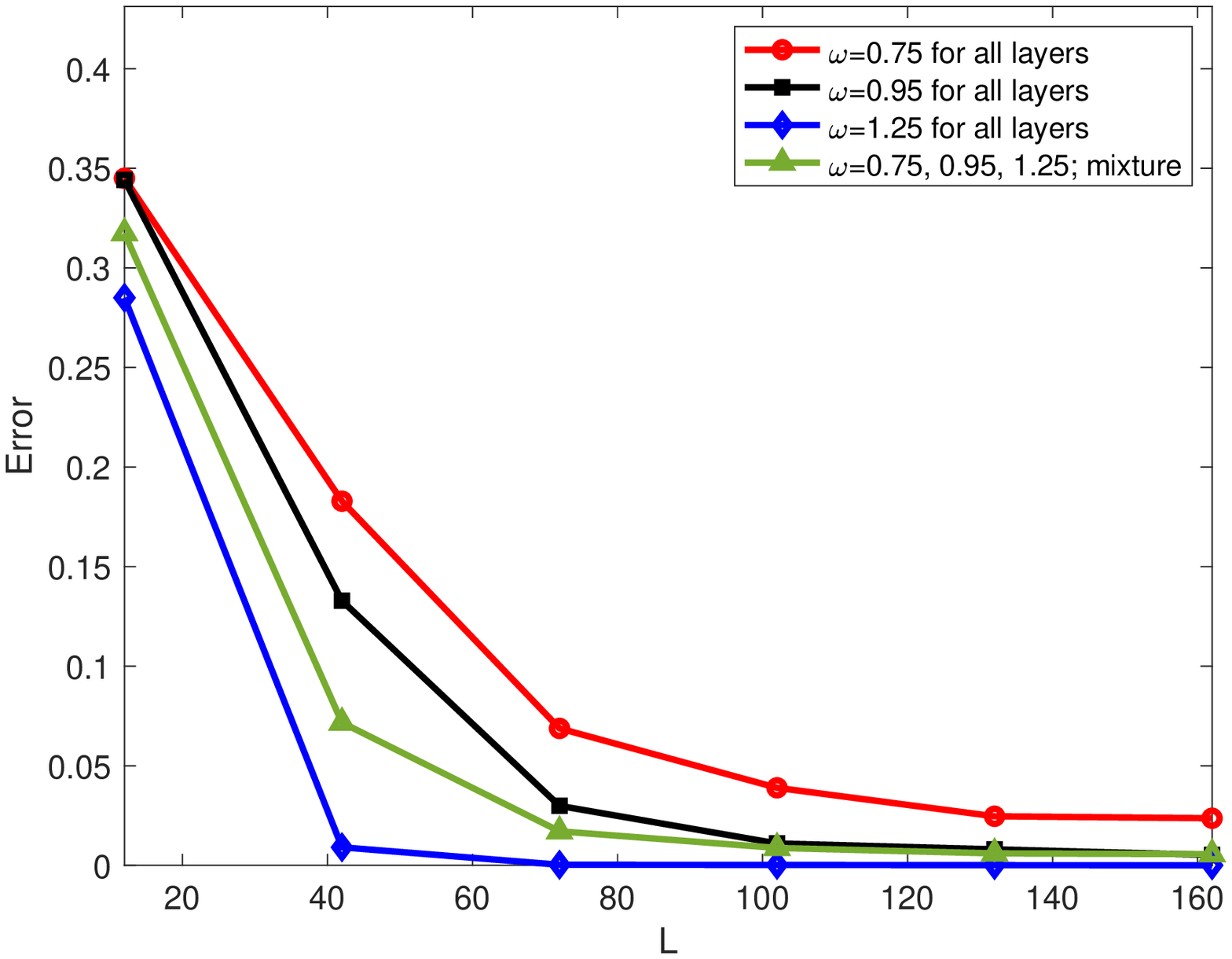}
\] 
\caption{ The clustering errors for $K=3$, $M=3$ (top left), $K=3$, $M=6$ (top right), $K=6$, $M=3$ (bottom left), 
and $K=6$, $M=6$ (bottom right). The number of layers ranges from $L=12$ to $L=162$ with the increments of 30.
The number of nodes, $n=100$, is fixed for all panels. The lines with different colors represent the results for different values of $\om$: 
$\om=0.75$  (red); $\om=0.95$  (black); $\om=1.25$  (blue); $\om=0.75, 0.95, 1.25$  (green). 
The errors are evaluated over 100 simulation runs. 
}
\label{mn:fig2}
\end{figure}

}

In our simulations, we generate layer and node memberships as multinomial random variables, 
as it is described in Section~\ref{sec:assump}, where $\pi_k = 1/K$, $k \in [K]$, and 
$\varpi_m = 1/M$, $m \in [M]$. The block probability matrices $B^{(l)}$ are generated as follows.
First, elements of the diagonal and the lower halves of   $B^{(l)}$ are generated as independent 
uniform random variables on the interval $[a,b]$, and the upper halves are obtained by symmetry.
Subsequently, all non-diagonal elements of  $B^{(l)}$ are multiplied by assortativity parameter $\om > 0$. 
When $\om <1$ is small, the layer networks are assortative; when $\om >1$ is large, they are  
disassortative; otherwise, they can be neither.
We find the probability matrices $P^{(l)}$ using \eqref{eq:model}, and generate symmetric adjacency matrices 
$A^{(l)}$, $l=1, ...,L$, with the lower halves obtained as   independent Bernoulli variables  
$A^{(l)}_{i,j}\sim \mbox{Bernoulli}(P^{(l)}_{i,j})$, $1 \leq j < i \leq n$.
Finally, we set $A^{(l)}_{i,j} = A^{(l)}_{j,i}$ when $j >i$, and $A^{(l)}_{i,i}=0$
since diagonal elements are not available.

We apply Algorithms~\ref{mn:SSC:alg} and \ref{alg:SSC:between-clust} to find the clustering matrix $\hC$. 
In Algorithm~\ref{mn:SSC:alg}, the tuning parameter $\lambda$ is chosen empirically from synthetic networks 
as $\lambda=4 \bar{\wh Q}$ where $\bar{\wh Q}$ is the average of the absolute values of entries of matrix $\wh Q$ defined in \eqref{eq:matrY}. 

We carry out simulations with $K = 3$ or 6 and $M=3$ or 6.
In our simulations, we choose $[a,b]=[0.3, 0.8]$ and use three different values for $\om$, $\om=0.75$, 0.95, and 1.25. 
Specifically, we generate four types of multilayer networks: (i) a multilayer network whose all layers are generated using $\om=0.75$; 
(ii) a multilayer network whose all layers are generated using $\om=0.95$; 
(iii) a multilayer network whose all layers are generated using $\om=1.25$; 
(iv) a multilayer network whose each one-third of layers corresponds to one of those three values of $\om$ 
(a mixture of types (i), (ii), and (iii)).

{\spacingset{1.4}

\begin{figure}[t]
\[ \includegraphics[width=2.47in] {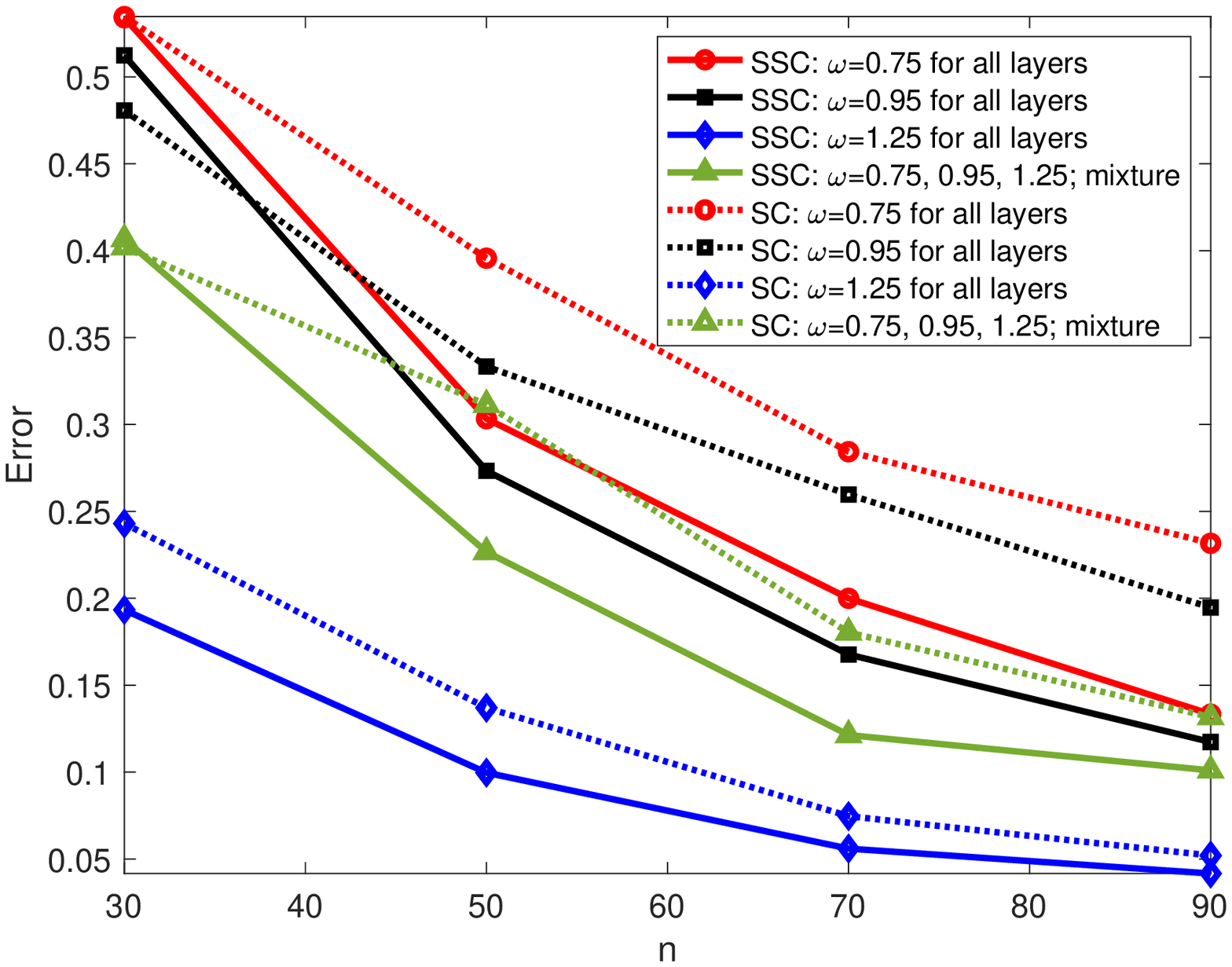}  
\includegraphics[width=2.47in]{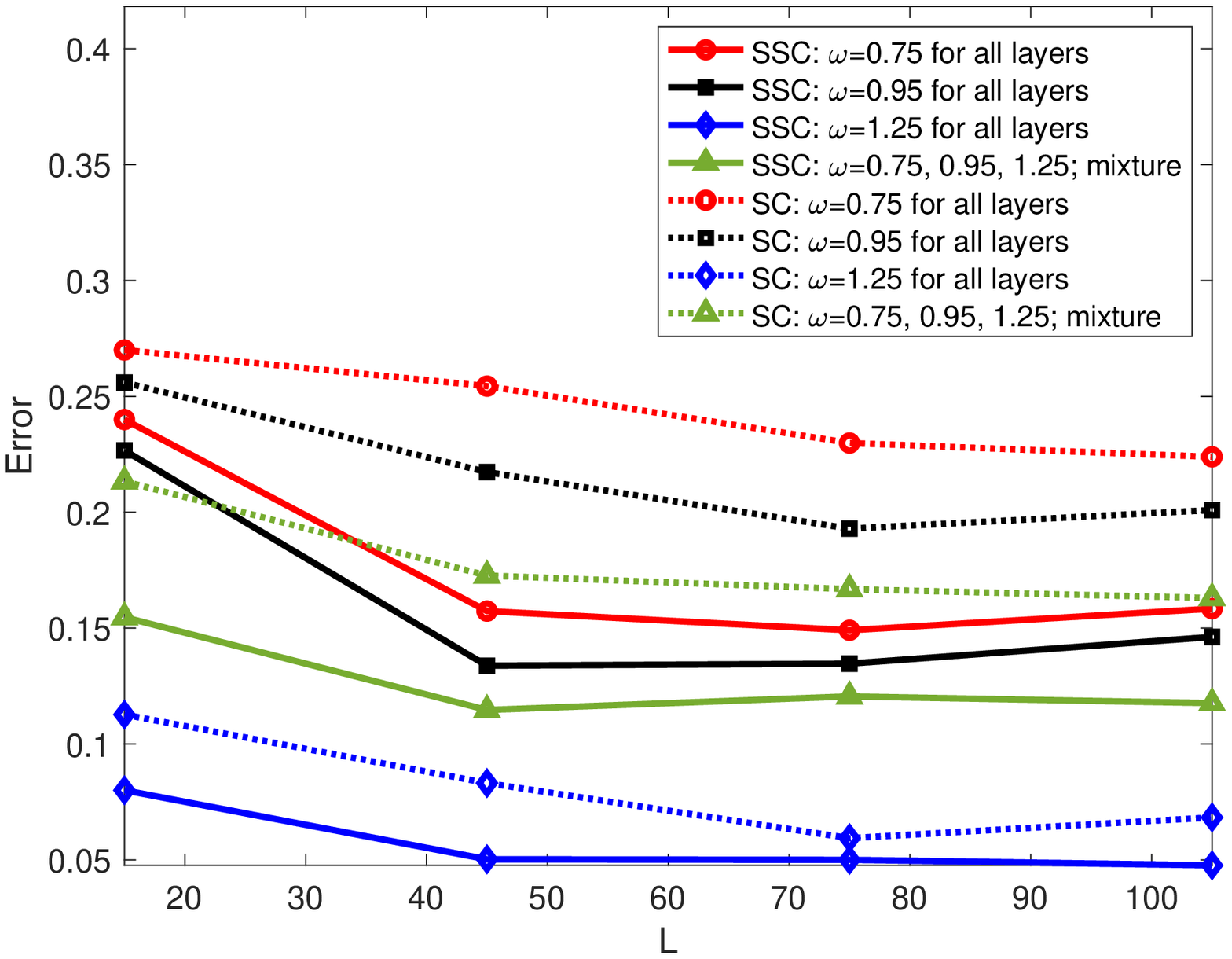}
\] 
\caption{ Comparison of clustering errors between Algorithm~\ref{alg:SSC:between-clust} (solid) 
and the method described in \cite{pensky2021clustering} (dotted) for $K=3$, $M=3$. 
Left panel: $L=60$ and the number of nodes ranges from $n=30$ to $n=90$ with the increments of 20. 
Right panel: $n=80$  and the number of layers ranges from $L=15$ to $L=105$ with the increments of 30.
}
\label{mn:fig3}
\end{figure}
 }

Figure 1 displays the between layer clustering errors for four types of multilayer networks with the fixed number of layers $L=60$,
and  $K=3$, $M=3$ (top left), $K=3$, $M=6$ (top right), $K=6$, $M=3$ (bottom left), and $K=6$, $M=6$ (bottom right). 
The number of nodes ranges from $n=60$ to $n=360$ with the increments of 60. 
Figure 2 displays the between layer clustering errors for four types of multilayer networks with the fixed number of nodes $n=100$,
and  $K=3$, $M=3$ (top left), $K=3$, $M=6$ (top right), $K=6$, $M=3$ (bottom left), and $K=6$, $M=6$ (bottom right). 
The number of layers ranges from $L=12$ to $L=162$ with the increments of 30.

Each of the panels presents all four scenarios for parameter $\om$. It is easy to see that $\om =1.25$ leads to the smallest 
and $\om = 0.75$   to the largest between-layer clustering errors. This   is due to the fact 
that smaller values of $\om$ lead to sparser networks, and the between-layer clustering error decreases  when $n \rho_n$ grows.
The latter shows that parameter $\om$ does not act as a ``signal-to-noise'' ratio in the between-layer clustering, as it happens 
in community detection in the SBM. Indeed, if this were true, then the between layer clustering error would be smaller for $\om = 0.75$ 
than for $\om = 0.95$.

It is easy to see that, for a fixed value of $L$,  the between layer clustering errors approach zero for all four 
types of networks as $n$ increases (since $\rho_n$ depends on the value of $\om$ only and is fixed). 
On the other hand, when $n$ is fixed and $L$ grows, the between layer clustering errors
decrease initially and then flattens. This agrees with the assessment of \cite{pensky2021clustering} where the authors 
observed the similar phenomenon. Indeed, growing  $L$  neither increases separation between subspaces, nor 
decreases the random deviations between the true vectors $\bx^{(l)}$ and their estimated versions $\by^{(l)}$.
Initial decrease in the error rate is due to initial decrease in the values of $\aleph_{w,K}$ in Assumption~{\bf A5},
the value of which flattens as $L$ grows.
Also, as both figures show, the errors, for fixed $n$ and $L$,  are   larger for larger values of $K$ and $M$ (in all cases), 
which agrees with our theoretical assessments.

Figure 3 shows the results of comparison of the between layer clustering errors of Algorithm~\ref{alg:SSC:between-clust} 
 and the technique described in \cite{pensky2021clustering}, which is based on the Spectral Clustering (SC). 
Since the algorithm used in \cite{pensky2021clustering}  is computationally very expensive as $n$ grows, 
we have compared the performances of these two methods using relatively small values of $n$. Specifically, 
Figure 3 illustrates the performances of the methods in two scenarios: fixed $L=60$ and  the number 
of nodes ranging from $n=30$ to $n=90$ with the increments of 20  (left panel); fixed $n=80$ and 
the number of layers ranges from $L=15$ to $L=105$ with the increments of 30 (right panel).
 For both panels, $K=3$ and $M=3$. It is easy to see that, Algorithm~\ref{alg:SSC:between-clust} is competitive 
with the clustering method used in \cite{pensky2021clustering}. In fact, the former outperforms the latter in almost all cases.



\section{A Real Data Example}
\label{sec:RealData}

In this section, we apply the proposed method to the Worldwide Food
Trading Networks data collected by the Food and Agriculture Organization of the United
Nations. The data have been described in \cite{de2015structural}, and it is available at\\
{\tt https://www.fao.org/faostat/en/\#data/TM}.  The data includes export/import trading
volumes among 245 countries for more than 300 food items.
In this multiplex network, layers represent food products, nodes are countries and edges at each layer 
represent import/export relationships of a specific food product among countries.


{\spacingset{1.4}

\begin{table}  [t]
\begin{center}
\caption{ List of three groups of food products obtained by Algorithm~\ref{alg:SSC:between-clust}}
\label{mn:table1}
\begin{tabular}{|c|c|}
\hline
 & Food Products  \\
\hline
Group 1 &  "Macaroni", "Pastry", "Rice, paddy (rice milled equivalent)",  \\ 
 & "Rice, milled", "Cereals, breakfast", "Mixes and doughs",  \\
 & "Food preparations, flour, malt extract", "Wafers", "Sugar nes", \\
 & "Sugar confectionery", "Nuts, prepared (exc. groundnuts)",\\ &"Vegetables, preserved nes", "Juice, orange, single strength",  \\
 & "Juice, fruit nes", "Fruit, prepared nes", "Beverages, non alcoholic",  \\
 & "Beverages, distilled alcoholic", "Food wastes", "Coffee, green", \\
 & "Coffee, roasted", "Chocolate products nes", "Pepper (piper spp.)", \\
 & "Pet food", "Food prep nes", "Crude materials"   \\
\hline
Group 2 & "Flour, wheat", "Flour, maize", "Infant food", "Sugar refined",    \\
& "Oil, sunflower", "Waters,ice etc", "Meat, cattle, boneless (beef \& veal)",  \\
& "Butter, cow milk", "Buttermilk, curdled, acidified milk", "Milk, whole dried",  \\
& "Milk, skimmed dried", "Cheese, whole cow milk", "Cheese, processed", \\
& "Ice cream and edible ice", "Meat, pig sausages", "Meat, chicken", \\
& "Meat, chicken, canned", "Margarine, short" \\
\hline
Group 3 & "Wheat", "Maize", "Potatoes", "Potatoes, frozen", "Sugar Raw Centrifugal", \\
& "Lentils", "Groundnuts, prepared", "Oil, olive, virgin", \\
& "Chillies and peppers, green", "Vegetables, fresh nes", \\
& "Vegetables, dehydrated", "Vegetables in vinegar", "Vegetables, frozen", \\
& "Juice, orange, concentrated", "Apples", "Pears", "Grapes",\\ & "Dates", "Fruit, fresh nes", "Fruit, dried nes", "Coffee, extracts",\\ 
& "Tea", "Spices nes", "Oil, essential nes", "Cigarettes"  \\
\hline
\end{tabular} 
\end{center}
\end{table}

}


In our analysis we used data for the year 2018.
As a pre-processing step, we remove low density layers and nodes. The original dataset contains 207 countries 
and 395 traded products. We choose the countries that are active in trading of at least 70\% of products, 
reducing the number of countries to 130. To create a network for each product, we draw an edge between   two countries 
if the export/import value of the  product   exceeds \$10,000. 
After that, we choose  the layers whose average degrees are larger than 10\%, 
that is, layers with average degrees of at least 13. This reduces the number of layers to 68,
so that the final multilayer network has 68 layers with 130 nodes in each layer.

Subsequently, we use Algorithm~\ref{alg:SSC:between-clust} to partition layers of the multiplex network (food products) into groups.
For this purpose, we choose $K=5$ which is consistent with the number of continents (Africa, Americas, Asia, Europe, and Oceania) 
and can be considered as a natural partition of countries, although different groups of layers may have different communities.
After experimenting with different values of $M$, we choose $M=3$ since it provides the most meaningful clustering results.

Table 1 represents the list of food products in the three resulting clusters. 
As is evident from Table 1, group 1 contains mostly cereals, stimulant crops, and derived products;  
group 2 consists mostly of animal products, and most products in group 3 are fruits, vegetables, and products
derived from them (like tea or vegetable oil).


{\spacingset{1.4}

\begin{table}   [t]
\begin{center}
\caption{ List of three groups of food products obtained by ALMA} 
\label{mn:table2}
\begin{tabular}{|c|c|}
\hline
 & Food Products  \\
\hline
Group 1 & "Pastry", "Sugar confectionery", "Fruit, prepared nes", \\
& "Beverages, non alcoholic", "Beverages, distilled alcoholic",  \\
& "Food wastes", "Chocolate products nes", "Food prep nes", "Crude materials"\\
\hline
Group 2 & "Cereals, breakfast", "Infant food", "Wafers", "Mixes and doughs", \\
& "Food preparations, flour, malt extract", "Potatoes", "Potatoes, frozen", \\
& "Oil, sunflower", "Chillies and peppers, green", "Vegetables, frozen",  \\
& "Apples", "Waters,ice etc", "Coffee, roasted", "Cigarettes", "Pet food", \\
& "Meat, cattle, boneless (beef \& veal)", "Butter, cow milk", \\
& "Buttermilk, curdled, acidified milk", "Milk, whole dried",  \\
& "Milk, skimmed dried", "Cheese, whole cow milk", "Cheese, processed",  \\
& "Ice cream and edible ice", "Meat, pig sausages", "Meat, chicken", \\
&"Meat, chicken, canned", "Margarine, short" \\
\hline
Group 3 & "Wheat", "Flour, wheat", "Macaroni", "Rice, paddy (rice milled equivalent)", \\
& "Rice, milled", "Maize", "Flour, maize", "Sugar Raw Centrifugal",  \\
& "Sugar refined", "Sugar nes", "Lentils", "Nuts, prepared (exc. groundnuts)", \\
& "Groundnuts, prepared", "Oil, olive, virgin", "Vegetables, fresh nes",  \\
& "Vegetables, dehydrated", "Vegetables in vinegar",  \\
& "Vegetables, preserved nes", "Juice, orange, single strength",  \\
& "Juice, orange, concentrated", "Pears", "Grapes", "Dates", "Fruit, fresh nes", \\
& "Fruit, dried nes", "Juice, fruit nes", "Coffee, green", "Coffee, extracts",  \\
& "Tea", "Pepper (piper spp.)", "Spices nes", "Oil, essential nes" \\
\hline
\end{tabular} 
\end{center}
\end{table}
}


In order to study communities for each of the three groups of layers, we apply the bias-adjusted clustering algorithm of 
\cite{lei2021biasadjusted}. Indeed, it is shown to be more robust than pure averaging of adjacency matrices since
we cannot be sure that all layers of the network are assortative. 
Figure~4 confirms that communities are indeed different for different types of food layers.
Indeed, for Group~1, community~1 mainly includes Middle East, South Asia and Australia, 
community~2 - Central and South America, community~3 - Africa, community~4 - former Soviet Union,
community~5 - Western Europe and Indochina. 
For Group~2, community~1  is comprised of countries in Middle East and Northern and Central Africa,
community~2 - Central and South America and  some Central Asian Countries,  community~3 - Southern and Central  Africa,
community~4 - Western Europe, community~5 -  Canada, US, Asia and Australia.
Finally, for Group~3, communities are much more mixed and scattered over the continents.

  \begin{figure}[th]
  \[ \includegraphics[width=3.05in] {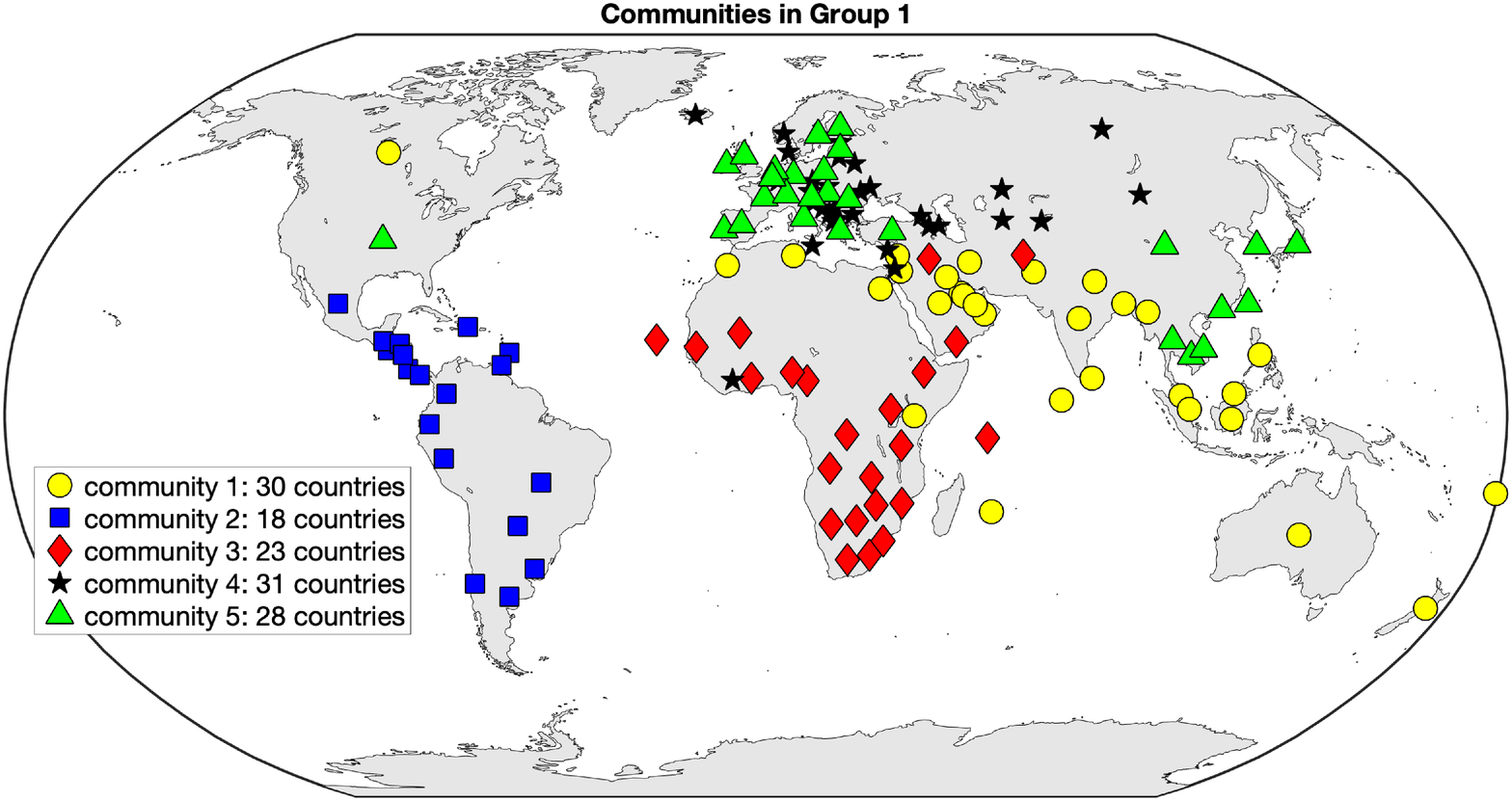}  \] 
  \[ \includegraphics[width=3.05in] {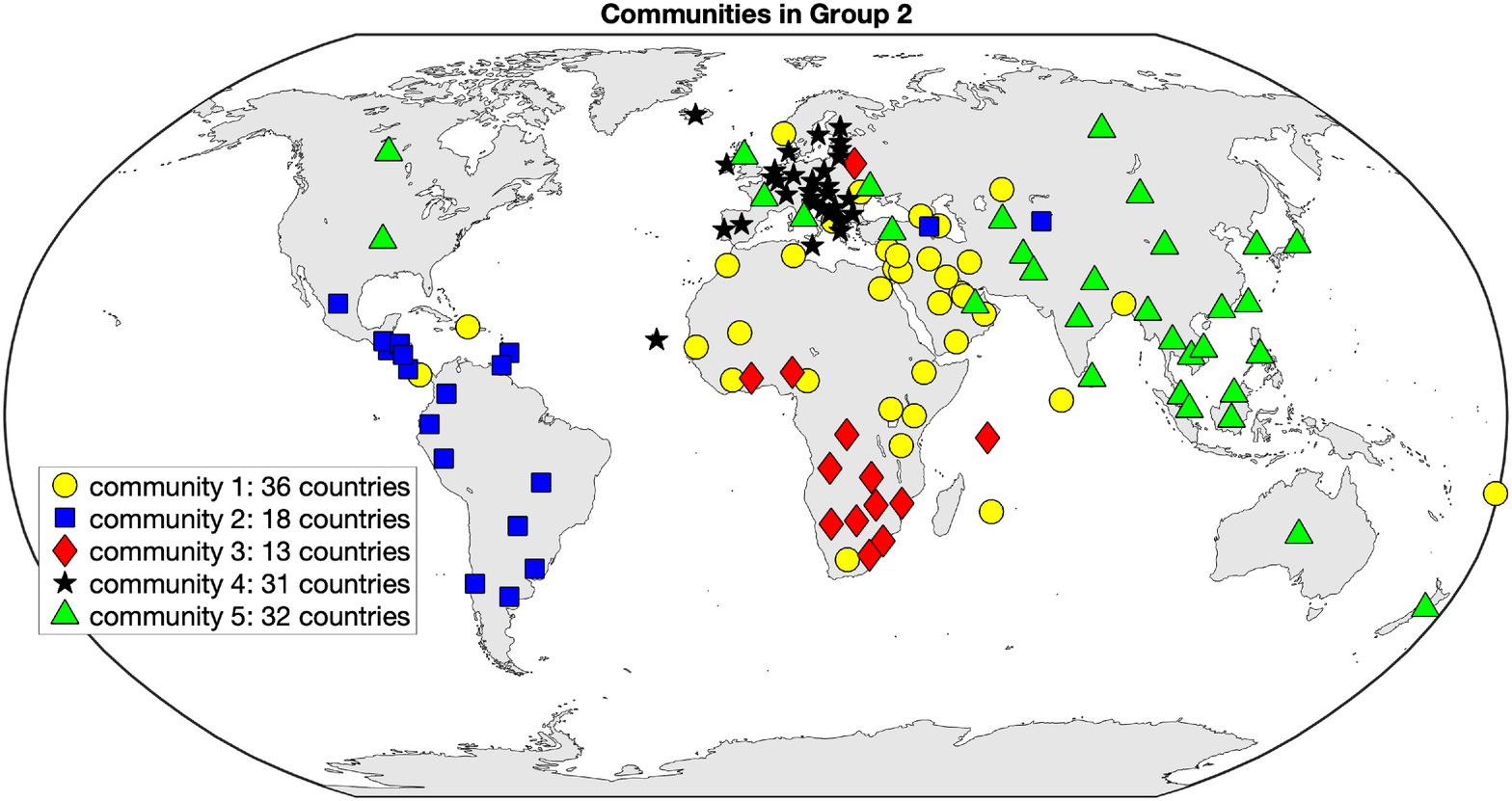}  \] 
  \[ \includegraphics[width=3.05in] {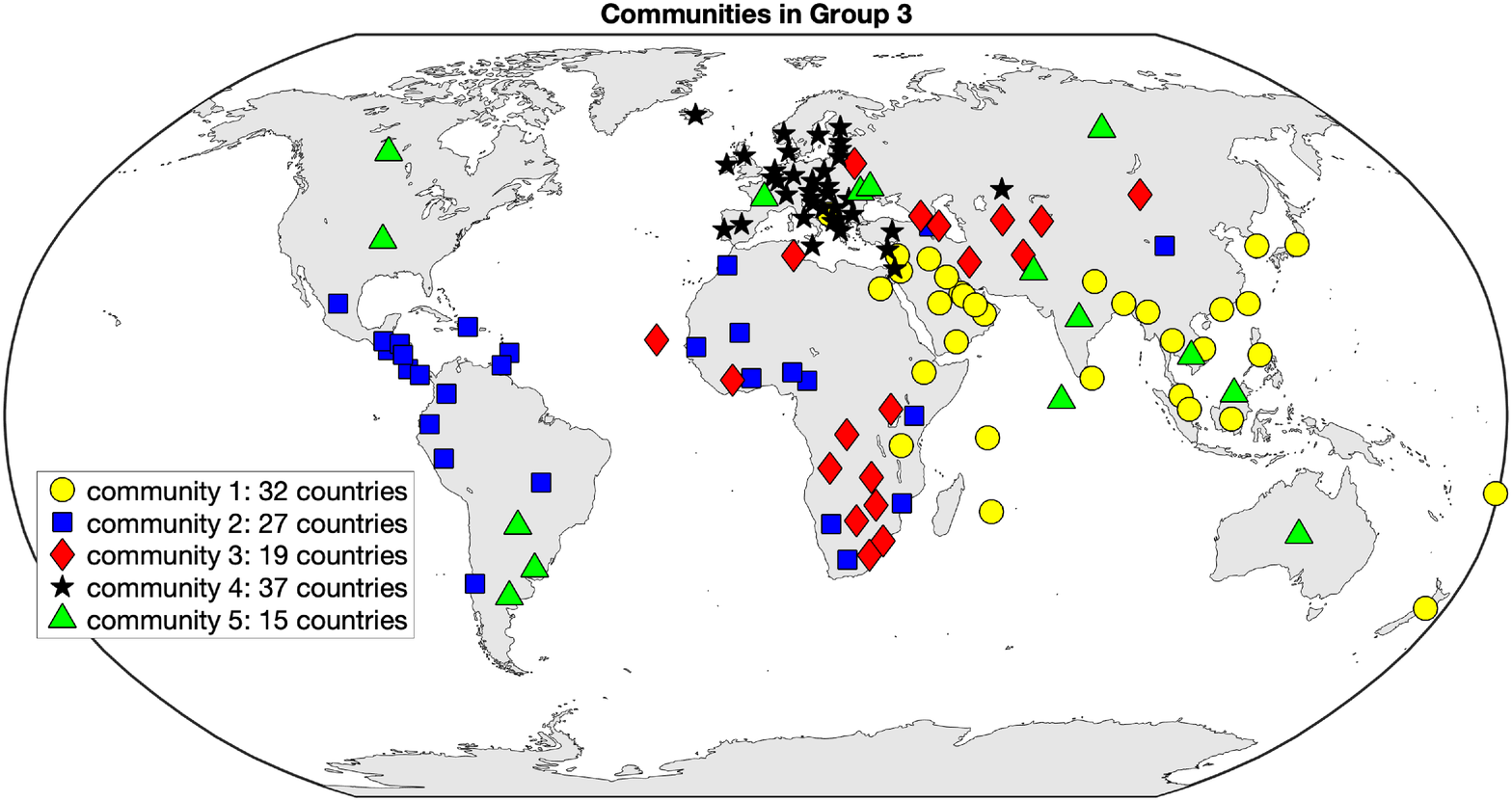}  \] 
   \caption{Trading communities of countries for products in groups 1,2 and 3.  
   }
   \label{mn:fig4}
  \end{figure}

As a comparison, we also carry out clustering of layers using  the Alternating Minimization Algorithm (ALMA) 
introduced in \cite{fan2021alma} for clustering a multiples network that follows the MMLSBM.
ALMA is known to be  competitive with TWIST (\cite{TWIST-AOS2079}), another clustering method employed for the MMLSBM.
The purpose of the comparison is to prove that, due to the flexibility of the DIMPLE model, 
it allows a better fit to the data than the MMLSBM.

Table 2 contains the list of food products in the three resulting groups obtained by ALMA 
with $K=5$ and $M=3$. We see that, similarly to the results in Table 1, most animal products 
are in group 2 and most fruits, vegetables, and derived products are classified in group 3. 
However, we do not see any dominant type of products in group 1 and the products 
do not seem to have meaningful relationships. Other possible values of $M$ don't lead to meaningful groups either. 
Hence, the network does not seem to fit the MMLSBM well. 

Therefore, based on results in Table 1 and Table 2, we can conclude that the DIMPLE model, 
with our proposed clustering method, is more suitable for this network.



\section {Discussion}
\label{sec:discussion}

The present paper considers the  DIverse MultiPLEx (DIMPLE) network  model, introduced in Pensky and Wang \cite{pensky2021clustering}.
However, while \cite{pensky2021clustering} applied spectral clustering to the proxy of the  adjacency tensor,
this paper uses  the SSC  for identifying groups of layers with identical community structures.
We provide algorithms for the between-layer clustering and formulate  sufficient conditions, under which these
algorithms lead to the strongly consistent clustering. Indeed, if the number of nodes is large enough, 
then the clustering error becomes zero with high probability.

The SSC has been applied to clustering single layer networks in \cite{noroozi2021hierarchy},
 \cite{noroozi2021sparse} and  \cite{noroozi2021estimation}.
To the best of our knowledge, our paper offers the first application of the SSC to the
Bernoulli multilayer network.  While the weights in Algorithm~\ref{mn:SSC:alg} are obtained 
in a relatively conventional manner, our between-layer clustering Algorithm~\ref{alg:SSC:between-clust}
is entirely original and very different from the one in \cite{pmlr-v51-wang16b}.

In addition, neither of \cite{noroozi2021hierarchy},
 \cite{noroozi2021sparse} and  \cite{noroozi2021estimation}
 offer any evaluation of clustering errors.
To the best of our knowledge,  this paper is the first one to provide  assessment of 
clustering precision of an SSC-based algorithm which is applied to a non-Gaussian network.
Specifically, majority of papers  provide theoretical guarantees 
for the sparse subspace clustering under the assumptions of spherical symmetry of the residuals
and  sufficient sampling density (see, e.g.,  \cite{soltanolkotabi2012},  \cite{soltanolkotabi2014}, 
\cite{10.5555/2946645.2946657}). It is easy to observe that rotational invariance fails 
in the case of the Bernoulli random vectors. In addition, the assumption that
the sampled vectors uniformly cover each of the subspaces may not be true either 
(for example, it does not hold for the MMLSBM).
For this reason, our paper offers a completely original proof of the clustering precision of the SSC-based
technique.

The present paper offers a strongly consistent between-layer clustering algorithm.
In comparison, the spectral clustering Algorithm~\ref{alg:PW_between} of \cite{pensky2021clustering}   leads,
with high probability, to the between layer clustering error of $O(K^2\, (n \rho_n)^{-1})$.
The latter results in a higher within-layer clustering error $R_{WL}^{(PW)}$. Indeed,
assumptions in both papers are  similar, and, with high probability, for some constants~$C_1$~and~$C_2$, 
\begin{align} 
R_{WL}^{(PW)} & \leq C_1\, \lkr  \frac{M  K^4 \log (L+n) }{n\, \rho_n\, L}  +   \frac{M  K^6}{n\, \rho_n}  \rkr \nonumber\\
 & \label{eq:PW-WL}\\
R_{WL}  & \leq C_2\, \lkr  \frac{M  K^4 \log(L+n)}{n\, \rho_n\, L}\   + \frac{K^4}{n^2} \rkr \nonumber 
\end{align}
where the second expression in \eqref{eq:PW-WL} is a repetition of formula     \eqref{eq:within_er}.
While the first terms in the expressions of $R_{WL}^{(PW)}$ and $R_{WL}$  coincide, the second term
in $R_{WL}^{(PW)}$ is significantly larger than the one in  $R_{WL}$. Indeed,   the second term in $R_{WL}^{(PW)}$
results from the between layer clustering error of $O(K^2\, (n \rho_n)^{-1})$. On the contrary $K^4\, n^{-2}$
in \eqref{eq:within_er} is due to smaller order terms.

Clustering methodology in this paper has a number of advantages. Not only is it  strongly consistent with high probability when 
the number of nodes is large, but also competitive with  (and often more precise  than) the spectral clustering   in 
\cite{pensky2021clustering}.  In addition, the algorithm of \cite{pensky2021clustering} requires SVD of 
$n(n-1)/2 \times L$ matrix,
which is challenging for large $n$, while in our case the SVD is applied to $L \times L$ matrix. 
Hence, the SSC-based technique allows to handle much larger networks. Moreover, the most time consuming part of the algorithm, finding
the weight matrix, is  perfectly suitable for application of parallel computing which can drastically reduce the computational time.


\appendix 

\section{Proofs}
\label{sec:proofs}


\subsection{Proof of Self-Expressiveness property under the separation condition}

Proof of Theorem~\ref{th:self_express} relies on the fact that the subspaces    $\calS_m$, $m=1, ...,M$, 
corresponding to different types of layers do not have large intersections. 
Specifically, we prove the following statement from which the validity of    Theorem~\ref{th:self_express}
will readily follow. 


\begin{proposition} \label{prop:self_express}
Let Assumptions  {\bf A1}, {\bf A2}, {\bf A3} and {\bf A5} hold. Let $L_m$ and $n_k\upm$be, respectively,
the number of layers of type $m$ and the number of nodes in the $k$-th community in the group of layers of type $m$, where
  $L_m$ and $n_k\upm$ satisfy condition \eqref{eq:bal_cond}.
Assume,  in addition, that there  exists $\tau \equiv \tau_{n,K} \in (0,1)$ such that for any arbitrary  vectors $\bm x \in \calS_m$ 
and $\bm x' \in \calS_{m'}$,  where $m \neq m'$,  one has $|\bm x^T \bm x'| \leq \tau\, \|\bm x\|\, \|\bm x'\|$. 
Let $t > 0$ and $\del = \delta_{n,K,t}$ be defined in \eqref{eq:delnk} where  $C_{t, \del}$ is a constant that depends only on $t$ 
and constants in Assumptions   {\bf A1}, {\bf A2}, {\bf A3} and {\bf A5}, and condition \eqref{eq:bal_cond}.

Let $\wh W$ be a solution of problem \eqref{mn:lasso} with $\lambda = \lam_{n,K}$ such that 
\be \label{eq:param_cond}
\lam_{n,K} \leq (4\, \aleph_{w,K})^{-1}, \quad 
\lim_{n \to \infty} \frac{(\delta_{n,K,t} + \tau_{n,K})(1 + \aleph_{w,K})}{\lam_{n,K}} = 0 
\ee
where $\aleph_{w,K}$ is defined in Assumption~{\bf A5}.
If $\di{\max_{1 \le l \le L} \|\by^{(l)} - \bx^{(l)}\|  \leq \delta} $
and $n$ is large enough,   then, 
matrix $\wh W$ (and, consequently, $\wh{\wt W}$)  satisfies the SEP. 
\end{proposition}


\medskip

\noindent
{\bf Proof of Proposition \ref{prop:self_express}. }
\\
Let matrices $Q, X  \in \RR^{n^2 \times L}$  and $\wh Q, Y \in \RR^{n^2 \times L}$ 
be defined in \eqref{eq:matrX} and \eqref{eq:matrY}, respectively. 
Choose an arbitrary $l_0 \in [L]$ and, without loss of generality, assume that $c(l_0) = 1$, i.e. $\bm x^{(l_0)} \in \calS_1$. 
Denote $\bx = \bx^{(l_0)}$, $\by = \by^{(l_0)}$,  $\wt \calS = \calS_1$ and $\wt{ \wt \calS} = \calS_2 \cup \ldots \cup \calS_M$, 
and present the remainder of matrix $X$ (i.e., $X$ with $X(:,l_0)$ removed)  as $[\wt X \, | \, \wt{\wt X}]$.
Here,   $\wt X$ and $\wt{\wt X}$ are portions of   $X$ with $X(:,l_0)$ removed, that correspond to $\wt \calS$ and $\wt{\wt S}$, respectively.
With some abuse of notations, we denote  $X$ with $X(:,l_0)$ removed by $X$ again, i.e., $X=[\wt X \, | \, \wt{\wt X}]$.

Denote $Z=Y-X$, $\bz^{(l)} = Z(:,l)$ and  $\bz = \bz^{(l_0)} = \by - \bx$, so that 
\bes 
\wt Y=\wt X+\wt Z, \quad \wt {\wt Y}=\wt{\wt X}+\wt{\wt Z}, \quad \bm y=\bm x+\bm z
\ees

\noindent
%
Let   $\bm w= [ \wt {\bm w}  \, | \, \wt{\wt {\bm w}} ]$ be the solution of problem \eqref{mn:lasso} for $l=l_0$. Then,
\eqref{mn:lasso}  implies that
\bes
\| \bm y - \wt Y \wt {\bm w} - \wt{\wt  Y} \wt{\wt {\bm w}} \|^2 + 2 \lambda \|\wt {\bm w} \|_1 + 
2 \lambda \|\wt{\wt {\bm w}} \|_1 \le \| \bm y - \wt Y \wt {\bm w} \|^2 + 2 \lambda \|\wt {\bm w} \|_1
\ees
By simplifying the inequality, obtain   
\be  \label{eq:Del_def}
\Delta \overset{def}{=} \| \wt{\wt  Y} \wt{\wt {\bm w}} \|^2 - 2  \langle \bm y - \wt Y \wt {\bm w} , \wt{\wt  Y} \wt{\wt {\bm w}} \rangle 
+ \lambda \|\wt{\wt {\bm w}} \|_1 \le 0
\ee
Note that  the Cauchy-Schwarz inequality and Assumption~{\bf A5} yield  
\begin{align*}
   \langle \bm y - \wt Y \wt {\bm w} , \wt{\wt  Y} \wt{\wt {\bm w}} \rangle 
\le & \tau   \| \bm x - \wt X \wt {\bm w} \| \| \wt{\wt  X} \wt{\wt {\bm w}} \| 
+  \| \bm z - \wt Z \wt {\bm w} \| \| \wt{\wt  X} \wt{\wt {\bm w}} \| \\
+ & \| \bm x - \wt X \wt {\bm w} \| \| \wt{\wt  Z} \wt{\wt {\bm w}} \| 
+  \| \bm z - \wt Z \wt {\bm w} \| \| \wt{\wt  Z} \wt{\wt {\bm w}} \|
\end{align*}
Moreover,
\bes
\| \bm z - \wt Z \wt {\bm w} \| \le [\| \wt {\bm w} \|_1 +1 ] \delta; \quad 
\| \wt{\wt  Z} \wt{\wt {\bm w}} \| \le \| \wt{\wt {\bm w}} \|_1  \delta; \quad 
\| \bm x - \wt X \wt {\bm w} \| \le \| \wt {\bm w} \|_1 +1 
\ees
Since 
$ \| \wt{\wt  Y} \wt{\wt {\bm w}} \|^2  \ge 0.5\, \| \wt{\wt X} \wt{\wt {\bm w}} \|^2 - \| \wt{\wt  Z} \wt{\wt {\bm w}} \|^2$, 
%
obtain
\be
\begin{aligned} \label{eq:Delta_bound}
 \Delta   \ge & \left [ 0.5\,  \| \wt{\wt  X} \wt{\wt {\bm w}} \|^2 
- 2 (\tau  + \delta) (\|\wt {\bm w} \|_1 +1) \| \wt{\wt  X} \wt{\wt {\bm w}} \| \right ] \\
&+ \| \wt{\wt {\bm w}} \|_1 \left [  \lambda - \delta   - 2( \|\wt {\bm w} \|_1 +1) \delta (1 + \delta) \right ]
\end{aligned}
\ee
To find an upper bound for $( \|\wt {\bm w} \|_1 +1)$, consider $\wt {\bm w}_*$, the solution of exact problem, that is
$\bm x = \wt X  \wt {\bm w}_*$. By Assumption~{\bf A6}, there exists a sub-matrix $\wt X_* \in \RR^{n^2 \times (K-1)^2}$  of $\wt X$,
such that $\bx = \wt X_*    {\bm w}_*$ and $\|\bm w_* \|_1 \leq \aleph_{w,K}$. 
Let $\wt Y_*$ be the portion of $\wt Y$ corresponding to $\wt X_*$ and $\wt Z_* = \wt Y_* - \wt X_*$. 
Since $\| \bm y - \wt Y \bm w_* \|^2  = \| \bm z - \wt Z \bm w_* \|^2 $, derive
\be \label{eq:exact_sol_lasso}
\| \bm y - \wt Y \bm w_* \|^2 + 2 \lambda \|\bm w_* \|_1 \le \delta^2 \left [  \| \bm w_* \|_1 +1 \right ]^2 + 2 \lambda \| \bm w_* \|_1
\ee
Note that, since $\bw_*$ is not an optimal solution, one has
\bes
\| \bm y - \wt Y \bm w_* \|^2 + 2 \lambda \| \bm w_* \|_1 \geq 
\| \bm y - \wt Y \wt {\bm w} - \wt{\wt  Y} \wt{\wt {\bm w}} \|^2 
+ 2 \lambda \|\wt{\bm w} \|_1 + 2 \lambda \|\wt{\wt{\bm w}} \|_1
\geq  2 \lambda \|\wt{\bm w} \|_1
\ees
Thus, $\|\wt{\bm w} \|_1 + 1 \leq (\| \bm w_* \|_1 + 1) +  \| \bm y - \wt Y \bm w_* \|^2/(2 \lam)$, so that  
\be \label{eq:w_tilde_bound}
\|\wt{\bm w} \|_1 + 1 \leq  (1 + \aleph_{w,K}) + 0.5\, \delta^2\, (1 + \aleph_{w,K})^2 / \lambda 
\ee
Then, using  \eqref{eq:Delta_bound}  and \eqref{eq:w_tilde_bound}, due to $\|\wt{\wt X} \wt{\wt {\bm w}} \| \le \|\wt{\wt{\bm w}} \|_1$, obtain
\begin{align*} 
  \Delta \ge  & \frac{1}{2}\,  \| \wt{\wt  X} \wt{\wt {\bm w}} \|^2  \\
  &+ \|\wt{\wt {\bm w}}\|_1 \left [ \lambda - \delta   -  2\,(1 + \aleph_{w,K}) 
\left ( 1 +    \frac{1}{2} \delta^2 (1 + \aleph_{w,K})/\lambda  \right) (2 \delta + 2 \tau  + \delta^2) \right ]
\end{align*}
Now,  observe that, due to condition \eqref{eq:param_cond}, $\delta <1$ and  $\delta^2 (1 + \aleph_{w,K})/\lambda$ 
tends to zero. Hence,  for $n$ large enough, arrive at
\bes 
\Delta \ge  \frac{1}{2}  \| \wt{\wt  X} \wt{\wt {\bm w}} \|^2 
 +   \lambda \|\wt{\wt {\bm w}}\|_1 \left [ 1 -  \frac{\delta}{\lambda} -  \frac{12\, (\delta + \tau)(1 + \aleph_{w,K})}{\lam}
  \right ] > 0
\ees
unless $\wt{\wt {\bm w}} = 0$. Since, by \eqref{eq:Del_def}, $\Delta \le 0$,
one has  $\wt{\wt {\bm w}} = 0$ and the SEP holds.

In order to complete the proof, we need to show that there exists $\lambda$ which is not too large, 
so the optimization problem \eqref{mn:lasso} for $l=l_0$ has a non-zero solution. 
If we show that, for some $\bm w \neq 0$, the objective function is smaller than that 
for $\bm w \equiv 0$, then \eqref{mn:lasso} for $l=l_0$ yields a non-zero solution. 
To this end, we find a sufficient condition such that 
$\| \bm y - \wt Y \bm w_* \|^2 + 2 \lambda \|\bm w_* \|_1 \le \|\bm y \|^2 =1$ 
holds.  
It follows from \eqref{eq:exact_sol_lasso} and Assumption~{\bf A6} that
\bes 
\| \bm y - \wt Y \bm w_* \|^2 + 2 \lambda \|\bm w_* \|_1 \le \delta^2 ( 1 + \aleph_{w,K})^2 + 
2 \lambda  \aleph_{w,K}  
\ees
Hence,
\bes 
\delta^2\, (1 + \aleph_{w,K})^2 + 2 \lambda \aleph_{w,K}   \leq 1
\ees
is sufficient for $\bm w \neq 0$. 
By condition \eqref{eq:param_cond},  one has 
$  \delta\, (1 + \aleph_{w,K})  \to 0$ 
 as $n \rightarrow \infty$, so that for $n$ large enough, 
$   \delta\, (1 + \aleph_{w,K}) \leq 1/2$.
Therefore, $2 \lambda \aleph_{w,K}   \leq 1/2$ 
is sufficient for $\bm w \neq 0$, which is equivalent to the first inequality in  
\eqref{eq:param_cond}. The latter completes the proof.


\subsection{Proof  of Theorem~\ref{th:self_express} }  
\label{sec:th1_proof}

In order to prove that Theorem~\ref{th:self_express} holds, we show that,
under assumptions of Theorem~\ref{th:self_express}, \eqref{eq:bal_cond} is true and that 
 $\tau_{n,K}  \leq \bbC K^2\,  n^{-1} \, \log n$ in Proposition~\ref{prop:self_express}. 
Let $\hL_m$ and $\hn_k\upm$ be defined in \eqref{eq:hnk_hLm}.
%
Then, the following statements are valid. 

\medskip


\begin{lemma} \label{lem:balanced_groups}  
Let Assumption~{\bf A4} hold.  
Let  $t > 0$ satisfy condition \eqref{eq:tcond}.
Then, there exists a set $\bar{\Om}_{t1}$ with 
\bes 
\PP(\Om_{t1}) \ge 1- 2 L^{-t} - 2K M\,  n^{-t}
\ees
such that, for $\om \in {\Om}_{t1}$, one has simultaneously
\be \label{eq:bounds_prob}
\frac{\lowc_{\varpi}\, L}{2\, M}  
\leq \hL_m \leq \frac{3\, \highc_{\varpi}\ L}{2\, M}, \quad {\rm and} \quad 
\bigcap_{m=1}^{M}  \bigcap_{k=1}^{K} \left \{ \omega : \frac{\lowc_{\pi}\, n}{2\, K}  
\leq \hat n_k\upm \leq \frac{3\, \highc_{\pi}\ n}{2\, K}  \right \} 
\ee
\\
\end{lemma} 

\medskip


\noindent
Apply the following lemma, proved later in Section~\ref{sec:suppl_proofs}, which ensures   the upper bound  
$ \di \max_l \, \|\by^{(l)} - \bx^{(l)}\| \leq \delta_{n,K,t}$ in Proposition~\ref{prop:self_express}.

\begin{lemma} \label{lem:scaling}  
Let Assumptions of Theorem \ref{th:self_express} hold and $t > 0$ satisfies condition \eqref{eq:tcond}. Let $\bq^{(l)}$ and $\hat {\bm q}^{(l)}$
be defined in \eqref{eq:q_l} and \eqref{eq:matrY}, respectively.
Let matrices $Q, \wh Q   \in \RR^{n^2 \times L}$   be defined in \eqref{eq:matrX} and \eqref{eq:matrY}, respectively.  
Then, 
\be \label{eq:minF_Pl}
\min_l \| \bm q^{(l)} \|  
\ge  \wt C_0   C_{\sig,0} \, \underline{C}\, K^{-1/2}\, n \, \rho_n, \quad  
\max_l  \| \bm q^{(l)} \|  
\leq \wt{\wt C}_0   \,\bar C  \,   K^{-1/2}\, n \, \rho_n 
\ee
Moreover,
there exists a set $\Om_{t2}$ such that $\PP(\Om_{t2}) \geq 1-Ln^{-t}$, and for $\om \in \Om_{t2}$, one has
\be \label{eq:upperbound}
 \max_l\    \|\hat {\bm q}^{(l)} - \bm q^{(l)}\|/\| \bm q^{(l)} \|    \leq C_{t,\rho, \sig}\,  K/\sqrt{n \rho_n} 
\ee
where $C_{t,\rho,\sig}$ depends only on   $t$ and constants in Assumptions {\bf A1}--{\bf A5}. 
\end{lemma} 

\medskip
\medskip


 \noindent
In addition, the following lemma provides an upper bound on 
$\tau_{n,K}$ in Proposition~\ref{prop:self_express}. 

\medskip

\medskip


\begin{lemma} \label{lem:inner_product_bound}  
Let Assumption~{\bf A4} hold, and $Z_{j,k}^{(m)}$,  $k \in [K]$, $j  \in [n]$, $m \in [M]$, be generated according to 
\eqref{eq:rand_gen2}.  
Let  $t > 0$ satisfy condition \eqref{eq:tcond}.
Then, there exists a set $\Om_{t3}$ with 
\bes 
\PP(\Om_{t3}) \ge 1- 2 L^{-t} - 2K M (M+1)\,  n^{-t}
\ees
such that, for $\om \in \Om_{t3}$, and
for any arbitrary  vectors $\bm x \in \calS_m$ 
and $\bm x' \in \calS_{m'}$,  where $m \neq m'$,  one has $|\bm x^T \bm x'| \leq \tau\, \|\bm x\|\, \|\bm x'\|$ 
with 
\be  \label{eq:taunk}
\tau \equiv \tau_{n,K}  \leq  2 \, (\sqrt{2} + 3/\lowc_{\pi})^2\, t \, K^2\,  n^{-1} \, \log n 
\ee
\end{lemma}


\medskip 

\medskip

\noindent
It is  easy to show that, for any $l \in [L]$, one has 
$\|\bz^{(l)}\| = \|\bm y^{(l)} - \bm x^{(l)} \|  \leq   2 \|\hat{\bm q}^{(l)} - \bm q^{(l)} \|/\|\bm q^{(l)}\|$. Hence,  
Lemma \ref{lem:scaling} implies that, for $\del \equiv \delta_{n,K,t}$ defined in \eqref{eq:delnk}, one has
\be \label{eq:y-x}
\PP \left\{ \max_{1 \le l \le L} \|\bz^{(l)} \|  \leq \delta  \right \} \ge 1-L n^{-t}
\ee
Now, in order to apply Proposition~\ref{prop:self_express},  it remains to show that condition \eqref{eq:param_cond_th1} implies \eqref{eq:param_cond}. 
For this purpose, note that, since columns of matrix $X$ have unit norms, one has $\aleph_{w,K} \geq 1$ in {\bf A5} and, 
hence, \eqref{eq:param_cond} implies that  $\delta_{n,K,t}/\lam_{n,K} \to 0$ as $n \to \infty$.
The latter furthermore yields that $n^{-1}\, K^2\, \log n \to 0$, so  that 
$\tau_{n,K} = o(\delta_{n,K,t})$ as $n \to \infty$, where $\tau_{n,K}$ and $\delta_{n,K,t}$ are defined in \eqref{eq:taunk}
and \eqref{eq:delnk}, respectively. This completes the proof.

\ignore{
so that 
\be  \label{eq:cond_large_n}
\lambda_{n,K} - \delta_{n,K,t}   -  2\,(1 + \aleph_{w,K}) 
\left ( 1 +    \delta_{n,K,t}^2 (1 + \aleph_{w,K})/(2\, \lambda_{n,K})  \right) (2 \delta_{n,K,t} + 2 \tau  + \delta_{n,K,t}^2) >0,
\ee 
{\bf CHECK THIS!!}\\
}


\subsection{Proof  of Theorem~\ref{th:SSC_ClustErrors} } 

Let $\wh W$ be the matrix of weights and $\Omega_t$ be the set in 
Theorem~\ref{th:self_express}, so that  $\Om_t$ is exactly the set  where SEP holds. 
Note that Algorithm~\ref{alg:SSC:between-clust} allows the situation where $\wt M < M$. 
However, if the SEP holds, then no two network layers in different  clusters can be a
part of the same connected component, and hence, $\wt M \geq M$.

Consider a clustering function $\phi: [L] \rightarrow [\wt M]$ and the corresponding clustering matrix 
$\Phi \in \{0, 1\}^{L \times \wt M}$, which partitions $L$ layers into $\wt M  \ge M$, disconnected components.  
Due to SEP,  some of the vectors that belong to different clusters, according to $\phi $, belong to the same cluster, according to $c$. 
On the other hand, if two vectors belong to different clusters according to $c$, 
they belong to different clusters according to $\phi$.  That is, for $l_1, l_2 = 1,\ldots, L$, $l_1 \neq l_2$,
one has 
\be \label{eq:tc_cond}
 \phi (l_1) = \phi (l_2) \,  \Longrightarrow \,  c(l_1) = c(l_2), \quad 
  c(l_1) \neq c(l_2) \, \Longrightarrow   \, \phi (l_1) \neq \phi (l_2)
\ee 
Hence, if $\wt M = M$, then $\phi = c$.

Let $\wt M >M$. Then, due to \eqref{eq:tc_cond}, one can partition $\wt M$ clusters into $M$ groups. Let 
$\theta : [\wt M] \rightarrow [M]$ be such clustering function,   and $\Theta$ be the corresponding clustering matrix. 
Then, for $\omega \in \Omega_t$, SEP holds and $C = \Phi \Theta$.
Observe that $\theta(\wt m_1) = \theta(\wt m_2)$ if $c(l_i) = c(l_j)$ 
for all $l_i, l_j$ with $\phi (l_i) = \wt m_1$ and $\phi (l_j) = \wt m_2$, where 
$l_i, l_j= 1, \ldots, L,$ and $\wt m_1, \wt m_2= 1, \ldots, \wt M$. 
To prove the theorem, we use  the following statement.

\begin{lemma} \label{lem:angles_within_subspace}  
Let Assumptions {\bf A1} - {\bf A5} hold  and $K^2/(n \rho_n) \to 0$ as $n \to \infty$. 
Then, 
if $\om \in \Om_t$, for some positive constant $\wc C$,  one has
\begin{align} 
|(\bm x^{(l_1)})^T \bm x^{(l_2)} | & \ge \wc C
\quad  \text{if} \quad c(l_1) = c(l_2); \nonumber\\
 &  \label{eq:scalar_products_x} \\
|(\bm x^{(l_1)})^T \bm x^{(l_2)} | & \le \tau_{n,K} \quad  \text{if} \quad c(l_1) \neq c(l_2)
\end{align}
Moreover, for  $\omega \in \Omega_t$ and $n$  large enough  
\begin{align} 
& \min_{\substack{l_1, l_2 \\ c(l_1) = c(l_2)}} |(\bm y^{(l_1)})^T \bm y^{(l_2)} | \ge \wc C/2, \nonumber\\ 
& \label{eq:scalar_products_y} \\
& \max_{\substack{l_1, l_2 \\ c(l_1) \neq c(l_2)}} |(\bm y^{(l_1)})^T \bm y^{(l_2)} | \le \tau_{n,K} + 2 \delta_{n, K, t} \nonumber
\end{align}
\\
\end{lemma} 


Consider matrices $\Up, \wh \Up \in \RR^{L \times L}$ with elements
\bes
\Up_{l_1, l_2} = | (\bm x^{(l_1)})^T \bm x^{(l_2)} | , \quad  \wh \Up_{l_1, l_2} = | (\bm y^{(l_1)})^T \bm y^{(l_2)} | , 
\quad l_1, l_2 = 1,\ldots, L
\ees
Denote  $D_{\Phi} = (\Phi)^T \Phi$ and define matrices $\wt \Up,  \wh{\wt \Up} \in \RR^{\wt M \times \wt M}$  
\bes
\wt \Up =  (D_{\Phi})^{-1/2} \Phi^T \Up \Phi (D_{\Phi})^{-1/2}, 
\quad \wh{\wt \Up} = (D_{\Phi})^{-1/2} \Phi^T \wh \Up \Phi (D_{\Phi})^{-1/2},
\ees
Then, due to \eqref{eq:thresh_cond_new},  by Lemma~\ref{lem:angles_within_subspace}, for $\wt m_1, \wt m_2 = 1, \ldots, \wt M$,
$\wt \Up_{\wt m_1, \wt m_2} \ge \wc C$ if $\theta(\wt m_1) = \theta(\wt m_2)$, and  
$\wt \Up_{\wt m_1, \wt m_2} \le \tau_{n, K}$ if $\theta(\wt m_1) \neq \theta(\wt m_2)$.
Also, for $\om \in \Om_t$, one has $\wh{\wt \Up}_{\wt m_1, \wt m_2} \ge  {\wc C}/2$ 
if  $\theta(\wt m_1) = \theta(\wt m_2)$, and 
$\wh{\wt \Up}_{\wt m_1, \wt m_2} \le \tau_{n, K} + 2 \delta_{n, K, t}$ 
if $\theta(\wt m_1) \neq \theta(\wt m_2)$.

Now, consider matrices $G, \wh G \in \{0, 1 \}^{\wt M \times \wt M}$ with
\bes
G_{\wt m_1, \wt m_2} = I(\theta(\wt m_1) = \theta(\wt m_2)), \, \, \,
\wh G_{\wt m_1, \wt m_2} = I(| \wh{\wt \Up}_{\wt m_1, \wt m_2} | \ge T), \, \, \,
\wt m_1, \wt m_2 = 1, \ldots, \wt M
\ees
Then,  $G = \Theta \Theta^T$.  Moreover, if $n$ is large enough,    
then   $\tau_{n, K} + 2 \delta_{n, K, t} < T  < {\wc C}/2$, 
whenever $T$ satisfies conditions~\eqref{eq:thresh_cond_new}. Consequently, 
$\wh G = G$ for $\omega \in \Omega_t$ and hence, spectral clustering 
of $\wh G$ correctly recovers $M$   clusters given by $\Theta$.
\\


\subsection{Proof  of Theorem~\ref{th:SSC_WithinErrors}  } 
\label{sec:th3_proof}


The proof of this theorem is very similar to the proof of Theorem 3 in 
\cite{pensky2021clustering}.
In this proof, same as before, we  denote by $\bbC$    an absolute constant which can be different at different instances.
Consider tensors $\bG \in \RR^{n \times n \times L}$ and $\bH = \bG \tim (C D_c^{-1/2})^T  \in \RR^{n \times n \times M}$ 
with layers, respectively, $G\upl = \bG(:,:,l)$ and $H\upm = \bH(:,:,m)$  of the forms
\be \label{eq:calG}
G\upl =   (P\upl)^2, \quad H\upm = L_m^{-1/2}\, \sum_{c(l)=m} \, G\upl, \quad
l \in [L],\ m \in [M]
\ee
In order to assess $R_{WL}$, one needs to examine the spectral structure of matrices  $H\upm$ 
and their deviation from the sample-based versions $\hH \upm = \hbH (:,:,m)$.
We start with the first task.

It follows from \eqref{eq:expans1} and \eqref{eq:BDl} that   
\be \label{eq:bHm_alt}
H\upm =  U_z\upm\,  \bar{Q}_D\upm\, (U_z\upm)^T \quad \mbox{with} \quad 
\bar{Q}_D\upm = L_m^{-1/2}\, \sum_{c(l)=m} \, \lkr B_D\upl \rkr^2
\ee
Since all eigenvalues of $(B_D\upl)^2$ are positive, applying 
the Theorem in Complement 10.1.2 on page 327 of \cite{Rao_Rao_1998}
and Assumptions~{\bf A1}--{\bf A5}, obtain that
\begin{align}
\sig_{\min} (H\upm) & = \sig_{K} \lkr \bar{Q}_D\upm \rkr \geq  L_m^{-1/2}\, \sum_{c(l)=m} \sig_K \lkr (B_D\upl)^2 \rkr \nonumber \\
& \geq  L_m^{-1/2}\,  \lkr \min_k (\hn_{k}\upm)\rkr^2\ \rho_n^2\,  \sum_{c(l)=m} \sig_K \lkr (B_0\upl)^2 \rkr
%
\nonumber \\
  & 
\geq \bbC \, n^2\, \rho_n^2 \,  K^{-2}\, \sqrt{L M^{-1}} 
\label{eq:sigminHm}
\end{align}
Note that the Euclidean separation $\gamma_m$  of rows of 
 $U_{H}\upm$ is the same as the Euclidean separation of rows of $U_z\upm$,
 and $\di \gamma_m^2 \geq 2 (\min_k (\hn_k\upm))^{-1}  \geq  \bbC K/n$  for 
$\om \in  \Om_t$.

Therefore, by Lemma~9 of \cite{lei2021biasadjusted},  derive that  the total number of clustering errors $\Del$ 
within all layers is bounded as  
\bes 
\Del \leq \bbC\, \frac{n}{K} \, \sum_{m=1}^M  \, 
\left\| \sin\Te \lkr \hU_{\hH}\upm, U_{H}\upm \rkr \right\|^2_F
\ees 
Using Davis-Kahan theorem and formula  \eqref{eq:sigminHm}, obtain 
\bes
\left\| \sin\Te \lkr \hU_{\hH}\upm,  U_{H}\upm \rkr \right\|^2_F \leq 
\frac{4 K \|\hH\upm - H\upm\|^2}{\sig^2_{\min} (H\upm)}
\leq \bbC \, \frac{K^5 M\,  \|\hH\upm - H\upm\|^2}{n^4 \rhon^4 L}   
\ees  
where we use  $\bbC$ for different constants that depend  on the constants in Assumptions~{\bf A1}-{\bf A5}. 
Combination of the last two inequalities yields that the total number of clustering errors 
within all layers  is bounded by  
\be   \label{eq:err_tot_alt1}
\Del \leq  \bbC \, \frac{K^4 M}{n^3 \rhon^4 L} \,  \sum_{m=1}^M \, \|\hH\upm - H\upm\|^2
\ee  
Recall that $H\upm = [\bG \tim  \Psi^T](:,:,m)$ and $\hH\upm = [\hbG \tim \hPsi^T] (:,:,m)$. Since, by Theorem~\ref{th:SSC_ClustErrors}, for $\om \in \Om_t$ one has $\hPsi = \Psi$, obtain that 
\bes
\|\hH\upm - H\upm\|^2    \leq    L_m^{-1} \left\|\hbarG\upm - \barG\upm \right\|^2
\ees
where
\bes
\barG\upm = \displaystyle \sum_{c(l) =m}  G\upl = \sqrt{L_m}\, H\upm, \quad 
\hbarG\upm =  \sqrt{L_m}\, [\hbG \tim \Psi^T] (:,:,m) = \displaystyle \sum_{c(l) =m}  \hG\upl
\ees
use   the following lemma that modifies upper bounds in 
 \cite{lei2021biasadjusted} in the absence of  the sparsity assumption $\rhon n \leq \bbC$:  
\\

\begin{lemma} \label{lem:lei_modified}  
Let   Assumptions {\bf A1}--{\bf A5} hold,  $G\upl = (P\upl)^2$ and 
$\hG\upl = (A\upl)^2 - \diag(A\upl {\bf 1})$, where $c(l) = m$, $l=1, ..., \tilL$.
Let 
\bes 
G = \sum_{l=1}^{\tilL}\, G\upl, \quad  \hG = \sum_{l=1}^{\tilL}\,  \hG\upl 
\ees
Then, for any $t >0$, there exists a constant $\wt C$
 that depends  only on   $t$ and constants in  Assumptions {\bf A1}--{\bf A5}, and $\wt C_{t, \eps}$ which depends  only on $t$ and 
$\eps$ in Algorithm~\ref{alg:PW_within}, such that one has 
\be \label{eq:lei_modified}
\PP \lfi \|\hG - G\|^2 \leq \wt C \lkv \rhon^3 n^3 \tilL \log (\tilL + n) + \rhon^4 n^2 \tilL^2 \rkv \rfi 
\geq 1 - \wt C_{t, \eps} (\tilL + n)^{1 - t}
\ee
\\
\end{lemma}
Applying Lemma~\ref{lem:lei_modified} with $\tilL = L_m$, obtain that there exists a set $\Om_{t4}$, with  
\bes
\PP(\Om_{t4}) \geq 1 - \wt C_{t, \eps} n^{1 - t},
\ees
and for $\om \in \Om_{t4}$, one has
\be \label{eq:Del1_up}
 \|\hH\upm - H\upm\|^2    \leq \bbC\, [ \rhon^{3} n^3  \,   \log (L+n)  + \rhon^4\, n^{2} L/M]  
\ee
To complete the proof, combine formulas \eqref{eq:err_tot_alt1}  and  \eqref{eq:Del1_up}, set $\tilOm_t = \Om_t \cap \Om_{t4}$
and  recall that $R_{WL} = \Del/(Mn)$ and $n \rhon \geq C_\rho \log n$.


\subsection{Proofs of supplementary statements }   
\label{sec:suppl_proofs}


\noindent 
{\bf Proof of Lemma~\ref{lem:condA5_a}. }  
First, we prove part (a). Recall that, for $l$ with $c(l)=m$, by formula \eqref{eq:q_l}, one has
\be \label{eq:lem1_1}
\bm q^{(l)}=  \rho_n\,  (\wt U^{(m)} \otimes \wt U^{(m)})  \left( \sqrt{D^{(m)}} \otimes \sqrt{D^{(m)}} \right) \bm b_0^{(l)}
\ee 
Since $\calB_0$ is a full rank matrix, one can present $\bb_0^{(l_0)}$  as $\bb_0^{(l_0)} = \calB_0 \bw$ for some vector $\bw$.
Note that, although vectors $\bb_0^{(l)} \in \RR^{K^2}$, due to symmetry of matrices $B_0^{(l)}$, the ambient dimension
of those vectors is $K(K+1)/2 = |{\cal L}_0|$. Then, by Assumption {\bf A1}, obtain
\begin{align*} 
\|\bw\|_1  & \leq  K \|\bw\|_2  
 \leq K\, (\sigma_{min} (\calB_0))^{-1}\, \|\bb_0^{(l_0)}\| \\
& \leq K\, (\sig_{0,K})^{-1}\, \|B_0^{(l_0)}\|_F   
  \leq \bar C\, (\sig_{0,K})^{-1}\, K \sqrt{K}  
\end{align*}
Now, \eqref{eq:lem1_1} and $\bb_0^{(l_0)} = \calB_0 \bw$  imply that 
\bes
\bq^{(l_0)} = \sum_{l \in {\cal L}_0} \bq^{(l)} \, \bw_l
\ees
Therefore, 
\bes
\bx^{(l_0)} = \sum_{l \in {\cal L}_0} \bx^{(l)} \, (\bw_*)_l
\ees
where $|(\bw_*)_l| = |\bw_l| \|\bq^{(l)}\|/\|\bq^{(l_0)}\| $.
By Lemma~\ref{lem:scaling}, one has
$\|\bq^{(l)}\|/\|\bq^{(l_0)}\| \leq (\wt C_0 \,  C_{\sig,0} \, \underline{C})^{-1} \, \wt{\wt C}_0   \,\bar C$,
and, hence,
\bes
\|\bw_*\|_1 \leq \frac{ (\bar C)^2\, \wt{\wt C}_0}{\underline{C}\, \wt C_0 \,  C_{\sig,0}}\
\frac{K \sqrt{K}}{\sig_{0,K}},
\ees
which proves part (a).


Validity of part (b) follows from the fact that there are at least two copies of any vector $\bx^{(l)}$ 
for any $l$ and any group of layers. 
\\

\medskip


\noindent 
{\bf Proof of Lemma~\ref{lem:balanced_groups}. }  
For a fixed $k$, note that $\hn_k\upm   \sim \text{Binomial}(\pi_k , n)$. By Hoeffding inequality, for any $x > 0$
\bes
\PP \left \{ \left |\hn_k\upm/n  - \pi_k \right | \ge x \right \} \le 2 \exp\{-2n x^2\}
\ees
Then, using \eqref{eq:bounds}, obtain
\bes
\PP \left \{   \lowc_{\pi}\,n/K - n x \leq \hn_k\upm \leq   \highc_{\pi}\,\,n/K + n x   \right \} \ge 1- 2 \exp\{-2n x^2\}
\ees
Now, set $x=\sqrt{t \log n/(2 n)}$ and let $n$ be large enough, so that
$K\sqrt{t \log n/(2 n)} < 1/2$, which is equivalent to $t < n/(2 K^2 \log n)$.
Then, combination of the union bound over $k$ and $m$ and 
\bes
\PP \left \{   \frac{\lowc_{\pi}\,n}{K} \lkr 1 - \frac{K\, \sqrt{t \log n}}{\lowc_{\pi} \sqrt{2\, n}}  \rkr \leq 
\hn_k\upm \leq    \frac{\highc_{\pi}\,n}{K} \lkr 1 + \frac{K\, \sqrt{t \log n}}{\highc_{\pi}\sqrt{2\, n}}  \rkr  \right \} 
\geq 1- 2  n^{-t}
\ees
implies the second inequality in  \eqref{eq:bounds_prob}. The first inequality in \eqref{eq:bounds_prob} can be proved in a similar manner.\\

\medskip


\noindent
{\bf Proof of Lemma~\ref{lem:scaling}. }  
Denote $D = \diag(n_1, \ldots, n_K)$, $\wh D^{(m)} = \left ( Z^{(m)} \right )^T \left ( Z^{(m)} \right ) =
\diag(\hat n_1^{(m)}, \ldots, \hat n_K^{(m)})$,
where $n_k = n \pi_k$ and $\hat n_k^{(m)}$ are defined in \eqref{eq:hnk_hLm}. Consider matrices
\bes
U^{(m)} = Z^{(m)} \left ( \wh D^{(m)} \right )^{-1/2} \in  \calO_{n,K}, \quad
\wt U^{(m)} = (I - \scrP) U^{(m)}, \quad m=1, \ldots, M, 
\ees
where $\scrP$ is defined in  \eqref{eq:scrP}, and note that $\calS_m = \text{span} \left ( \wt U^{(m)} \otimes \wt U^{(m)} \right )$.
For $m=1, ..., M$, denote
\be
\begin{aligned} \label{eq:btm}
& \bm t =  n^{-1/2}\, \lkr \sqrt{n_1}, \ldots,  \sqrt{n_K}\rkr^T, 
\quad \hat {\bm t}^{(m)} = n^{-1/2}\, \lkr  \sqrt{\hat n_1^{(m)}}, \ldots,  \sqrt{\hat n_K^{(m)}} \rkr^T, \\
& \Pi_{\hat{\bm t}^{(m)}} =  \hat{\bm t}^{(m)} (\hat{\bm t}^{(m)})^T
\end{aligned} 
\ee
where  $\Pi_{\hat{\bm t}^{(m)}}$ are the projection matrices 
and $\Pi_{\hat{\bm t}^{(m)}}^{\perp} =   I_K - \Pi_{\hat{\bm t}^{(m)}}$.
Then, for $m=1,\ldots, M$, due to $\bm 1_n = Z^{(m)} \bm 1_K$, one has
\bes
\wt U^{(m)}   = U^{(m)} \left ( I_K - \left ( \wh D^{(m)} \right )^{1/2} 
\frac{\bm 1_K \bm 1_n^T}{n} Z^{(m)} \left ( \wh D^{(m)} \right )^{-1/2} \right ) 
\ees
Now, since $\left ( \wh D^{(m)} \right )^{1/2} \bm 1_K =  \sqrt{n}\, \hbtm$ and 
$\bm 1_n^T Z^{(m)} \left ( \wh D^{(m)} \right )^{-1/2} =  \sqrt{n}\, (\hbtm)^T$, 
one obtains
\be \label{eq:wtUm}
\wt U^{(m)} = (I - \scrP) U^{(m)}   = U^{(m)} \left ( I_K - \hat {\bm t}^{(m)} (\hat{\bm t}^{(m)})^T \right )   
= U^{(m)} \Pi_{\hat{\bm t}^{(m)}}^{\perp},
\ee 
Note that, $\Pi_{\hat{\bm t}^{(m)}}^{\perp} =  \wh V^{(m)} (\wh V^{(m)})^T$, for some matrix $\wh V^{(m)} \in \calO_{K,K-1}$. Denote
\be \label{eq:tWm}
\wt W^{(m)} =  U^{(m)} \wh V^{(m)} \in \calO_{n,K-1}, \quad m=1,\ldots, M
\ee
Hence, 
$$\wt U^{(m)} =  \wt W^{(m)} (\wh V^{(m)})^T \quad \text{and} \quad
\calS_m = \Span \left [\left ( \wt W^{(m)} \otimes \wt W^{(m)} \right )  \left ( \wh V^{(m)} \otimes \wh V^{(m)} \right )^T \right ].$$

Consider $\bm q^{(l_i)}$ with $c(l_i) = m_i$, $i=1,2$. 
Due to \eqref{eq:span1}, \eqref{eq:tilPl}--\eqref{eq:q_l} and \eqref{eq:wtUm}, obtain
\bes 
\bm q^{(l_i)} = \left (  U^{(m_i)} \otimes  U^{(m_i)} \right ) 
\left (  \Pi_{\bm t^{(m_i)}}^{\perp} \otimes  \Pi_{\bm t^{(m_i)}}^{\perp} \right )  \wt{\bm b}^{(l_i)}, \quad i = 1, 2 
\ees 
If $m_1=m_2=m$, then, due to $(U^{(m)})^T U^{(m)} = I_K$ and using  Theorem 1.2.22 in \cite{GuptaNagar1999}, obtain
\begin{align*} 
 (\bm q^{(l_1)})^T \bm q^{(l_2)}  & =  (\wt{\bm b}^{(l_1)})^T 
\left (  \Pi_{\bm t^{(m)}}^{\perp} \otimes  \Pi_{\bm t^{(m)}}^{\perp} \right )  \wt{\bm b}^{(l_2)} \\   
  &= \left [ \vect(\wt B^{(l_1)}) \right ]^T \vect \left [ \Pi_{\bm t^{(m)}}^{\perp} \wt B^{(l_2)} \Pi_{\bm t^{(m)}}^{\perp} \right ] \\
& = \Tr \left [ \wt B^{(l_1)} \Pi_{\bm t^{(m)}}^{\perp} \wt B^{(l_2)} \Pi_{\bm t^{(m)}}^{\perp} \right ]  \\
 & =   \left ( \vect(\Pi_{\bm t^{(m)}}^{\perp}) \right )^T (\wt B^{(l_1)} \otimes \wt B^{(l_2)}) \vect(\Pi_{\bm t^{(m)}}^{\perp}),   
\end{align*}
so that 
\bes
| (\bm q^{(l_1)})^T \bm q^{(l_2)}| \geq 
\sigma_{\min}(\wt B^{(l_1)}) \sigma_{\min}(\wt B^{(l_2)}) \left \| \vect(\Pi_{\bm t^{(m)}}^{\perp}) \right \|^2
\ees
Since $\wt B^{(l_i)} = \sqrt{D^{(m)}} B^{(l_i)} \sqrt{D^{(m)}}$, by Assumptions~{\bf (A1)-\bf (A3)}, one has 
\bes
 \sigma_{\min}(\wt B^{(l_i)}) \geq \sigma_{\min}( D^{(m)}) \, \sigma_{\min}( B_0^{(l_i)}) \, \rho_n  
\geq \wt C_0 C_{\sigma,0} \, \sigma_{\max}( B_0^{(l_i)}) n  \, \rho_n/K  
\ees 
and
$\left \| \vect(\Pi_{\bm t^{(m)}}^{\perp}) \right \|^2 =\left  \| \Pi_{\bm t^{(m)}}^{\perp} \right \|_F^2 = K - 1$.
Hence,  for $K \ge 2$ and $c(l_1) = c(l_2)$, one has
\be  \label{eq:lem5_1}
|(\bm q^{(l_1)})^T \bm q^{(l_2)} | \ge  (\wt C_0 \, C_{\sigma,0})^2  \,  
\sigma_{\max}( B_0^{(l_1)}) \,  \sigma_{\max}( B_0^{(l_2)})  n^2 \rho_n^2/(2K)
\ee
Using \eqref{eq:lem5_1} with $l_1 = l_2 = l$ and taking into account that 
$\sigma_{\max}(B_0^{(l)}) \geq \lowC$ by Assumption~{\bf A1}, obtain that,  
for $K \ge 2$,
\bes   
\| \bm q^{(l)} \|  \geq 
 0.5\,  \wt C_0 \, \lowC\, C_{\sigma,0}    \,  
 K^{-1/2}\, n  \rho_n 
\ees
which implies the first inequality in \eqref{eq:minF_Pl}.
On the other hand, if $l_1 = l_2 = l$, then
\be \label{eq:lem5_11} 
 \|\bm q^{(l)} \|  \leq   \sigma_{\max}(D^{(m)}) \, \rho _n \,  \sigma_{\max}(B_0^{(l)}) \, \|\vect(\Pi_{\bm t^{(m)}}^{\perp}) \| 
 \leq \wt{\wt C}_0   \,\sigma_{\max}(B_0^{(l)}) \,   n \, \rho_n \, K^{-1/2}
 \ee 
which yields the second inequality in  \eqref{eq:minF_Pl}.

In order to prove \eqref{eq:upperbound}, 
note that, due to $ \| \Pi_{(K-1)}  (  \wt A^{(l)} ) - \wt A^{(l)} \| ^2 \le  \| \wt P^{(l)} - \wt A^{(l)} \| ^2 $ and 
$\| \wt P^{(l)} - \wt A^{(l)} \|  \leq   \| P^{(l)} -  A^{(l)} \|$,
one derives
\begin{align*}
  \| \wh {\wt P}^{(l)} - \wt P^{(l)} \|_F^2 & \leq 2K \| \Pi_{(K-1)}  (  \wt A^{(l)} ) - \wt P^{(l)} \| ^2 \\
  & \le 2K \left [ 2 \| \Pi_{(K-1)}  (  \wt A^{(l)} ) - \wt A^{(l)} \| ^2 + 2 \| \wt A^{(l)} - \wt P^{(l)} \| ^2 \right ]\\  
  & \le 8\, K\,  \| P^{(l)} -  A^{(l)} \|^2 
\end{align*}
Using Theorem 5.2 of \cite{lei2015consistency}, for any $t > 0$, with probability at least $1-n^{-t}$, obtain 
$ \| P^{(l)} -  A^{(l)} \|  \le C_{t,\rho} \sqrt{n \rho_n}$, 
where $C_{t,\rho}$ depends on $C_{\rho}, \bar C$ and $t$ only. 
Hence, with probability at least $1-n^{-t}$, one has 
\bes
\|\hat{\bm q}^{(l)} - \bm q^{(l)}\| = \| \wh {\wt P}^{(l)} - \wt P^{(l)} \|_F      \leq 2 \sqrt{2}\, C_{t,\rho}  \sqrt{K\, n\, \rho_n}
\ees
Application of the union bound and \eqref{eq:minF_Pl} yields that, with probability at least $1-Ln^{-t}$, 
\bes  
\max_l\ \frac{\|\hat{\bm q}^{(l)} - \bm q^{(l)}\|}{\| \bm q^{(l)} \|}   \le 
\frac{2 \sqrt{2} C_{t,\rho} \sqrt{\rho_n K n} \, \sqrt{K}}{\wt C_0 \, \lowC\, \, C_{\sigma,0} \rho_n n},
\ees
which completes the proof.

\medskip


\noindent
{\bf Proof of Lemma~\ref{lem:inner_product_bound}. }  
Consider $\bm x \in \calS_m$ and $\bm x' \in \calS_{m'}$, where $m \neq m'$. Then
$\bm x = \left ( \wt W^{(m)} \otimes \wt W^{(m)} \right ) \bm v$,    
where $\bm v \in \RR^{(K-1)^2}$ and $\wt W^{(m)}$ is defined in  \eqref{eq:tWm}, and
\bes
\| \bm x \|^2= \bm v^T \left ( \left ( \wt W^{(m)} \right )^T  \wt W^{(m)}  \otimes \left ( \wt W^{(m)} \right )^T  \wt W^{(m)}  \right )  \bm v  = \| \bm v \|^2,
\ees
Similarly, $\bm x' = \left ( \wt W^{(m')} \otimes \wt W^{(m')} \right ) \bm v'$,
where $\bm v' \in \RR^{(K-1)^2}$ and $\| \bm x' \| = \| \bm v' \|$. 
Then, using the Cauchy-Schwarz inequality, obtain
\bes
| \bm x^T \bm x' |  \leq \|\bm x\| \left \|  \left ( \wt W^{(m)} \right )^T  \wt W^{(m')}  
\otimes \left ( \wt W^{(m)} \right )^T  \wt W^{(m')}  \right \|  \|\bm x' \|
\ees
Since $\wt W^{(m)} = \wt U^{(m)} \wh V^{(m)}$ and $\wh V^{(m)} \in \calO_{K,K-1}$, 
$m \in [M]$, it is easy to see that
\begin{align*}
\left \|  \left ( \wt W^{(m)} \right )^T  \wt W^{(m')}  \otimes \left ( \wt W^{(m)} \right )^T  \wt W^{(m')}  \right \|  
&= \left \| \left ( \wt W^{(m)} \right )^T  \wt W^{(m')} \right \|^2\\
&\leq \left \| \left ( \wt U^{(m)} \right )^T  \wt U^{(m')} \right \|^2
\end{align*}
Therefore, if $\bm x \in \calS_m, \, \bm x' \in \calS_{m'},$ and $\|\bm x \| = \|\bm x' \| = 1$, $m \ne m'$, then
\be \label{eq:sc_prod}
|\bm x^T \bm x' | \le \left \| \left ( \wt U^{(m)} \right )^T  \wt U^{(m')} \right \|^2
\ee
In order to derive an upper bound for \eqref{eq:sc_prod} when $m \ne m'$, note that matrix $\wt U^{(m)}$, defined in 
\eqref{eq:wtUm},  has elements 
\bes 
\wt U_{j,k}^{(m)} =  (\hat n_k^{(m)})^{-1/2}\, \lkv  I(\xi_j^{(m)} = k) -  
n^{-1}\, \sum_{i=1}^{n}  I(\xi_i^{(m)} = k) \rkv
\quad \mbox{with} \quad \sum_{j=1}^{n} \wt U_{j,k}^{(m)} = 0
\ees 
Rows of matrix $\wt U^{(m)}$ are identically distributed but not independent, 
which makes the analysis difficult. For this reason, we consider proxies $\wt {\wt U}^{(m)}$ for $\wt U^{(m)}$ with elements 
\bes
\wt{\wt U}_{j,k}^{(m)} = \frac{1}{\sqrt{n_k}} I(\xi_j^{(m)} = k) - \frac{\sqrt{n_k}}{n} \equiv   \frac{1}{\sqrt{n \pi_k}} 
\left [ I(\xi_j^{(m)} = k) - \pi_k \right ],\ j \in [n], k \in [K]
\ees
so that $E \wt{\wt U}_{j,k}^{(m)} = 0$. Rows of $\wt{\wt U}^{(m)}$ are i.i,d, 
and also $\wt{\wt U}^{(m)}$ and $\wt{\wt U}^{(m')}$ are independent when $m \neq m'$. 
Hence, matrices $\wt{\wt U}^{(m)}$ are i.i.d with $E \wt{\wt U}^{(m)} = 0$. 
We shall use the following statement, proved later in Section~\ref{sec:suppl_proofs}.
\\

\begin{lemma} \label{lem:U_bound} 
Let $\bar {\bs{\pi}}  = (\pi_1, \ldots, \pi_K)$ be such that $\pi_k \ge  \lowc_{\pi}/K$  for $k=1, \ldots, K$. 
Then, there exists a set $\wt \Omega_{t}$ with $\PP (\wt \Omega_{t} ) \ge 1- 2 K M^2 n^{-t}$ 
such that, for any $\omega \in \wt \Omega_{t}$,
\be \label{eq:U_bound}
\max_{\substack{1 \le m_1, m_2 \le M \\ m_1 \neq m_2}} \left \| \left (\wt{\wt U}^{(m_1)} \right )^T \wt{\wt U}^{(m_2)} \right \|  
\leq \frac{2 K \sqrt{t \log n}}{\sqrt{n}}\\
\ee
\end{lemma} 

\medskip  

\noindent
In order to obtain an upper bound for \eqref{eq:sc_prod} when $m \ne m'$, use the fact that proxies 
$\wt {\wt U}^{(m)}$ are close to $\wt U^{(m)}$. Indeed, the following statement is valid.

\begin{lemma} \label{lem:U_dif_bound}  
Let $\bar {\bs{\pi}} = (\pi_1, \ldots, \pi_K)$ be such that $\pi_k \ge  \lowc_{\pi}/K$, $k=1, \ldots, K$. 
Then, there exists a set $\wt{\wt \Omega}_t$ with $\PP (\wt{\wt \Omega}_t) \ge 1- 2 K M n^{-t}$ such that, 
for any $\omega \in \wt{\wt \Omega}_t$, one has
\be \label{eq:U_dif_bound}
\Del \equiv \max_{1 \le m \le M} \left \| \wt{\wt U}^{(m)} - \wt U^{(m)} \right \|  \le \frac{K\, \sqrt{2 t \log n}}{\lowc_{\pi}\, \sqrt{n}} \\
\ee
\\
\end{lemma} 

\medskip

\noindent
Then, due to 
$$\left \| \wt U^{(m)} \right \|  = \left \| (I - \scrP) U^{(m)} \right \|  \leq 1 \quad
\text{and} \quad \left \| \wt{\wt U}^{(m)} \right \|  \le \left \| \wt U^{(m)} \right \|  + \left \| \wt{\wt U}^{(m)} - \wt U^{(m)}  \right \|,$$
derive for any $m_1, m_2$
\begin{align*} 
\left \|\left (\wt U^{(m_1)} \right )^T \wt U^{(m_2)}\right \| & \le \left \|\left (\wt{\wt U}^{(m_1)} \right )^T \wt{\wt U}^{(m_2)} \right \|  
+ \left \| \left [ \wt{\wt U}^{(m_1)} - \wt U^{(m_1)} \right ]^T \wt{\wt U}^{(m_2)}  \right \|  \\
& + \left \| \left (\wt U^{(m_1)} \right )^T \left [ \wt{\wt U}^{(m_2)} - \wt U^{(m_2)} \right ] \right \| \\
& \leq   \left \| \left (\wt{\wt U}^{(m_1)} \right )^T \wt{\wt U}^{(m_2)} \right \|  + \Delta (1+ \Delta) + \Delta
\end{align*}
Now, let $\check \Omega_t = \wt \Omega_t \cap \wt{\wt \Omega}_t$. 
Note that $\Delta < 1$ for $n$ large enough. Then,
$\PP (\check \Omega_t ) \ge 1- 2 K M (M+1) n^{- t}$ 
and, for $\omega \in \check \Omega_t$, one has
\bes
\max_{m \ne m'} \left \| \left (\wt U^{(m)} \right )^T \wt U^{(m')} \right \|  
\leq \frac{K \sqrt{2 t \log n}}{\sqrt{n}} \left [ \sqrt{2} + \frac{3}{\lowc_{\pi}} \right ]
\ees
which completes the proof.
\\

\medskip \medskip

 
\noindent
{\bf Proof of Lemma~\ref{lem:angles_within_subspace}. }  
\ignore{
Consider $\bm q^{(l_i)}$ with $c(l_i) = m_i$, $i=1,2$. 
Due to \eqref{eq:span1}, \eqref{eq:tilPl}--\eqref{eq:q_l} and \eqref{eq:wtUm}, obtain
\bes 
\bm q^{(l_i)} = \left (  U^{(m_i)} \otimes  U^{(m_i)} \right ) 
\left (  \Pi_{\bm t^{(m_i)}}^{\perp} \otimes  \Pi_{\bm t^{(m_i)}}^{\perp} \right )  \wt{\bm b}^{(l_i)}, \quad i = 1, 2 
\ees 
If $m_1=m_2=m$, then, due to $(U^{(m)})^T U^{(m)} = I_K$ and using  Theorem 1.2.22 in \cite{GuptaNagar1999}, obtain
\begin{align*} 
 (\bm q^{(l_1)})^T \bm q^{(l_2)}  & =  (\wt{\bm b}^{(l_1)})^T 
\left (  \Pi_{\bm t^{(m)}}^{\perp} \otimes  \Pi_{\bm t^{(m)}}^{\perp} \right )  \wt{\bm b}^{(l_2)}    
  = \left [ \vect(\wt B^{(l_1)}) \right ]^T \vect \left [ \Pi_{\bm t^{(m)}}^{\perp} \wt B^{(l_2)} \Pi_{\bm t^{(m)}}^{\perp} \right ] \\
& = \Tr \left [ \wt B^{(l_1)} \Pi_{\bm t^{(m)}}^{\perp} \wt B^{(l_2)} \Pi_{\bm t^{(m)}}^{\perp} \right ]  
  =   \left ( \vect(\Pi_{\bm t^{(m)}}^{\perp}) \right )^T (\wt B^{(l_1)} \otimes \wt B^{(l_2)}) \vect(\Pi_{\bm t^{(m)}}^{\perp}),   
\end{align*}
so that 
\bes
| (\bm q^{(l_1)})^T \bm q^{(l_2)}| \geq 
\sigma_{\min}(\wt B^{(l_1)}) \sigma_{\min}(\wt B^{(l_2)}) \left \| \vect(\Pi_{\bm t^{(m)}}^{\perp}) \right \|^2
\ees
Since $\wt B^{(l_i)} = \sqrt{D^{(m)}} B^{(l_i)} \sqrt{D^{(m)}}$, by Assumptions~{\bf (A1)-\bf (A3)}, one has 
\bes
 \sigma_{\min}(\wt B^{(l_i)}) \geq \sigma_{\min}( D^{(m)}) \, \sigma_{\min}( B_0^{(l_i)}) \, \rho_n  
\geq \wt C_0 C_{\sigma,0} \, \sigma_{\max}( B_0^{(l_i)}) n  \, \rho_n/K  
\ees 
and
$\left \| \vect(\Pi_{\bm t^{(m)}}^{\perp}) \right \|^2 =\left  \| \Pi_{\bm t^{(m)}}^{\perp} \right \|_F^2 = K - 1$.
Hence,  for $K \ge 2$,
\be  \label{eq:lem5_1}
|(\bm q^{(l_1)})^T \bm q^{(l_2)} | \ge  (\wt C_0 \, C_{\sigma,0})^2  \,  
\sigma_{\max}( B_0^{(l_1)}) \,  \sigma_{\max}( B_0^{(l_2)})  n^2 \rho_n^2/(2K)
\ee
which implies the first inequality in \eqref{eq:minF_Pl}.
On the other hand, if $l_1 = l_2 = l$, then
\be \label{eq:lem5_11} 
 \|\bm q^{(l)} \|  \leq   \sigma_{\max}(D^{(m)}) \, \rho _n \,  \sigma_{\max}(B_0^{(l)}) \, \|\vect(\Pi_{\bm t^{(m)}}^{\perp}) \| 
  \leq \wt{\wt C}_0   \,\sigma_{\max}(B_0^{(l)}) \,   n \, \rho_n \, K^{-1/2}   
\ee 
which yields the second inequality in  \eqref{eq:minF_Pl}.
} 
In addition,  the last inequality and \eqref{eq:lem5_1} imply that 
\be
|(\bm x^{(l_1)})^T \bm x^{(l_2)} | = \frac{|(\bm q^{(l_1)})^T \bm q^{(l_2)} | }{\|\bm q^{(l_1)} \| \|\bm q^{(l_2)} \|} \ge 
\frac{(\wt C_0 \, C_{\sigma,0})^2 }{2 (\wt{\wt C}_0)^2},
\ee
which completes the proof of the first inequality in \eqref{eq:scalar_products_x}.
The second inequality in \eqref{eq:scalar_products_x} is true by    by {\bf A5}.

 
To prove  \eqref{eq:scalar_products_y}, note
 that, for any $l_1$ and $l_2$, by the Cauchy-Schwarz inequality and \eqref{eq:y-x}, one has
\begin{align*}
 |(\bm y^{(l_1)})^T \bm y^{(l_2)} - (\bm x^{(l_1)})^T \bm x^{(l_2)}|   
& \leq |(\bm y^{(l_1)})^T \lkv \bm y^{(l_2)} - \bm x^{(l_2)} \rkv | + |\lkv \bm y^{(l_1)} - \bm x^{(l_1)}\rkv^T \bm x^{(l_2)} | \\
  & \leq 2 \max_{1 \le l \le L} \| \bm y^{(l)} - \bm x^{(l)} \|   
  \le 2 \delta_{n, K, t} 
\end{align*}
for $\omega \in \Omega_t$, where $\Om_t$ and $\delta_{n, K, t}$ are defined, respectively, in Theorem~\ref{th:self_express} 
and \eqref{eq:delnk}. 
Then, using  \eqref{eq:scalar_products_x},  for $c(l_1) = c(l_2) = m$ and   $\omega \in \Omega_t$, obtain
\bes
\min_{\substack{l_1, l_2}} |(\bm y^{(l_1)})^T \bm y^{(l_2)} | \ge |(\bm x^{(l_1)})^T \bm x^{(l_2)} | - 2 \delta_{n, K, t} \ge  \wc C/2
\ees
if $n$ is large enough, due to $\delta_{n, K, t} \to 0$ as $n \to \infty$.
If $c(l_1) \neq c(l_2)$, then, again by \eqref{eq:scalar_products_x},
for $\omega \in \Omega_t$, derive
\bes
\max_{\substack{l_1, l_2}} |(\bm y^{(l_1)})^T \bm y^{(l_2)} | \leq   \tau_{n,K} + 2 \delta_{n, K, t}
\ees
which completes the proof.
 
\medskip \medskip


\noindent
{\bf Proof of Lemma~\ref{lem:U_bound}. }  
Note that $\wt{\wt U}^{(m)}$ are i.i.d. for $m=1,\ldots,M,$ so, for simplicity,  we can consider $m=1,2$.
Let $S= \left (\wt{\wt U}^{(1)} \right )^T \wt{\wt U}^{(2)} \in \RR^{K \times K}$. 
Since $\wt{\wt U}^{(1)}$ and $\wt{\wt U}^{(2)}$ are independent and $\EE \left ( \wt{\wt U}^{(m)} \right ) = 0$,  obtain $\EE S =0$. 
Now let $ \bm{u}_j^{(m)} = \wt{\wt U}^{(m)}(j,:)$ be the $j$-th row of $\wt{\wt U}^{(m)}$, $j=1,\ldots,n$. Then,
\bes 
 S= \sum_{j=1}^{n} S^{(j)}, \quad S^{(j)} = \lkr \bm{u}_j^{(1)}\rkr^T \bm{u}_j^{(2)} \in \RR^{K \times K}, \quad  j=1, \ldots, n
\ees
Note that $S^{(j)}$ are independent, $\EE S^{(j)} = 0,$ and $\rank(S^{(j)}) =1$. Hence,
$\|S^{(j)}\|  = \|S^{(j)}\|_{F}= \| \bm{u}_j^{(1)} \| \| \bm{u}_j^{(2)} \|$.
 Also, note that,  due to $ \sum_{k=1}^{K} I(\xi_j^{(m)}=k) = 1$ and $1/\pi_k  \leq   K/\lowc_{\pi}$, one has
\bes
   \| \bm{u}_j^{(m)} \|^2 = \sum_{k=1}^{K} \left [ \wt{\wt U}^{(m)}_{j,k}    \right ]^2 
= \sum_{k=1}^{K} \frac{1}{n_k} \left [ I(\xi_j^{(m)}=k) - \pi_k \right ]^2 
  \le \frac{K}{\lowc_{\pi} n}  
\ees
Hence, $\|S^{(j)}\|  \le   K/(\lowc_{\pi} n)$.

Now, we are going to apply matrix Bernstein inequality to matrix $S$. Observe that 
\be  \label{eq:SST}
 \EE (S^T S) = \EE (S S^T) = \sum_{j=1}^{n} \EE \left ( S^{(j)} (S^{(j)})^T \right )    
\ee
where $\EE \left ( S^{(j)} (S^{(j)})^T \right ) = \EE \left ( [\bm u_j^{(1)}]^T \bm u_j^{(1)} \right ) \EE \left \| \bm u_j^{(2)} \right \|^2$.
Therefore, 
\bes
\left \| \EE \left ( S^{(j)} (S^{(j)})^T \right ) \right \| 
= \EE \left \| \bm u_j^{(2)} \right \|^2  \left \| \EE \left ( [\bm u_j^{(1)}]^T \bm u_j^{(1)} \right ) \right \|
\ees  
Since the operator norm  is a convex function, by Jensen inequality and due to $\rank \left ( [\bm u_j^{(1)}]^T \bm u_j^{(1)}  \right ) = 1$,
obtain
\bes
 \left \| \EE \left ( [\bm u_j^{(1)}]^T \bm u_j^{(1)} \right ) \right \|  
\leq  \EE \left  \|   [\bm u_j^{(1)}]^T \bm u_j^{(1)}  \right \|  = \EE \left  \|   \bm u_j^{(1)}  \right \| ^2
\ees
On the other hand, it is easy to show that, for any $m$, one has 
$ \EE \left \|  \bm u_j^{(m)} \right \|^2 \leq  K/n$.
Therefore,  
$\left \| \EE \left ( S^{(j)} (S^{(j)})^T \right ) \right \|  \le  n^{-2}\,  K^2$, so that $ \|\EE(SS^T) \| \leq n^{-1}\, K^2$.
Now applying Theorem~1.6.2 (matrix Bernstein inequality) in \cite{Tropp2012UserFriendlyTF},
derive that, for any $x>0$, one has
\be \label{eq:m_Bern}
\PP \left\{ \| S  \|  > x \right \} \leq 2 K\,  \exp \left\{- \frac{x^2/2}{n^{-1}\, K^2 + n^{-1}\,K x/(3\, \lowc_{\pi})} \right\} 
\ee 
For any $t>0$, setting $x= 2K\, n^{-1/2}\, \sqrt{t \log n}$ ensures that, for $n$ large enough,  the denominator of the exponent in \eqref{eq:m_Bern}
is bounded above by $2 K^2\, n^{-1}$.  Then, for any $m_1, m_2 = 1,\ldots, M$, obtain
\be
\begin{aligned} \label{eq:U_bound_prob}
 \PP \left ( \|S\|  \geq  2 K\, n^{-1/2}\,  \sqrt{t \log n} \right ) &= 
\PP \left ( \left \|\left (\wt{\wt U}^{(m_1)} \right )^T \wt{\wt U}^{(m_2)} \right \|  \geq 2 K\, n^{-1/2}\,  \sqrt{t \log n} \right )\\ 
& \leq 2 K n^{-t}
\end{aligned}
\ee 
To complete the proof, apply the union bound to \eqref{eq:U_bound_prob} and let $\wt \Omega_{t}$ be the set where
this union bound holds.

\medskip \medskip


 \noindent
{\bf Proof of Lemma~\ref{lem:U_dif_bound}. }  
Since $\wt U^{(m)}$ and $\wt{\wt U}^{(m)}$ are i.i.d for every $m$, for simplicity, we drop the index $m$. By definition, 
for $k=1,\ldots, K$, one has 
\bes
\wt U(:,k) = U(:,k) -  n^{-1/2}\, \bm 1_n \cdot \hat {\bm t}_k, \quad 
 \wt {\wt U}(:,k) =  U(:,k) \,\sqrt{\hat n_k}/\sqrt{n_k}  - n^{-1/2}\, \bm 1_n \cdot   {\bm t}_k.
\ees
Hence,
\bes 
\wt U = U - n^{-1/2}\, \bm 1_n  \hat{\bm t}^T, \quad
\wt {\wt U} = U \Lambda - n^{-1/2}\, \bm 1_n {\bm t}^T, \, \, \mbox{with} \, \,  
\Lambda = \diag \left ( \frac{\sqrt{\hat n_1}}{\sqrt{n_1}}, \ldots, \frac{\sqrt{\hat n_K}}{\sqrt{n_K}}  \right ), 
\ees
where $\hat{\bm t}$ and ${\bm t}$ are defined in \eqref{eq:btm}. 
Then,
\begin{align}   \label{eq:Utt_Ut_bound}
  \left \| \wt {\wt U} - \wt U \right \|  & \leq \| U ( \Lambda - I) \|  + 
 n^{-1/2}\,  \left \|  \bm 1_n (\hat {\bm t} - \bm t)^T  \right \|  
   \leq \|  I - \Lambda \|  + \left \| \hat{\bm t} - \bm t \right \|\\
   & = \max_{1 \le k \le K} \left |1 - \frac{\sqrt{\hat n_k}}{\sqrt{n_k}} \right | + 
\left [ \sum_{k=1}^{K} \frac{(\sqrt{\hat n_k} - \sqrt{n_k})^2}{n} \right ]^{1/2} 
 \nonumber 
\end{align}
Since,  for $a, b > 0$, one has $| \sqrt{a} - \sqrt{b} | \le  |a-b|/\sqrt{b} $, and 
$n_k=n \pi_k \ge  \lowc_{\pi} n/K$, one can easily show that
\bes 
\left |1 - \frac{\sqrt{\hat n_k}}{\sqrt{n_k}} \right | \le \frac{K}{\lowc_{\pi} n} | \hat n_k - n_k |, \quad
\sum_{k=1}^{K} \frac{(\sqrt{\hat n_k} - \sqrt{n_k})^2}{n} \le \frac{K}{\lowc_{\pi} n^2} \sum_{k=1}^{K} (\hat n_k - n_k)^2
\ees
Now, recall that  $\hat n_k  =  \sum_{j=1}^{n} I(\xi_j = k)$ and $\EE(\hat n_k) = n_k$, and, using Hoeffding inequality, for any $x > 0$, obtain
\be 
\PP \left \{ \left | \hat n_k - n_k  \right | \ge n\, x \right \} \le 2 \exp{\{-2 n x^2\}}
\ee
For any $t > 0$, setting $x= \sqrt{t \log n/(2 n)}$ and taking the union bound, derive
\be \label{eq:n_k_union_prob}
\PP \left \{ \max_{\substack{1 \le m \le M \\ 1 \le k \le K}} \left | \frac{\hat n_k^{(m)} - n_k^{(m)}}{n} \right | 
\leq \sqrt{\frac{t \log n}{2 n}} \right \} \ge 1 - 2 K M n^{-t}
\ee
Now let $\wt{\wt \Omega}_t$ be the set where \eqref{eq:n_k_union_prob} holds. Then for $\omega \in \wt{\wt \Omega}_t$, one has
\be \label{eq:I_Lambda_bound}
\| I - \Lambda \|  \le \frac{K}{\lowc_{\pi}} \sqrt{\frac{t \log n}{2 n}}, \quad
\|\hat{\bm t} - \bm t \| \le \sqrt{\frac{K^2}{\lowc_{\pi}}} \sqrt{\frac{t \log n}{2 n}}
\ee
Finally, combining \eqref{eq:Utt_Ut_bound} and \eqref{eq:I_Lambda_bound}, for $\omega \in \wt{\wt \Omega}_t$,  we arrive at 
\bes 
\max_{1 \le m \le M} \left \| \wt {\wt U}^{(m)} - \wt U^{(m)} \right \|  \le  n^{-1/2}\, K \,  \sqrt{2 t \log n}/\lowc_{\pi} 
\ees
which completes the proof.




\bibliographystyle{Chicago}

\bibliography{References}

\begin{thebibliography}{}

\bibitem[\protect\citeauthoryear{Abbe}{Abbe}{2018}]{JMLR:v18:16-480}
Abbe, E. (2018).
\newblock Community detection and stochastic block models: Recent developments.
\newblock {\em J. Mach. Learn. Res.\/}~{\em 18\/}(177), 1--86.

\bibitem[\protect\citeauthoryear{Bhattacharyya and Chatterjee}{Bhattacharyya
  and Chatterjee}{2020}]{bhattacharyya2020general}
Bhattacharyya, S. and S.~Chatterjee (2020).
\newblock General community detection with optimal recovery conditions for
  multi-relational sparse networks with dependent layers.
\newblock {\em ArXiv:2004.03480\/}.

\bibitem[\protect\citeauthoryear{Bickel, Choi, Chang, and Zhang}{Bickel
  et~al.}{2013}]{10.1214/13-AOS1124}
Bickel, P., D.~Choi, X.~Chang, and H.~Zhang (2013).
\newblock {Asymptotic normality of maximum likelihood and its variational
  approximation for stochastic blockmodels}.
\newblock {\em The Annals of Statistics\/}~{\em 41\/}(4), 1922 -- 1943.

\bibitem[\protect\citeauthoryear{Bickel and Chen}{Bickel and
  Chen}{2009}]{doi:10.1073/pnas.0907096106}
Bickel, P.~J. and A.~Chen (2009).
\newblock A nonparametric view of network models and newman–girvan and other
  modularities.
\newblock {\em Proceedings of the National Academy of Sciences\/}~{\em
  106\/}(50), 21068--21073.

\bibitem[\protect\citeauthoryear{Chi, Gaines, Sun, Zhou, and Yang}{Chi
  et~al.}{2020}]{JMLR:v21:18-155}
Chi, E.~C., B.~J. Gaines, W.~W. Sun, H.~Zhou, and J.~Yang (2020).
\newblock Provable convex co-clustering of tensors.
\newblock {\em Journal of Machine Learning Research\/}~{\em 21\/}(214), 1--58.

\bibitem[\protect\citeauthoryear{De~Domenico, Nicosia, Arenas, and
  Latora}{De~Domenico et~al.}{2015}]{de2015structural}
De~Domenico, M., V.~Nicosia, A.~Arenas, and V.~Latora (2015).
\newblock Structural reducibility of multilayer networks.
\newblock {\em Nature communications\/}~{\em 6\/}(1), 1--9.

\bibitem[\protect\citeauthoryear{{Elhamifar} and {Vidal}}{{Elhamifar} and
  {Vidal}}{2009}]{Vidal:2009aa}
{Elhamifar}, E. and R.~{Vidal} (2009).
\newblock Sparse subspace clustering.
\newblock In {\em 2009 IEEE Conference on Computer Vision and Pattern
  Recognition}, pp.\  2790--2797.

\bibitem[\protect\citeauthoryear{Elhamifar and Vidal}{Elhamifar and
  Vidal}{2013}]{Elhamifar:2013:SSC:2554063.2554078}
Elhamifar, E. and R.~Vidal (2013).
\newblock Sparse subspace clustering: Algorithm, theory, and applications.
\newblock {\em IEEE Trans. Pattern Anal. Mach. Intell.\/}~{\em 35\/}(11),
  2765--2781.

\bibitem[\protect\citeauthoryear{Fan, Pensky, Yu, and Zhang}{Fan
  et~al.}{2022}]{fan2021alma}
Fan, X., M.~Pensky, F.~Yu, and T.~Zhang (2022).
\newblock Alma: Alternating minimization algorithm for clustering mixture
  multilayer network.
\newblock {\em Journal of Machine Learning Research\/}~{\em 23\/}(330), 1--46.

\bibitem[\protect\citeauthoryear{Greene and Cunningham}{Greene and
  Cunningham}{2013}]{greene2013producing}
Greene, D. and P.~Cunningham (2013).
\newblock Producing a unified graph representation from multiple social network
  views.
\newblock In {\em Proceedings of the 5th annual ACM web science conference},
  pp.\  118--121.

\bibitem[\protect\citeauthoryear{Gupta and Nagar}{Gupta and
  Nagar}{1999}]{GuptaNagar1999}
Gupta, A. and D.~Nagar (1999).
\newblock {\em {Matrix Variate Distributions}}.
\newblock Chapman and Hall/CRC.

\bibitem[\protect\citeauthoryear{Han, Luo, Wang, and Zhang}{Han
  et~al.}{2021}]{han2021exact}
Han, R., Y.~Luo, M.~Wang, and A.~R. Zhang (2021).
\newblock Exact clustering in tensor block model: Statistical optimality and
  computational limit.
\newblock {\em ArXiv:2012.09996\/}.

\bibitem[\protect\citeauthoryear{Jing, Li, Lyu, and Xia}{Jing
  et~al.}{2020}]{jing2020community}
Jing, B.-Y., T.~Li, Z.~Lyu, and D.~Xia (2020).
\newblock Community detection on mixture multi-layer networks via regularized
  tensor decomposition.
\newblock {\em arXiv preprint arXiv:2002.04457\/}.

\bibitem[\protect\citeauthoryear{Jing, Li, Lyu, and Xia}{Jing
  et~al.}{2021}]{TWIST-AOS2079}
Jing, B.-Y., T.~Li, Z.~Lyu, and D.~Xia (2021).
\newblock {Community detection on mixture multilayer networks via regularized
  tensor decomposition}.
\newblock {\em The Annals of Statistics\/}~{\em 49\/}(6), 3181 -- 3205.

\bibitem[\protect\citeauthoryear{Karrer and Newman}{Karrer and
  Newman}{2011}]{Karrer2011StochasticBA}
Karrer, B. and M.~E.~J. Newman (2011).
\newblock Stochastic blockmodels and community structure in networks.
\newblock {\em Physical review. E, Statistical, nonlinear, and soft matter
  physics\/}~{\em 83}, 016107.

\bibitem[\protect\citeauthoryear{Kivel{\"a}, Arenas, Barthelemy, Gleeson,
  Moreno, and Porter}{Kivel{\"a} et~al.}{2014}]{kivela2014multilayer}
Kivel{\"a}, M., A.~Arenas, M.~Barthelemy, J.~P. Gleeson, Y.~Moreno, and M.~A.
  Porter (2014).
\newblock Multilayer networks.
\newblock {\em Journal of complex networks\/}~{\em 2\/}(3), 203--271.

\bibitem[\protect\citeauthoryear{Le and Levina}{Le and
  Levina}{2015}]{Le2015EstimatingTN}
Le, C.~M. and E.~Levina (2015).
\newblock Estimating the number of communities in networks by spectral methods.
\newblock {\em ArXiv:1507.00827\/}.

\bibitem[\protect\citeauthoryear{Lei, Chen, and Lynch}{Lei
  et~al.}{2019}]{10.1093/biomet/asz068}
Lei, J., K.~Chen, and B.~Lynch (2019, 12).
\newblock {Consistent community detection in multi-layer network data}.
\newblock {\em Biometrika\/}~{\em 107\/}(1), 61--73.

\bibitem[\protect\citeauthoryear{Lei and Lin}{Lei and
  Lin}{2021}]{lei2021biasadjusted}
Lei, J. and K.~Z. Lin (2021).
\newblock Bias-adjusted spectral clustering in multi-layer stochastic block
  models.
\newblock {\em ArXiv:2003.08222\/}.

\bibitem[\protect\citeauthoryear{Lei and Rinaldo}{Lei and
  Rinaldo}{2015}]{lei2015consistency}
Lei, J. and A.~Rinaldo (2015).
\newblock Consistency of spectral clustering in stochastic block models.
\newblock {\em The Annals of Statistics\/}~{\em 43\/}(1), 215--237.

\bibitem[\protect\citeauthoryear{Liu, Lin, and Yu}{Liu
  et~al.}{2010}]{Liu2010RobustSS}
Liu, G., Z.~Lin, and Y.~Yu (2010).
\newblock Robust subspace segmentation by low-rank representation.
\newblock In {\em Proceedings of the 27th International Conference on
  International Conference on Machine Learning}, ICML'10, USA, pp.\  663--670.
  Omnipress.

\bibitem[\protect\citeauthoryear{Lorrain and White}{Lorrain and
  White}{1971}]{doi:10.1080/0022250X.1971.9989788}
Lorrain, F. and H.~C. White (1971).
\newblock Structural equivalence of individuals in social networks.
\newblock {\em The Journal of Mathematical Sociology\/}~{\em 1\/}(1), 49--80.

\bibitem[\protect\citeauthoryear{MacDonald, Levina, and Zhu}{MacDonald
  et~al.}{2020}]{macdonald2020latent}
MacDonald, P.~W., E.~Levina, and J.~Zhu (2020).
\newblock Latent space models for multiplex networks with shared structure.
\newblock {\em arXiv preprint arXiv:2012.14409\/}.

\bibitem[\protect\citeauthoryear{MacDonald, Levina, and Zhu}{MacDonald
  et~al.}{2021}]{macdonald2021latent}
MacDonald, P.~W., E.~Levina, and J.~Zhu (2021).
\newblock Latent space models for multiplex networks with shared structure.
\newblock {\em ArXiv:2012.14409\/}.

\bibitem[\protect\citeauthoryear{Mairal, Bach, Ponce, Sapiro, Jenatton, and
  Obozinski}{Mairal et~al.}{2014}]{mairal2014spams}
Mairal, J., F.~Bach, J.~Ponce, G.~Sapiro, R.~Jenatton, and G.~Obozinski (2014).
\newblock Spams: A sparse modeling software, v2.3.
\newblock {\em URL http://spams-devel. gforge. inria. fr/downloads. html\/}.

\bibitem[\protect\citeauthoryear{Nasihatkon and Hartley}{Nasihatkon and
  Hartley}{2011}]{5995679}
Nasihatkon, B. and R.~Hartley (2011).
\newblock Graph connectivity in sparse subspace clustering.
\newblock In {\em CVPR 2011}, pp.\  2137--2144.

\bibitem[\protect\citeauthoryear{Noroozi and Pensky}{Noroozi and
  Pensky}{2022}]{noroozi2021hierarchy}
Noroozi, M. and M.~Pensky (2022).
\newblock The hierarchy of block models.
\newblock {\em Sankhya A\/}~{\em 84}, 64--107.

\bibitem[\protect\citeauthoryear{Noroozi, Pensky, and Rimal}{Noroozi
  et~al.}{2021}]{noroozi2021sparse}
Noroozi, M., M.~Pensky, and R.~Rimal (2021).
\newblock Sparse popularity adjusted stochastic block model.
\newblock {\em Journal of Machine Learning Research\/}~{\em 22\/}(193), 1--36.

\bibitem[\protect\citeauthoryear{Noroozi, Rimal, and Pensky}{Noroozi
  et~al.}{2021}]{noroozi2021estimation}
Noroozi, M., R.~Rimal, and M.~Pensky (2021).
\newblock Estimation and clustering in popularity adjusted block model.
\newblock {\em Journal of the Royal Statistical Society: Series B (Statistical
  Methodology)\/}~{\em 83\/}(2), 293--317.

\bibitem[\protect\citeauthoryear{Paul and Chen}{Paul and Chen}{2016}]{paul2016}
Paul, S. and Y.~Chen (2016).
\newblock Consistent community detection in multi-relational data through
  restricted multi-layer stochastic blockmodel.
\newblock {\em Electron. J. Statist.\/}~{\em 10\/}(2), 3807--3870.

\bibitem[\protect\citeauthoryear{Paul and Chen}{Paul and
  Chen}{2020a}]{paul2020spectral}
Paul, S. and Y.~Chen (2020a).
\newblock Spectral and matrix factorization methods for consistent community
  detection in multi-layer networks.
\newblock {\em The Annals of Statistics\/}~{\em 48\/}(1), 230--250.

\bibitem[\protect\citeauthoryear{Paul and Chen}{Paul and
  Chen}{2020b}]{paul2020}
Paul, S. and Y.~Chen (2020b, 02).
\newblock Spectral and matrix factorization methods for consistent community
  detection in multi-layer networks.
\newblock {\em Ann. Statist.\/}~{\em 48\/}(1), 230--250.

\bibitem[\protect\citeauthoryear{Pensky and Wang}{Pensky and
  Wang}{2021}]{pensky2021clustering}
Pensky, M. and Y.~Wang (2021).
\newblock Clustering of diverse multiplex networks.
\newblock {\em arXiv preprint arXiv:2110.05308\/}.

\bibitem[\protect\citeauthoryear{Rao and Rao}{Rao and Rao}{1998}]{Rao_Rao_1998}
Rao, C. and M.~Rao (1998).
\newblock {\em Matrix Algebra and its Applications to Statistics and
  Econometrics\/} (1st ed.).
\newblock World Scientific Publishing Co.

\bibitem[\protect\citeauthoryear{Sengupta and Chen}{Sengupta and
  Chen}{2018}]{RePEc:bla:jorssb:v:80:y:2018:i:2:p:365-386}
Sengupta, S. and Y.~Chen (2018).
\newblock A block model for node popularity in networks with community
  structure.
\newblock {\em Journal of the Royal Statistical Society Series B\/}~{\em
  80\/}(2), 365--386.

\bibitem[\protect\citeauthoryear{Soltanolkotabi and Candes}{Soltanolkotabi and
  Candes}{2012}]{soltanolkotabi2012}
Soltanolkotabi, M. and E.~J. Candes (2012).
\newblock A geometric analysis of subspace clustering with outliers.
\newblock {\em Ann. Statist.\/}~{\em 40\/}(4), 2195--2238.

\bibitem[\protect\citeauthoryear{Soltanolkotabi, Elhamifar, and
  Candes}{Soltanolkotabi et~al.}{2014}]{soltanolkotabi2014}
Soltanolkotabi, M., E.~Elhamifar, and E.~J. Candes (2014).
\newblock Robust subspace clustering.
\newblock {\em Ann. Statist.\/}~{\em 42\/}(2), 669--699.

\bibitem[\protect\citeauthoryear{Tropp}{Tropp}{2012}]{Tropp2012UserFriendlyTF}
Tropp, J.~A. (2012).
\newblock {\em User-Friendly Tools for Random Matrices: An Introduction}.

\bibitem[\protect\citeauthoryear{Tseng}{Tseng}{2000}]{tseng2000nearest}
Tseng, P. (2000).
\newblock Nearest q-flat to m points.
\newblock {\em Journal of Optimization Theory and Applications\/}~{\em
  105\/}(1), 249--252.

\bibitem[\protect\citeauthoryear{Vidal}{Vidal}{2011}]{vidal2011subspace}
Vidal, R. (2011).
\newblock Subspace clustering.
\newblock {\em IEEE Signal Processing Magazine\/}~{\em 28\/}(2), 52--68.

\bibitem[\protect\citeauthoryear{Vidal, Ma, and Sastry}{Vidal
  et~al.}{2005}]{vidal2005generalized}
Vidal, R., Y.~Ma, and S.~Sastry (2005).
\newblock Generalized principal component analysis (gpca).
\newblock {\em IEEE Trans. Pattern Anal. Mach. Intell.\/}~{\em 27\/}(12),
  1945--1959.

\bibitem[\protect\citeauthoryear{von Luxburg}{von
  Luxburg}{2007}]{vonLuxburg2007}
von Luxburg, U. (2007, Dec).
\newblock A tutorial on spectral clustering.
\newblock {\em Statistics and Computing\/}~{\em 17\/}(4), 395--416.

\bibitem[\protect\citeauthoryear{Wang and Zeng}{Wang and
  Zeng}{2019}]{NEURIPS2019_9be40cee}
Wang, M. and Y.~Zeng (2019).
\newblock Multiway clustering via tensor block models.
\newblock In H.~Wallach, H.~Larochelle, A.~Beygelzimer, F.~Alch\'{e}-Buc,
  E.~Fox, and R.~Garnett (Eds.), {\em Advances in Neural Information Processing
  Systems}, Volume~32. Curran Associates, Inc.

\bibitem[\protect\citeauthoryear{Wang, Wang, and Singh}{Wang
  et~al.}{2016}]{pmlr-v51-wang16b}
Wang, Y., Y.-X. Wang, and A.~Singh (2016).
\newblock Graph connectivity in noisy sparse subspace clustering.
\newblock In A.~Gretton and C.~C. Robert (Eds.), {\em Proceedings of the 19th
  International Conference on Artificial Intelligence and Statistics},
  Volume~51 of {\em Proceedings of Machine Learning Research}, Cadiz, Spain,
  pp.\  538--546. PMLR.

\bibitem[\protect\citeauthoryear{Wang and Xu}{Wang and
  Xu}{2016}]{10.5555/2946645.2946657}
Wang, Y.-X. and H.~Xu (2016).
\newblock Noisy sparse subspace clustering.
\newblock {\em J. Mach. Learn. Res.\/}~{\em 17\/}(1), 320--360.

\bibitem[\protect\citeauthoryear{Zhang, Szlam, Wang, and Lerman}{Zhang
  et~al.}{2012}]{Zhang2012}
Zhang, T., A.~Szlam, Y.~Wang, and G.~Lerman (2012, Dec).
\newblock Hybrid linear modeling via local best-fit flats.
\newblock {\em International Journal of Computer Vision\/}~{\em 100\/}(3),
  217--240.

\bibitem[\protect\citeauthoryear{Zhu and Ghodsi}{Zhu and
  Ghodsi}{2006}]{ZHU2006918}
Zhu, M. and A.~Ghodsi (2006).
\newblock Automatic dimensionality selection from the scree plot via the use of
  profile likelihood.
\newblock {\em Computational Statistics \& Data Analysis\/}~{\em 51\/}(2),
  918--930.

\end{thebibliography}


\end{document}